\documentclass[12pt]{elsarticle}
\usepackage[pdfborder={0 0 0}]{hyperref}
\usepackage{bm}
\usepackage{listings}
\usepackage{xcolor}
\definecolor{codegreen}{rgb}{0,0.6,0}
\definecolor{codegray}{rgb}{0.5,0.5,0.5}
\definecolor{codepurple}{rgb}{0.58,0,0.82}
\definecolor{backcolour}{rgb}{0.95,0.95,0.92}
\lstdefinestyle{mystyle}{
    backgroundcolor=\color{backcolour},   
    commentstyle=\color{codegreen},
    keywordstyle=\color{magenta},
    numberstyle=\tiny\color{codegray},
    stringstyle=\color{codepurple},
    basicstyle=\ttfamily\footnotesize,
    breakatwhitespace=false,         
    breaklines=true,                 
    captionpos=b,                    
    keepspaces=true,
    numbersep=5pt,                  
    showspaces=false,                
    showstringspaces=false,
    showtabs=false,                  
    tabsize=2
}
\makeatletter
\def\ps@pprintTitle{%
	\let\@oddhead\@empty
	\let\@evenhead\@empty
	\let\@oddfoot\@empty
	\let\@evenfoot\@oddfoot
}
\makeatother
\usepackage{graphics,fancyhdr,graphicx}
\usepackage{amsthm,amsmath,amsfonts,latexsym,amssymb}
\usepackage{lineno}
\usepackage[ruled,linesnumbered]{algorithm2e}
\usepackage[margin=1in]{geometry}
\usepackage{multirow,booktabs}
\usepackage{float}
\usepackage{tikz}
\usepackage{listings}
\usepackage{lscape}
\usepackage{tabularx}
\usepackage{caption}
\usepackage[skip=0.15pt]{subcaption}
\usepackage{multirow}
\usepackage{tikz}
\modulolinenumbers[5]

\newcommand{\tsr}[1]{\ensuremath\boldsymbol{#1}}

\newcommand{\wh}[1]{\widehat{#1}}

\def\equationautorefname~#1\null{%
  Eq.~(#1)\null
  }
\def\subfigureautorefname~#1\null{%
  Fig.~#1\null
}

\definecolor{listinggray}{gray}{0.9}
\definecolor{lbcolor}{rgb}{0.9,0.9,0.9}
\definecolor{Darkgreen}{RGB}{0,100,0}

\begin{document}
\abovedisplayskip=6.0pt
\belowdisplayskip=6.0pt
\begin{frontmatter}

\title{An Energy Approach to the Solution of Partial Differential Equations in Computational Mechanics via Machine Learning: Concepts, Implementation and Applications}

     \author[rvt]{E. Samaniego}
\address[rvt]{School of Engineering and Departamento de Recursos
	H\'idricos y Ciencias Ambientales, Universidad de Cuenca, Av. 12
	de Abril s/n., Cuenca, Ecuador}
\author[els]{C. Anitescu}
\address[els]{Institute of Structural Mechanics, Bauhaus-Universit\"at Weimar, Marienstra{\ss}e 15 99423 Weimar}
\author[els]{S. Goswami}
\author[hls]{V.M. Nguyen-Thanh}
\address[hls]{Institute of Continuum Mechanics, Leibniz Universit\"at Hannover, Appelstra{\ss}e 11, 30167 Hannover, Germany}
\author[hls]{H. Guo}
\author[hls]{K. Hamdia}
\author[els]{T. Rabczuk\corref{cor1}}
\cortext[cor1]{Corresponding author.}
\ead{timon.rabczuk@uni-weimar.de}
\author[hls]{X. Zhuang}

\begin{abstract}
Partial Differential Equations (PDE) are fundamental to model different phenomena in science and engineering mathematically. Solving them is a crucial step towards a precise knowledge of the behaviour of natural and engineered systems. In general, in order to solve PDEs that represent real systems to an acceptable degree, analytical methods are usually not enough. One has to resort to discretization methods. For engineering problems, probably the best known option is the finite element method (FEM). However, powerful alternatives such as mesh-free methods and Isogeometric Analysis (IGA) are also available. The fundamental idea is to approximate the solution of the PDE by means of  functions specifically built to have some desirable properties. In this contribution, we explore  Deep Neural Networks (DNNs) as an option for approximation. They have shown impressive results in areas such as visual recognition. DNNs are regarded here as function approximation machines. There is great flexibility to define their structure and important advances in the architecture and the efficiency of the algorithms to implement them make DNNs a very interesting alternative to approximate the solution of a PDE. We concentrate in applications that have an interest for Computational Mechanics. Most contributions that have decided to explore this possibility have adopted a collocation strategy. In this contribution, we concentrate in mechanical problems and analyze the energetic format of the PDE. The energy of a mechanical system seems to be the natural loss function for a machine learning method to approach a mechanical problem. As proofs of concept, we deal with several problems and explore the capabilities of the method for applications in engineering.
\end{abstract}


\end{frontmatter}

\section{Introduction}
Computational mechanics aims at solving mechanical problems using computer methods. These mechanical problems can originate from the study of either natural or engineered systems. In order to describe their behaviour in a precise manner, mathematical models have to be devised. In engineering applications, these mathematical models are often based on partial differential equations (PDEs). When realistic models are considered, one has to resort to numerical methods to solve them. The idea is to look for an approximate solution for the problem, in a finite-dimensional space. Then, the problem reduces to find a finite set of parameters that define this approximate solution. Conventional ways to tackle the solution of PDEs are the finite element method (FEM) \cite{hughes2012finite}, mesh-free methods \cite{huerta2018meshfree}, and isogeometric analysis \citep{hughes2018isogeometric}. 

In recent times, the use of deep neural networks (DNNs) have led to outstanding achievements in several areas, such as visual recognition \citep{goodfellow2016deep}. Feed-forward neural networks are devised to approximate the target functions. The approximating functions depend on certain parameters (the weights and the biases) that have to be "learned" by means of a "training" process. Then, it is  conceivable that Neural Networks may be used to approximate the solution of a PDE. This is the perspective that we are going to adopt with respect to DNNs in this article: they are function approximation machines.

The success of DNNs in important learning tasks can be related to the development of very powerful computational tools. Libraries such as Tensorflow \citep{abadi2016tensorflow} and PyTorch \citep{ketkar2017python} provide the building blocks to devise learning machines for very different problems. Interfaces that allow to code in a readable languages like Python and the availability of optimized numerical algorithms leads to achieving a near-mathematical notation, reminiscent of computing plataforms like FEniCS \citep{alnaes2014unified}.

There are some works in which the idea of using DNNs to solve PDEs has been pursued \citep{raissi2018deep}. However, in general, they have dealt with the strong form of the PDE, leading to an approach based on collocation \citep{raissi2017physics}. Then, the training process is based in devising an objective function, the empirical loss function, whose minimization leads to the fulfilment of the governing equations. Also, the problems dealt with are not in general problems directly related to engineering applications.

A way to approach the solution of a PDE is to write it as a variational problem \citep{dacorogna2014introduction}. From the mechanical point of view, the corresponding functional has the meaning of an energy. Given the fact that the training process in machine learning can be regarded as a process of minimizing the loss function, it seems natural to regard the energy of the system as a very good candidate for this loss function. In addition, the near-mathematical syntax achieved by the platforms associated to Tensorflow and Pytorch imply a high degree of readability and ease of implementation of the PDE solver.  Moreover, once the variational problem is approximated in a finite-dimensional space, it becomes an optimization problem, which is very convenient given that machine learning libraries are specially oriented to optimization techniques.

It is worth mentioning that our approach points at solving the PDEs by means of DNNs as an approximation strategy. In that respect, what we propose is different from approaches such as \cite{baiges2019finite}. They use labeled data from numerical simulations (although it could be obtained from experiments, in principle) to help the solution of a Boundary Value Problem in some specific aspect, where detailed knowledge of the phenomena that is being modelled is lacking. For instance, \citep{ortiz2016data} replaces the constitutive model for a data-driven model. In summary, they use machine learning to build a surrogate model, while we use it to build the approximation space.

In this work, we explore the possibility of using a DNN based solver for PDEs. The paper is organized as follows. In \autoref{sec:generalized}, we introduce the reader to the generalized problem setup. \autoref{sec:deep_neural_network} provides a brief introduction to DNN. The strategy for solving PDEs with DNNs based on collocation method and deep energy method is explained in \autoref{sec:solution_strat}. The implementation technique is explained in \autoref{sec:implementation}. To explain the implementation, we have used snippets from the code. Some representative applications in computational mechanics are tackled in \autoref{sec:application}, to explore the possibilities of this approach. Finally, \autoref{sec:conclusion} concludes the study by summarizing the key results of the present work.

\section{Mathematical modeling of continuous physical systems}
The general aim of this work is to set the foundations for a new paradigm in the field of computational mechanics that enriches deep learning with the long standing developments in mathematical physics. To that end, we first consider a generalized PDE and later briefly introduce the energy approach.
\label{sec:generalized}
\subsection{Partial Differential Equations}
Consider a generalized PDE expressed as:
\begin{equation}\label{BVP_SF}
\mathcal{N}[\boldsymbol{u}]=0\;\;\text{ on }\;\;\Omega,
\end{equation}
where $\mathcal{N}$ stands for a (possibly non-linear) differential operator acting upon a (possibly vector-valued) function, $\boldsymbol{u}$. In addition, $\boldsymbol{u}$ must satisfy both the Dirichlet and the Neumann-boundary conditions. The solution of this PDE subjected to appropriate boundary conditions is referred to a boundary value problem (BVP). Many problems in science and engineering can be written as specific cases of this generalized format. Sometimes, this version of the corresponding BVP is called the strong form.

Let us consider how this general setting can be applied to the case of a linear elastic body. We will have the displacement, $\boldsymbol{u}$ as the primal variable. Then, the second order total strain tensor,$\boldsymbol{\epsilon}(\boldsymbol{u})$ is defined as follows:
\begin{equation} \label{Strain}
\boldsymbol{\epsilon}(\boldsymbol{u})=\frac{1}{2}\bigg(\nabla\boldsymbol{u}+\nabla\boldsymbol{u}^{T}\bigg).
\end{equation}
For the sake of simplicity, in \autoref{Strain} we have dropped the explicit dependence of the involved fields on position, $\boldsymbol{x}$ and time, $t$. The Cauchy stress tensor, $\boldsymbol{\sigma}(\boldsymbol{\epsilon})$ can be computed as:
\begin{equation}
\label{Stress1}
\boldsymbol{\sigma}(\boldsymbol{\epsilon})=\mathbb{C}:\boldsymbol{\epsilon},
\end{equation}
where $\mathbb{C}$ is the elasticity tensor. The momentum balance equation, neglecting the inertial effects and considering zero body forces reads as:
\begin{equation}
\label{Balance}
\nabla \cdot \boldsymbol{\sigma}=\boldsymbol{0}.
\end{equation}
Hence, we define \autoref{BVP_SF} as:
\begin{equation}\label{Operator}
\mathcal{N}[\boldsymbol{u}]=\nabla \cdot \boldsymbol{\sigma}\big(\boldsymbol{\epsilon}(\boldsymbol{u})\big),
\end{equation}
which for sufficiently smooth $\boldsymbol{u}$ can be solved by a collocation-type method as detailed in \autoref{sec:solution_strat}. 

\subsection{Energy Approach}
For the energy approach to a BVP, consider the following problem
\begin{equation}
\label{BVP_EF}
\underset{\boldsymbol{u}}{\operatorname{min}} \, \mathcal{E}[\boldsymbol{u}],
\end{equation}
where $\boldsymbol{u}$ is constrained by Dirichlet boundary conditions. Here, we will assume that the total variational energy of the system, $\mathcal{E}$ is such that there exists a unique solution of the problem defined in \autoref{BVP_EF} and is the same as the solution of \autoref{BVP_SF}. In that case, \autoref{BVP_SF} is called the Euler-Lagrange equation of the variational problem defined in \autoref{BVP_EF}. 

The first step to go from the variational energy formulation to the corresponding strong form is to find a stationary state of $\mathcal{E}$ by equating its first variation to zero: 
\begin{equation}\label{variation}
\delta \mathcal{E}[\boldsymbol{u},\delta \boldsymbol{u}] = \left. \frac{\mathrm{d}}{\mathrm{d} \tau} \mathcal{E}[\boldsymbol{u}+\tau \delta \boldsymbol{u}] \right|_{\tau=0} = 0,
\end{equation}
which has to hold for any admissible $\delta \boldsymbol{u}$ (i.e., for any smooth enough function having homogeneous boundary conditions on the Dirichlet part of the boundary). The resulting expression is the so-called weak form of the problem defined by \autoref{BVP_SF} plus appropriate boundary conditions. In mechanical problems, this corresponds to the principle of virtual work. The weak form is the point of departure of important discretization methods such as the FEM. A characteristic feature of the approach used in this contribution is that it deals directly with the minimization of the variational energy, $\mathcal{E}$ circumventing the need to derive the weak form from the energy of the system explicitly.

In order to illustrate this approach, let us again consider a linear elastic body. One can define the stored elastic strain energy, $\Psi(\boldsymbol{\epsilon})$ as:
\begin{equation} \label{Energy_Density}
\Psi(\boldsymbol{\epsilon})=\frac{1}{2}\boldsymbol{\epsilon}:\mathbb{C}:\boldsymbol{\epsilon}.
\end{equation}
Notice that the Cauchy stress tensor, $\boldsymbol{\sigma}(\boldsymbol{\epsilon})$ can be computed as follows:
\begin{equation} \label{Cauchy}
\boldsymbol{\sigma}(\boldsymbol{\epsilon})=\frac{\partial\Psi}{\partial\boldsymbol{\epsilon}}.
\end{equation}

Then, for the case of zero body forces and homogeneous Neumann boundary conditions, we can define the energy of the solid as
\begin{equation}
\label{elast_energy}
\mathcal{E}[\boldsymbol{u}]=\int_{\Omega} \Psi \big(\boldsymbol{\epsilon}(\boldsymbol{u})\big)d\Omega.
\end{equation}

\section{Deep Neural Networks for PDE discretization}
\label{sec:deep_neural_network}
Artificial Neural Networks (ANNs) are learning machines loosely inspired by the biological neural networks. The general idea is to transform some given input into an output. For this, there are several units, called neurons, whose state can change depending on the inputs and the functional relations between these states. ANNs are usually represented by a graph, whose nodes are the neurons. 

Given an input $\boldsymbol{x}$, an ANN would map it to an output $\boldsymbol{u}_{p}(\boldsymbol{x})$. The functional structure of the ANN is defined by a set of parameters (which can be collected in a vector, $\boldsymbol{p}$). Deep Neural Networks are ANNs obtained by means of the composition of simple functions:
\begin{equation} \label{DNN_mapping}
\boldsymbol{u}_{p}(\boldsymbol{x})=A_L(\sigma(A_{L-1}(\sigma(\ldots\sigma(A_1(\boldsymbol{x}))\ldots)))),
\end{equation}
where $A_l$ (with $l=1,2,\ldots,L$) are affine mappings and $\sigma$ is a the activation function applied element-wise. It is important to note that the non-linearity of the DNNs come from the activation function. Due to the graphical representation of the composition, each of these functions is called a layer, where $L$ denotes the number of layers of the network. So, loosely speaking, we say that an ANN is a DNN with one layer, barring the input and the output layers..

Each affine mapping $A_l$ can be defined by a (in general not square) matrix and a vector. The elements of the matrix are called $\it{weights}$ and the elements of the vector are called $\it{biases}$. They together parametrize the function $\boldsymbol{u}_{p}(\boldsymbol{x})$, and can be grouped in a vector $\boldsymbol{p}$. In the following, we consider feed-forward fully-connected DNNs, where each neuron is connected to all the neurons in the adjacent layers. Other types of neural networks can be considered as well, such as convolutional neural networks \cite{Ruthotto2018deep}, where the layers and the connections between them are organized in an hierarchical structure.

Once the architecture of the network has been decided, that is, once the number of layers, the number of neurons per layer, and the activation functions have been frozen, defining $\boldsymbol{u}_{p}(\boldsymbol{x})$ boils down to determining $\boldsymbol{p}$. The process of determining these parameters (the weights and the biases) is called $training$ the network. 
Usually, when solving problems with machine learning, this entails having a large amount of input and output data.  In the approach used here, data for training is generated from evaluations of the PDEs at given point in the domain, as well as from certain points on the boundary. This issue is explained more in detail below.

\section{Solution strategies} \label{sec:solution_strat}
\subsection{Collocation Methods} 
Collocation methods are based on the solution of the BVP in strong form. Given a set of points $\{\boldsymbol{x}_1,\boldsymbol{x}_2,\ldots,\boldsymbol{x}_n \}$ in the domain $\Omega$, the idea is to evaluate $\mathcal{N}[\boldsymbol{u}_{p}]$ at each point $\boldsymbol{x}_i$ and construct a loss function whose optimization tends to impose the condition $\mathcal{N}[\boldsymbol{u}_{p}](\boldsymbol{x}_i)=0$. For instance, loss functions based on the mean square error of $\mathcal{N}[\boldsymbol{u}_{p}](\boldsymbol{x}_i)$ can be used.

\subsection{Deep Energy Method}
The main idea of the method advocated in this contribution is to take advantage of the variational (energetic) structure of some BVPs. To that end, the energy of the system is used as the loss function for the DNN, as proposed by \citep{weinan2018deepRitz}. Due to its mechanical flavor, we name it the Deep Energy Method (DEM) here. One of the key ingredients is to approximate the energy of the body by a weighted sum of the energy density at integration points. Then, the following form for the loss function, $\mathcal{L}(\boldsymbol{p})$ is obtained:
\begin{equation}\label{elast_energy2}
\mathcal{E}[\boldsymbol{u}_{p}] \approx \mathcal{L}(\boldsymbol{p}) = \sum_{i}  \Psi \big(\boldsymbol{\epsilon}(\boldsymbol{u}_{p}(\boldsymbol{x}_i))\big)w_i,
\end{equation}
where $\boldsymbol{u}_{p}(\boldsymbol{x}_i)$ is the displacement approximation function evaluated at the integration point, $\boldsymbol{x}_i$ and its corresponding weight, $w_i$. A key issue is the prescription of the boundary conditions. In an variational energy formulation, Neumann boundary conditions can in principle be imposed in a natural way. However, Dirichlet boundary conditions have to be imposed in a somehow more involved way, since DNN approximation functions do not fulfill the Kronecker delta property. We will address these issues in \autoref{sec:application}.

\subsubsection{Optimization in machine learning}
Once the architecture of a DNN and the empirical loss function have been selected, finding the parameters $\boldsymbol{p}$ (weights and biases) that define the approximation $\boldsymbol{u}_p$ of the the unknown $\boldsymbol{u}$ is the main task. Then, having in mind that the loss function can be expressed as
\begin{equation}
\label{finite_sum_loss}
\mathcal{L}(\boldsymbol{p}) = \sum_{i}  f_i(\boldsymbol{p}),
\end{equation}
we end up facing a non-convex finite-sum optimization problem \citep{Jordan2017finitesum}:
\begin{equation}
\label{finite_sum_optim}
\underset{\boldsymbol{p}}{\operatorname{min}} \, \sum_{i}  f_i(\boldsymbol{p}).
\end{equation}
In machine learning, first order methods are almost universal. In fact, one may say that most optimization strategies in machine learning are variants of the so-called gradient descent method. A typical iteration of this method is the following:
\begin{equation}
\label{GD}
\boldsymbol{p}^{k+1}= \boldsymbol{p}^k + \gamma_k \nabla_p \left( \sum_{i}  f_i(\boldsymbol{p}) \right),
\end{equation}
where $\boldsymbol{u}_p^k$ is the DNN approximation of the unknown $\boldsymbol{u}$ parametrized by vector $ \boldsymbol{p}^k$, which groups all weights and biases at iteration $k$. In addition, $\gamma_k$ stands for a parameter called learning rate. Several variations of this algorithm can be considered. First of all, not all the points are always considered. In fact, stochasticity in the selection of the points considered in the finite sum is a general practice, resulting in the so-called stochastic gradient descent (SGD) algorithm. Acceleration strategies are also very common. Another very common variant implies the addition of a penalty term to avoid unbounded values of the parameters. Sometimes, even some approximation of second order information is included, for example, by means of the L-BFGS method, a quasi-Newton method.

One important issue related to the optimization problem is the structure of the search space. Recent mathematical results describe the presence of highly unfavorable properties of the topological structure of DNN spaces. This means that, although the space of functions generated by a DNN can be very expressive, the structure of the search space may cause difficulties for optimization algorithms \citep{petersen2018topological}. However, some unexpected properties of the gradient-descent-based algorithms seem to be able to overcome these difficulties \citep{jordan2018dynamical}, at least for some applications. This an issue that is the object of intense research currently. 

\section{Implementation}
\label{sec:implementation}

In this section, we illustrate how the proposed DEM approach can be used for solving problems in the field of continuum mechanics. The primary goal of the approach is to obtain the field variables by solving the governing partial differential equation, which in turn requires to compute the integral \autoref{elast_energy}. We do this by using machine learning tools available in open-source libraries like TensorFlow and PyTorch. This allows us to use collective knowledge accumulated in a very vibrant community. 

We describe the typical steps that one would have to follow for a TensorFlow implementation. The idea is to illustrate the general procedure. Specific details for each application are given below. The full source code will be published on GitHub or can be obtained by contacting the authors. 

First, the geometry is generated using uniformly spaced points in the whole domain and obtain the training points, $\bm{X}_f$ and prediction points, $\bm{x}_{pred}$.
Before we start to train the network, we initialize the weights of the network randomly from a Gaussian distribution using the Xavier initialization technique \cite{glorot2010understanding}. The network is initialized in the following manner:

\begin{lstlisting}[language=Python]
    def initialize_NN(self,layers):
        weights = []
        biases = []
        num_layers = len(layers)
        for l in range(0, num_layers - 1):
            W = self.xavier_init(size=[layers[l], layers[l + 1]])
            b = tf.Variable(tf.zeros([1,layers[l + 1]]),
                dtype=tf.float32)
            weights.append(W)
            biases.append(b)
        return weights, biases

    def xavier_init(self, size):
        in_dim = size[0]
        out_dim = size[1]
        xavier_stddev = np.sqrt(2.0 / (in_dim + out_dim))
        w_init = tf.Variable(tf.truncated_normal([in_dim, out_dim], 
                stddev=xavier_stddev))
        return w_init
\end{lstlisting}
where the network is initialized with $layers$. Once the weights are initialized, we begin training the neural network. 

Next, we modify the neural network outputs in such a way so that the Dirichlet boundary conditions are exactly satisfied. This can be done as in the following listing, where $uNN$ and $vNN$ denote the output from the neural network.
\begin{lstlisting}[language=Python]
    def net_uv(self,x,y,vdelta):
        X = tf.concat([x,y],1)
        uv = self.neural_net(X,self.weights,self.biases)
        uNN = uv[:,0:1]
        vNN = uv[:,1:2]
        u = (1-x)*x*uNN
        v = y*(y-1)*vNN
        return u, v
\end{lstlisting}
 The neural network subroutine can be defined according to the type of activation function chosen between subsequent layers as in the following example:
\begin{lstlisting}[language=Python]
    def neural_net(self,X,weights,biases):
        num_layers = len(weights) + 1
		
        H = 2.0 * (X - self.lb) / (self.ub - self.lb) - 1.0
        for l in range(0, num_layers - 2):
            W = weights[l]
            b = biases[l]
            H = tf.nn.relu(tf.add(tf.matmul(H, W), b))**2
        W = weights[-1]
        b = biases[-1]
        Y = tf.add(tf.matmul(H, W), b)
        return Y
\end{lstlisting}
Here $self.lb$ and $self.ub$ denote the lower bound and the upper bound of the inputs to the neural network, which implies the spatial co-ordinates.

In the next step, automatic differentiation is used to compute the displacement gradients and compute the variational energy at any point in the domain.
\begin{lstlisting}[language=Python]
    def net_energy(self,x,y):
        u, v = self.net_uv(x,y)
        u_x = tf.gradients(u,x)[0]
        v_y = tf.gradients(v,y)[0]
        u_y = tf.gradients(u,y)[0]
        v_x = tf.gradients(v,x)[0]
        u_xy = (u_y + v_x)
        sigmaX = self.c11*u_x + self.c12*v_y
        sigmaY = self.c21*u_x + self.c22*v_y
        tauXY = self.c33*u_xy
        energy = 0.5*(sigmaX*u_x + sigmaY*v_y + tauXY*u_xy)
        return energy
\end{lstlisting}
Finally, we minimize the mean error of the energy losses computed in the previous step. For optimization, we use the Adam (adaptive momentum) optimizer followed by a quasi-Newton method (L-BFGS). Once the network is optimized, we predict the values of the field variables at $x_{pred}$ points.

\section{Applications}
\label{sec:application}
In this section, we explore the application of DEM to solve PDEs in various domains of continuum mechanics. The first section considers a one-dimensional problem and the results obtained using DEM is compared to the available analytical solution. Later, we discuss the application of DEM on linear elasticity, hyperelasticity. In the latter part of the section, we solve PDEs for coupled problems. For coupled problems, we address phase-field modeling of fracture, piezoelectricity and lastly the bending of a Kirchhoff plate.

The implementation has been carried out using the \texttt{TensorFlow} framework \cite{abadi2015tensorflow}. Details on the network architecture, such as number of layers, number of neurons in each layer, the activation function used etc. have been provided with each example. 

\subsection{DNN with ReLU activation functions and FE in 1D}
\label{subsec:DNN_relu}
ANNs are known to be universal approximators. Moreover, recent results in approximation theory show that DNNs using rectified linear units (ReLU) as activation functions can reproduce linear finite element spaces \citep{he2018relu}. ReLU functions are defined by $ReLU(x)=max(0,x)$. These results have been extended to more general Finite Element spaces using Sobolev norms in \citep{petersen2019deep}. Here we illustrate the relationship between FE approximations and DNNs by means of a one-dimensional example, as done in \citep{he2018relu}.

Let us consider a discretization of the interval $[0,l]$ in the real line: $x_0 < x_1 < \ldots < x_n$, where $x_0=0$ and $x_n=l$. Then, consider a piecewise linear FE approximation of a function $u(x)$ taking values on $[0,l]$:

\begin{equation}
\label{uh_FE}
u^h(x)=\sum\limits_{i=0}^{n}u_i N_i(x)
\end{equation}
where $N_i(x)$ is a linear hat function with compact support in the subinterval $[x_{i-1},x_{i+1}]$, such that $N_i(x_i)=1$. For $i=0$ and $i=n$, $N_i(x)$ has compact support in $[x_{0},x_{1}]$ and $[x_{n-1},x_{n}]$, respectively. Function $N_i(x)$, for intermediate nodes, can be expressed in terms of rectified linear unis $ReLU$ as follows:

\begin{equation}
\label{Ni_ReLU}
N_i(x)=\frac{ReLU(x-x_{i-1})-ReLU(x-x_i)}{h_i} - \frac{ReLU(x-x_i)-ReLU(x-x_{i+1})}{h_{i+1}}
\end{equation}
where ${h_i}=x_i-x_{i-1}$

Rearranging terms, we get
\begin{equation}
\label{Ni_ReLU2}
N_i(x)=\frac{1}{h_i} ReLU(x-x_{i-1})-\left(\frac{1}{h_i} + \frac{1}{h_{i+1}} \right) ReLU(x-x_i)+ \frac{1}{h_{i+1}} ReLU(x-x_{i+1})
\end{equation}

Using Eq. \ref{Ni_ReLU2} and Eq. \ref{uh_FE}, the following expression can be obtained:

\begin{equation}
\label{uh_ReLU}
u^h(x)=u_0+\sum\limits_{i=0}^{n-1} \left(\frac{\Delta_{i+1}}{h_{i+1}} - \frac{\Delta_{i}}{h_i} \right) ReLU(x-x_i)
\end{equation}
where $\Delta_{i}=u_i-u_{i-1}$, for $i=1,2,\ldots,n$, and $\Delta_{i}=0$, for $i=0$. Then, $u^h$ can be expressed as an ANN with one hidden layer, having $x$ as input, $u^h(x)$ as output, and $ReLU$ as the activation function:

\begin{equation}
\label{uh_ReLU2}
u^h(x)=u_0+\sum\limits_{i=0}^{n-1}  w_i ReLU(x-x_i)
\end{equation}
with
\begin{equation}
w_i=\left(\frac{\Delta_{i+1}}{h_{i+1}} - \frac{\Delta_{i}}{h_i} \right)
\end{equation}

Notice that $u^h$ is parametrized by the weights $w_i$, which can be grouped in a vector $\boldsymbol{w}$, and by the position of the nodes $x_i$ (the biases of the hidden layer), which can be grouped in the vector $\boldsymbol{x}_p$. For fixed $\boldsymbol{x}_p$, and prescribing $u_0$, the vector value of the elements of $\boldsymbol{w}$ is defined by the parameters $u_i$, i.e., the values of $u^h$ evaluated at the nodes $x_i$.

Consider the approximation in Eq. \eqref{uh_ReLU2}. Then, if $\Psi \big(\epsilon)$ is the one-dimensional linear elastic energy density, we can define an energy \begin{equation}
\label{elast_energy_1D}
\mathcal{E}(\boldsymbol{w},\boldsymbol{x}_p)=\int_{0}^{l} \Psi \big(\epsilon(u^h(x;\boldsymbol{w},\boldsymbol{x}_p))\big)dx.
\end{equation}
in terms of the weights vector $\boldsymbol{w}$ and a vector containing the position of the nodes, $\boldsymbol{x}_p$:

\begin{equation}
\label{BVP_EF_1D}
\min\limits_{\boldsymbol{w},\boldsymbol{x}_p} \mathcal{E}(\boldsymbol{w},\boldsymbol{x}_p)
\end{equation}
For fixed $\boldsymbol{x}_p$, the problem would be equivalent to solving a standard FE problem (remember the relationship between $\boldsymbol{w}$ and the nodal FE unknowns). When $\boldsymbol{x}_p$ is not fixed, from the mathematical point of view, we would be solving an adaptive FE problem. However, since the problem is now non-linear (and probably non-convex), there may be issues related to the use of typical machine learning algorithms to optimize the loss function defined in \ref{BVP_EF_1D}.

Although this 1D problem is illustrative of the meaning of the parameters of a DNN approximation, it is also clear that, for general situations, assigning physical meaning to these parameters would be very difficult. This is may be a consequence of the mesh-free character of DNN approximations for the solution of PDEs. This has several implications. Collocation points, for instance, are not to be confused with nodes in a Finite Element mesh. Points in a FE mesh are tightly related to the parameters of the approximation space, while collocation points are the inputs for the training process in the DNN approximation.

\subsection{Linear Elasticity problem}
\label{subsec:linear_elasticity}
In this section, the physics informed neural network is developed to solve the governing partial differential equation for linear-elastic problems using DEM approach. The solution of the displacement field is obtained by minimizing the stored elastic strain energy of the system. The problem statement is written as:
\begin{equation}\label{eq:elasticenergy}
    \begin{split}
    \text{Minimize:}\;\;\;\; \mathcal{E} &= \Psi(\epsilon),\\
    \text{where:}\;\;\;\; \Psi(\epsilon) &= \int\limits_\Omega{f(\bm{x})}dx,\\
    \text{and}\;\;\;\; f(\bm{x}) &= \frac{1}{2}\bm{\epsilon}:\mathbb{C}:\bm{\epsilon},\\
    \text{subject to:}\;\;\;\; \bm{u} &= \bm{\overline u} \text{ on } \partial \Omega_{D},\\
    \text{and}\;\;\;\; \bm{\sigma}\cdot \bm{n} &= \bm{t}_N \text{ on } \partial\Omega_{N} \\
    \end{split}
\end{equation}
where $\Psi(\epsilon)$ denotes the stored elastic strain energy of the system expressed in terms of the strain tensor, $\epsilon(\bm{u})$, $\mathbb{C}$ represents the constitutive elastic matrix, $\bm{\overline u}$ is the prescribed displacement on the Dirichlet boundary, $\partial \Omega_{D}$ and $\bm{t}_N$ is the prescribed boundary forces on the Neumann boundary, $\partial \Omega_{N}$. Using the DEM approach, the homogeneous Neumann boundary conditions are automatically satisfied. The field variables are obtained by minimizing the mean error of the energy functional at the integration points. Additionally, the non-homogeneous Neumann boundary conditions are satisfied by minimizing the mean error at the integration points on the boundary. For optimization, the neural network is trained by using a combination of Adam optimizer and second-order quasi-Newton method (L-BFGS). 

We shall now present the results obtained using DEM for two and three-dimensional linear elastic problems. Analytical solution is available for all these problems, hence the accuracy of the method is measured by computing the $\mathcal L_2$ error for the displacement field and the strain energy.

\subsubsection{Pressurized thick-cylinder}
\label{subsec2:LE_prob1}
\begin{figure}
 \centering
 \begin{subfigure}[b] {0.4\textwidth}
   \centering
   \includegraphics[width=0.7\textwidth]{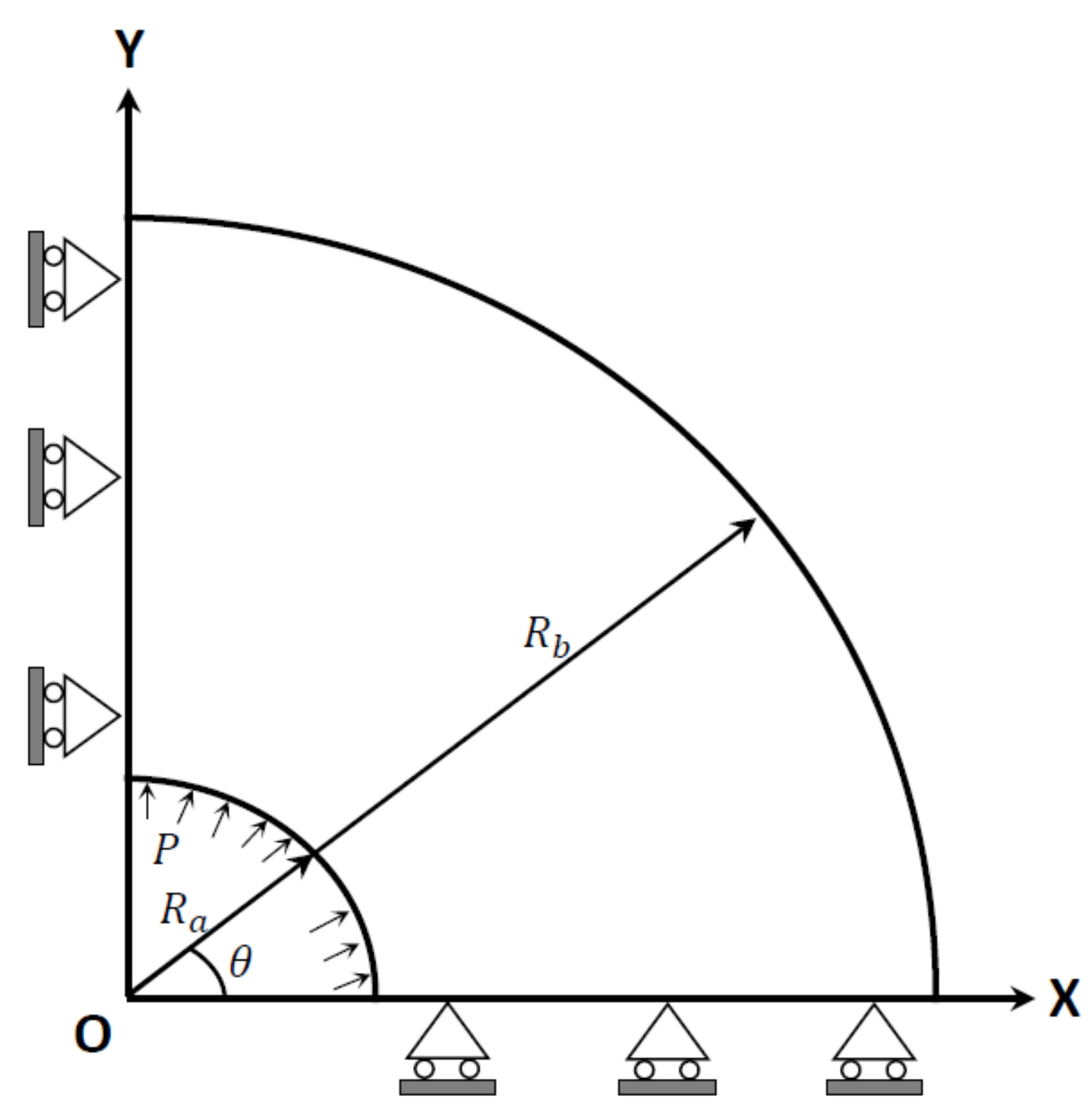}\hfill
    \caption{Geometrical setup and boundary conditions}
 \end{subfigure}\hspace{5pt}
 \begin{subfigure}[b] {0.4\textwidth}
	\centering
	\includegraphics[width=\textwidth]{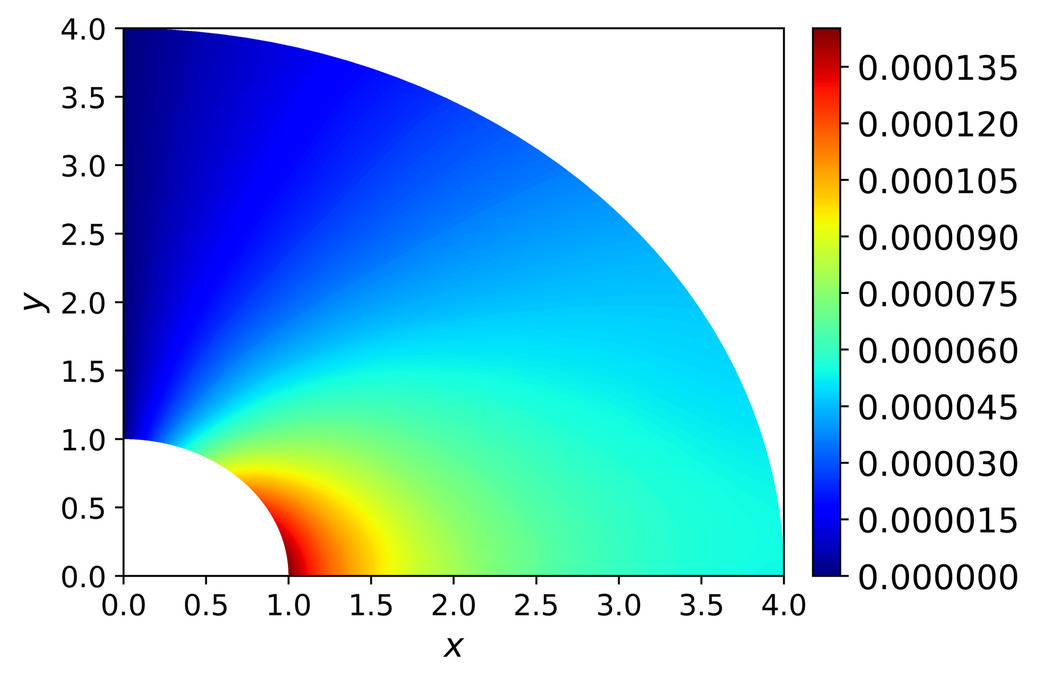}\hfill
	\caption{Displacement in $x$-axis.}
\end{subfigure}\hspace{5pt}
\begin{subfigure}[b] {0.4\textwidth}
	\centering
	\includegraphics[width=\textwidth]{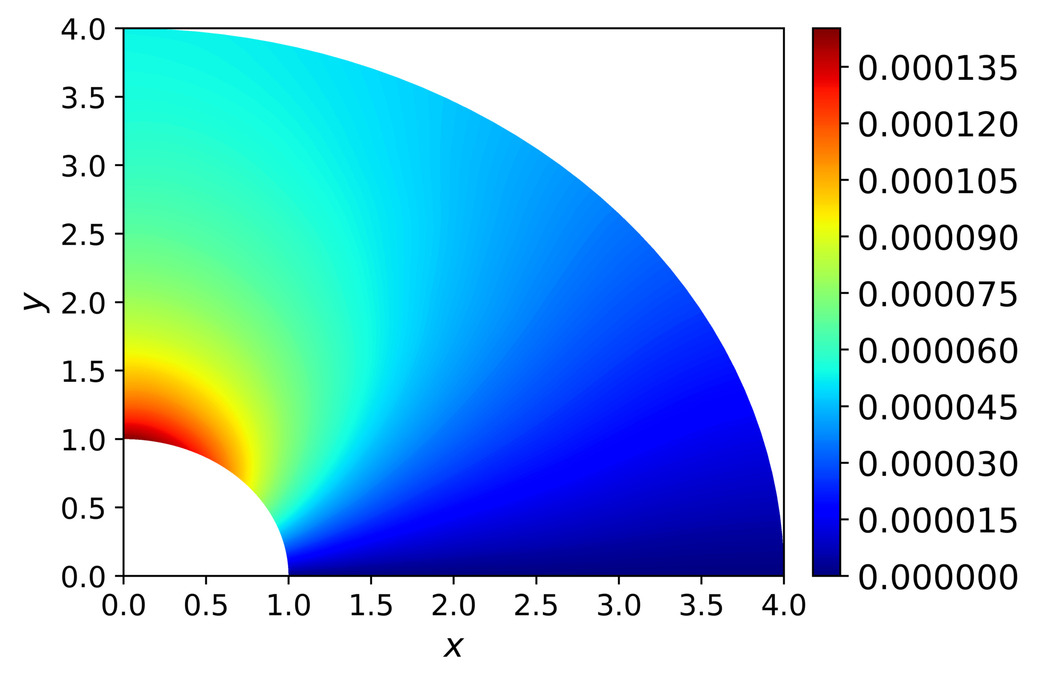}\hfill
	\caption{Displacement in $y$-axis.}
\end{subfigure}\hspace{5pt}
\begin{subfigure}[b] {0.4\textwidth}
	\centering
	\includegraphics[width=\textwidth]{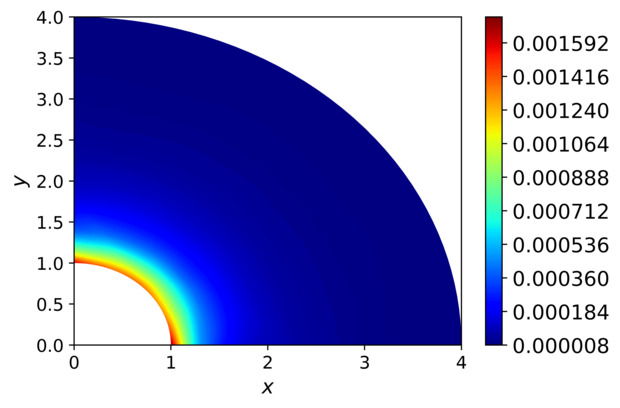}\hfill
	\caption{Strain energy.}
\end{subfigure}\hspace{5pt}
   \caption{Setup and numerical results for the pressurized thick cylinder example}
 \label{fig:thickcylinder}
 \end{figure}

The first example considers the benchmark problem of of a thick cylinder subjected to internal pressure under plane stress condition \cite{jia2019adaptive}. Due to symmetry in the geometry and boundary conditions, only a quarter section represented by an annulus is analyzed. The cylinder is subjected to internal pressure applied on the inner circular edge. The geometrical setup and the boundary conditions are shown in \autoref{fig:thickcylinder}(a). In this example, we have considered $E = 1\times10^{5}$ and $\nu = 0.3$. The analytical solution, in terms of the stress components, is \cite{timoshenko2011theory}:
\begin{equation}\label{eq:analytical_thickcylinder}
    \begin{split}
        \sigma_{rr} &= \frac{R_a^2P}{R_a^2 - R_b^2}\left(1-\frac{R_a^2}{r^2}\right),\\
        \sigma_{\theta \theta} &= \frac{R_a^2P}{R_a^2 - R_b^2}\left(1+\frac{R_a^2}{r^2}\right),\\
        \sigma_{r\theta} &= 0,\\
        u_{rad} &= \frac{R_a^2Pr}{E(R_b^2-R_a^2)}\left(1-\nu+\left(\frac{R_b}{r}\right)^2(1+\nu)\right),\\
        u_{exact} &= u_{rad}\cos \theta,\\
        v_{exact} &= u_{rad}\cos \theta,\\
    \end{split}
\end{equation}
where $R_a$ and $R_b$ represents the inner and the outer radius of the cylinder, respectively with $R_a = 1$ and $R_b = 4$. $P$ is the pressure exerted along the inner circular edge. In \autoref{eq:analytical_thickcylinder}, $u_{exact}$ and $v_{exact}$ are the analytical solution of the displacement field in $\it{x}$ and $\it{y}$-axis, respectively. The minimization problem reads as stated in \autoref{eq:energyterms}. The Dirichlet and Neumann boundary conditions for the thick cylinder under internal pressure are:
\begin{equation}
    \begin{split}
        u(0,y) &= v(x,0) = 0,\\
        t_{N,x}(R_a,\theta) &= Pu,\\
        t_{N,y}(R_a,\theta) &= Pv,\\
    \end{split}
\end{equation}
where $u$ and $v$ are the solutions of the elastic field in $\it{x}$ and $\it{y}$-axis, respectively and $t_N(R_a, \theta)$ denotes the traction force which is expressed in terms of the applied internal pressure and the displacement field. To find the solution of the displacement field, the trial solution is defined as:
\begin{equation}\label{eq:boundary_thickcylinder}
\begin{split}
    u &= x\hat{u},\\
    v &= y\hat{v}, 
\end{split}
\end{equation}
where $\hat{u}$ and $\hat{v}$ are obtained from the neural network. The solution of $u$ and $v$ are chosen in such a way that it satisfies all the boundary conditions as in \cite{avrutskiy2017enhancing, berg2018unified}. As a consequence, no component corresponding to the boundary loss is needed in the loss function. This significantly simplifies the objective function to be minimized.
We have used a network with 3 hidden layers of $30$ neurons each and $N_x\times N_y$ uniformly spaced points in the interior of the annulus, with $N_x = N_y = 80$. The inner circular edge is discretized using $N_{Bound}$ uniformly spaced points to apply the internal pressure in terms of traction force, with $N_{Bound} = 80$. For the input layer and the first two hidden layers, we have considered rectified linear units, \texttt{ReLU}$^2$ activation function; whereas for the last layer connected to the output layer, linear activation function has been considered. The loss function is computed as 
\begin{equation}\label{eq:loss_thickcylinder}
\begin{split}
    \mathcal{L}_{elastic} &= \mathcal{L}_{int} - \mathcal{L}_{neu},\\
    \mathcal{L}_{int} & = \frac{A_{\Omega}}{N_x\times N_y}\sum_{i = 1}^{N_x\times N_y}f(\bm{x}_i),\\
    \mathcal{L}_{neu} & = \frac{A_{\Omega}}{N_{bound}}\sum_{i = 1}^{N_{Bound}}f_{neu}(\bm{x}_i),\\
    f(\bm{x}) &= \frac{1}{2}\bm{\epsilon}:\mathbb{C}:\bm{\epsilon},\\
    f_{neu}(\bm{x}) &= t_{N,x}(\bm{x}) + t_{N,y}(\bm{x}),\\
\end{split}
\end{equation}
where $\mathcal{L}_{int}$ and $\mathcal{L}_{neu}$ defines the losses in the elastic strain energy and the Neumann boundary loss, respectively and $A_{\Omega}$ is the area of the analyzed domain. The predicted displacement field using the DEM approach is shown in \autoref{fig:thickcylinder}(b)-(c). The strain energy obtained using the neural network is presented in \autoref{fig:thickcylinder}(d).
 \begin{figure}
 \centering
 \begin{subfigure}[b] {0.49\textwidth}
   \centering
   \includegraphics[width=\textwidth]{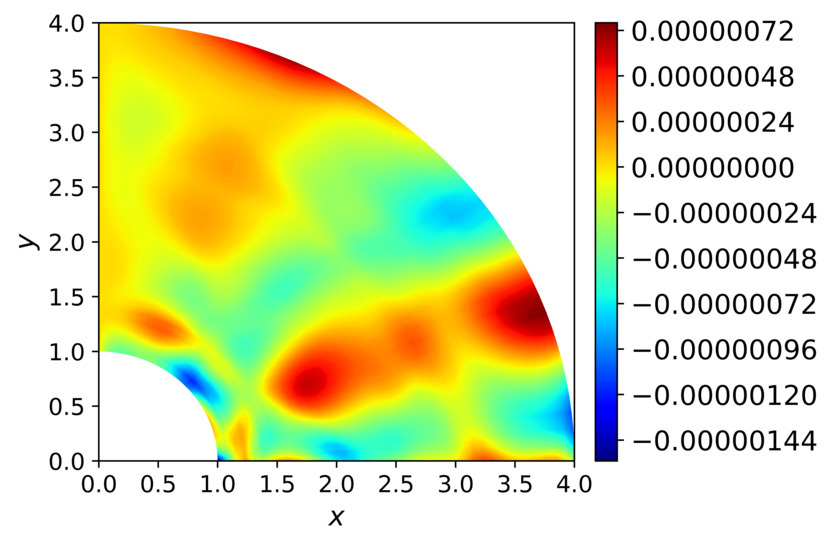}\hfill
   \caption{Error in the displacement field in $x$-axis.}
 \end{subfigure}\hspace{5pt}
 \begin{subfigure}[b] {0.49\textwidth}
   \centering
   \includegraphics[width=\textwidth]{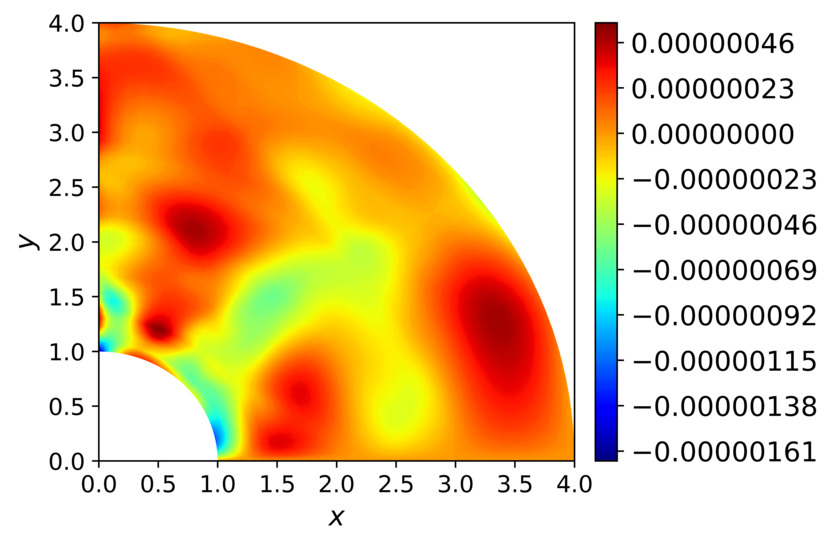}\hfill
   \caption{Error in the displacement field in  $y$-axis.}
 \end{subfigure}\hspace{5pt}
 \caption{Error plots for pressurized thick-cylinder.}
 \label{fig:error_thickcylinder}
 \end{figure}
In order to quantify the accuracy of the results obtained using the proposed approach, we compute the relative $\mathcal L_2$ error as
\begin{equation}\label{eq:L2norm_elastic}
    \mathcal{L}_2^{rel} = \frac{\sqrt{\sum_{i = 1}^{N_{pred}}{\mathcal{V}_{err}^2(x_i)}dx}}{\sqrt{\sum_{i = 1}^{N_{pred}}{\mathcal{V}_{exact}^2(x_i)}dx}},
\end{equation}
where $\mathcal{V}_{exact}$ corresponds to the results obtains using the analytical solution of the field variable, $\mathcal{V}$ and $\mathcal{V}_{err} = \mathcal{V}_{exact} - \mathcal{V}_{pred}$, where $\mathcal{V}_{pred}$ is obtained from the neural network. Corresponding to the displacement and the energy norms, relative prediction errors of 0.5\% and 3.7\%, respectively have been observed. \autoref{fig:error_thickcylinder} shows the errors in the displacement field.

\subsubsection{Plate with a circular hole}
\label{subsec2:LE_prob2}
\begin{figure}[t]
	\centering
	\includegraphics[width=0.35\textwidth]{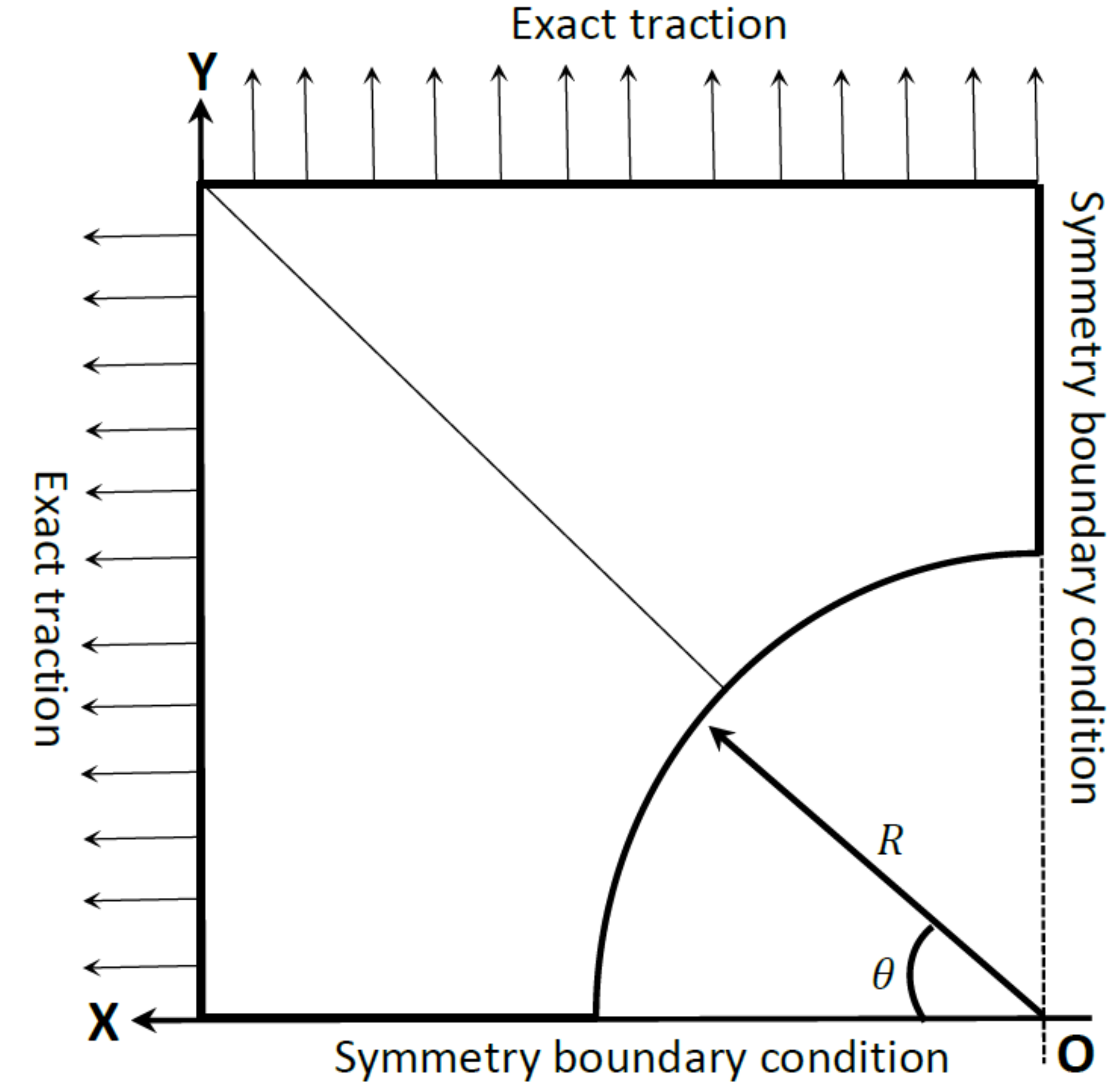}\hfill
	\caption{Problem setup for the plate with a circular hole example}
	\label{fig:setup_platehole}
\end{figure}

\begin{figure}
	\centering
	\begin{subfigure}[b] {0.4\textwidth}
		\centering
		\includegraphics[width=\textwidth]{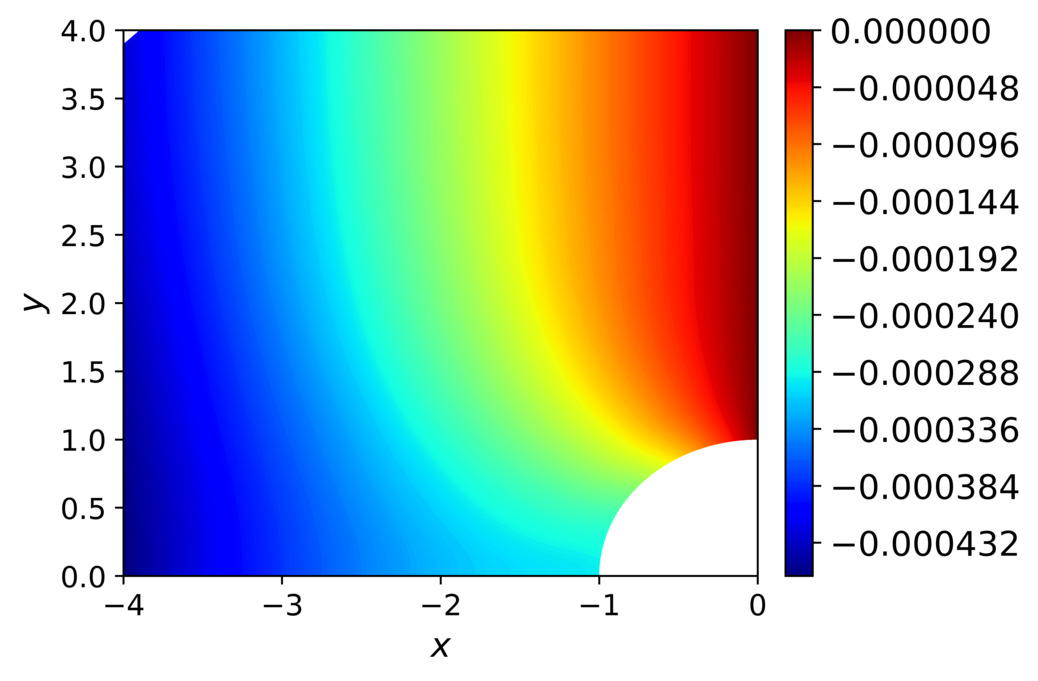}\hfill
		\caption{Displacement in $x$-axis.}
	\end{subfigure}\hspace{5pt}
	\begin{subfigure}[b] {0.4\textwidth}
		\centering
		\includegraphics[width=\textwidth]{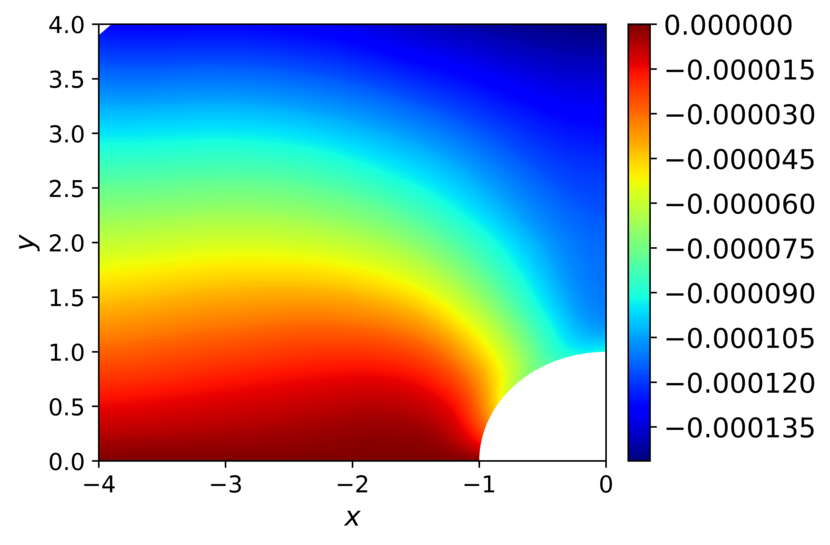}\hfill
		\caption{Displacement in $y$-axis.}
	\end{subfigure}\hspace{5pt}
	\begin{subfigure}[b] {0.4\textwidth}
		\centering
		\includegraphics[width=\textwidth]{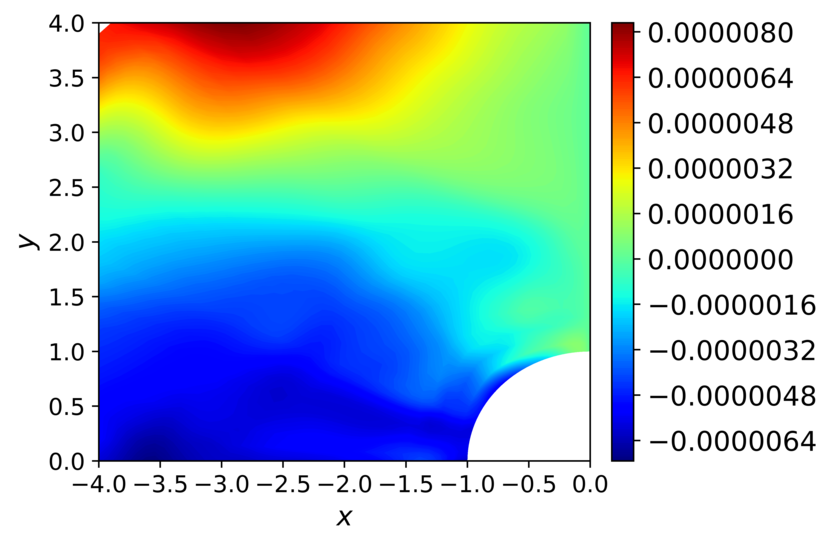}\hfill
		\caption{Error in the displacement field in $x$-axis.}
	\end{subfigure}\hspace{5pt}
	\begin{subfigure}[b] {0.4\textwidth}
		\centering
		\includegraphics[width=\textwidth]{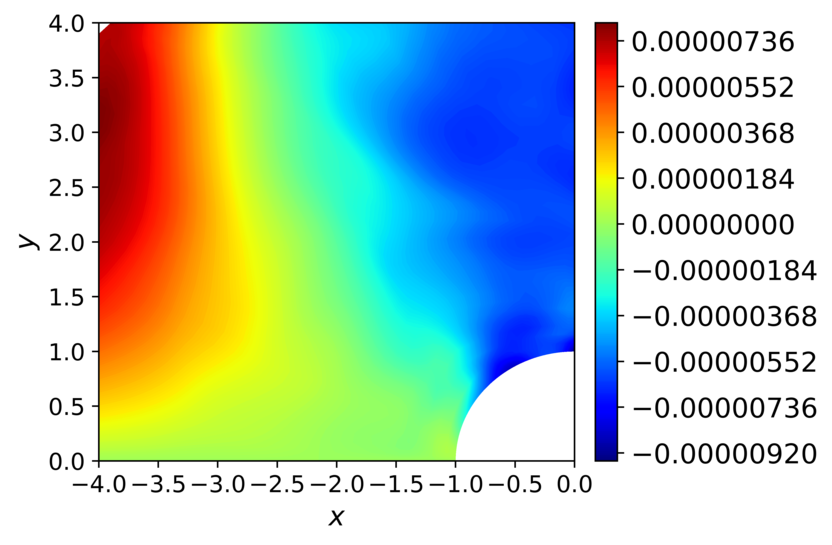}\hfill
		\caption{Error in the displacement field in $y$-axis.}
	\end{subfigure}\hspace{5pt}
	\caption{Computed solution and errors for plate with hole.}
	\label{fig:disp_platewithhole}
\end{figure}

The second problem is a benchmark problem of an infinite plate with a hole, in plane stress. To solve the problem numerically, we consider a finite domain. Exploiting the symmetry of the problem, only one quarter of the plate is analyzed. The geometrical setup and the boundary conditions for the analyzed domain are depicted in \autoref{fig:setup_platehole}. The material properties considered are $E = 1\times 10^5$ and $\nu = 0.3$. A traction force, $T_x = 10$ is applied on the plate as shown in \autoref{fig:setup_platehole}. The analytical solution of the problem, in terms of polar coordinates is given by \cite{gould1994introduction}:
\begin{equation}
    \begin{split}
        \sigma_{rr} &= \frac{T_x}{2}\left(1-\frac{R^2}{r^2}\right)+\frac{T_x}{2}\left(1+3\frac{R^4}{r^4}-4\frac{R^2}{r^2}\right)\cos 2\theta,\\
        \sigma_{\theta \theta} &=  \frac{T_x}{2}\left(1+\frac{R^2}{r^2}\right)-\frac{T_x}{2}\left(1+3\frac{R^4}{r^4}\right)\cos 2\theta,\\
        \sigma_{r\theta} &= -\frac{T_x}{2}\left(1+2\frac{R^2}{r^2}-3\frac{R^4}{r^4}\right)\sin 2\theta,
    \end{split}
\end{equation}
where $R$ denotes the radius of the circular hole and $(r,\theta)$ is used to represent the radial distance and the angle for locating any point on the plate. Representing the stresses in the Cartesian co-ordinate system, we obtain
\begin{equation}
    \left(
    \begin{array}{c}
        \sigma_{xx}(x,y) \\
        \sigma_{yy}(x,y)\\
        \sigma_{xy}(x,y)
    \end{array}
    \right) = A^{-1}\left(\begin{array}{c}
        \sigma_{rr}(r,\theta) \\
        \sigma_{\theta\theta}(r,\theta)\\
        \sigma_{r\theta}(r,\theta)
    \end{array}
    \right),
\end{equation}
where the transformation matrix $A$ is expressed as
\begin{equation}
   A = \left(\begin{array}{ccc}
        \cos ^2 \theta & \sin ^2 \theta & 2\sin \theta \cos \theta \\
        \sin ^2 \theta & \cos ^2 \theta & -2\sin \theta \cos \theta\\
        -\sin \theta \cos \theta & \sin \theta \cos \theta & \cos ^2 \theta - \sin ^2 \theta 
    \end{array}\right).
\end{equation}
 \hspace{5pt}
The minimization problem reads as stated in \autoref{eq:energyterms}. The Dirichlet and Neumann boundary conditions are:
\begin{equation}
    \begin{split}
        u(0,y) &= v(x,0) = 0,\\
        t_{N,x} &= T_x,\\
    \end{split}
\end{equation}
where $u$ and $v$ are the solutions of the elastic field in $\it{x}$ and $\it{y}$-axis, respectively and $t_{N,x}$ denotes the traction force applied on the plate along the $\it{x}$-axis. To find the solution of the displacement field, the trial solution is defined as in \autoref{eq:boundary_thickcylinder}. Similar to the previous example, we have used a network with 3 hidden layers of $30$ neurons each. For the first two layers, we have considered rectified linear units, \texttt{ReLU}$^2$ activation function; whereas for the last layer, linear activation function has been considered. The domain is discretized into $N_x\times N_y$ uniformly spaced points in the interior of the plate and $N_{Bound}$ uniformly spaced points on the edge for applying the traction force, with $N_{Bound} = 80$. We use \autoref{eq:loss_thickcylinder} to compute the loss function to optimize the network. The predicted displacement field for the plate with a circular hole, using the DEM approach, is shown in \autoref{fig:disp_platewithhole} (a)-(b). 

To quantify the accuracy of the neural network, the relative $\mathcal L_2$ error corresponding to $u$ and strain energy has been computed using \autoref{eq:L2norm_elastic}. Corresponding to $u$ and strain energy, a prediction error of 1.8\% and 3.19\%, respectively have been observed. \autoref{fig:disp_platewithhole}(c)-(d) shows the errors in the displacement field.
\subsubsection{Hollow sphere under internal pressure}
\label{subsec2:LE_prob3}
\begin{figure}[ht]
	\centering
	\begin{subfigure}[b] {0.4\textwidth}
		\centering
		\includegraphics[width=\textwidth]{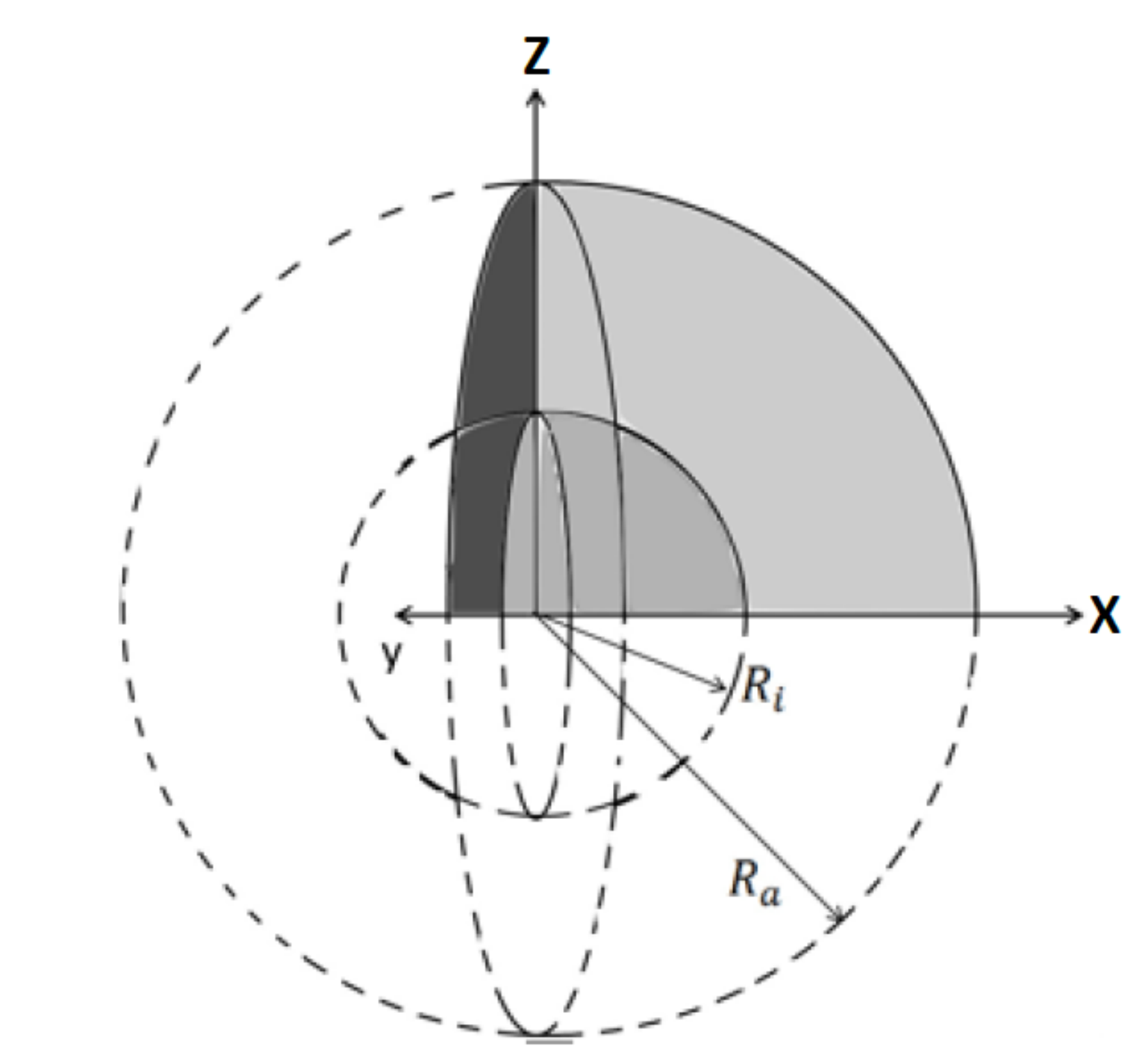}\hfill
		\caption{Hollow sphere under internal pressure.}
	\end{subfigure}\hspace{5pt}
	\begin{subfigure}[b] {0.4\textwidth}
		\centering
		\includegraphics[width=\textwidth]{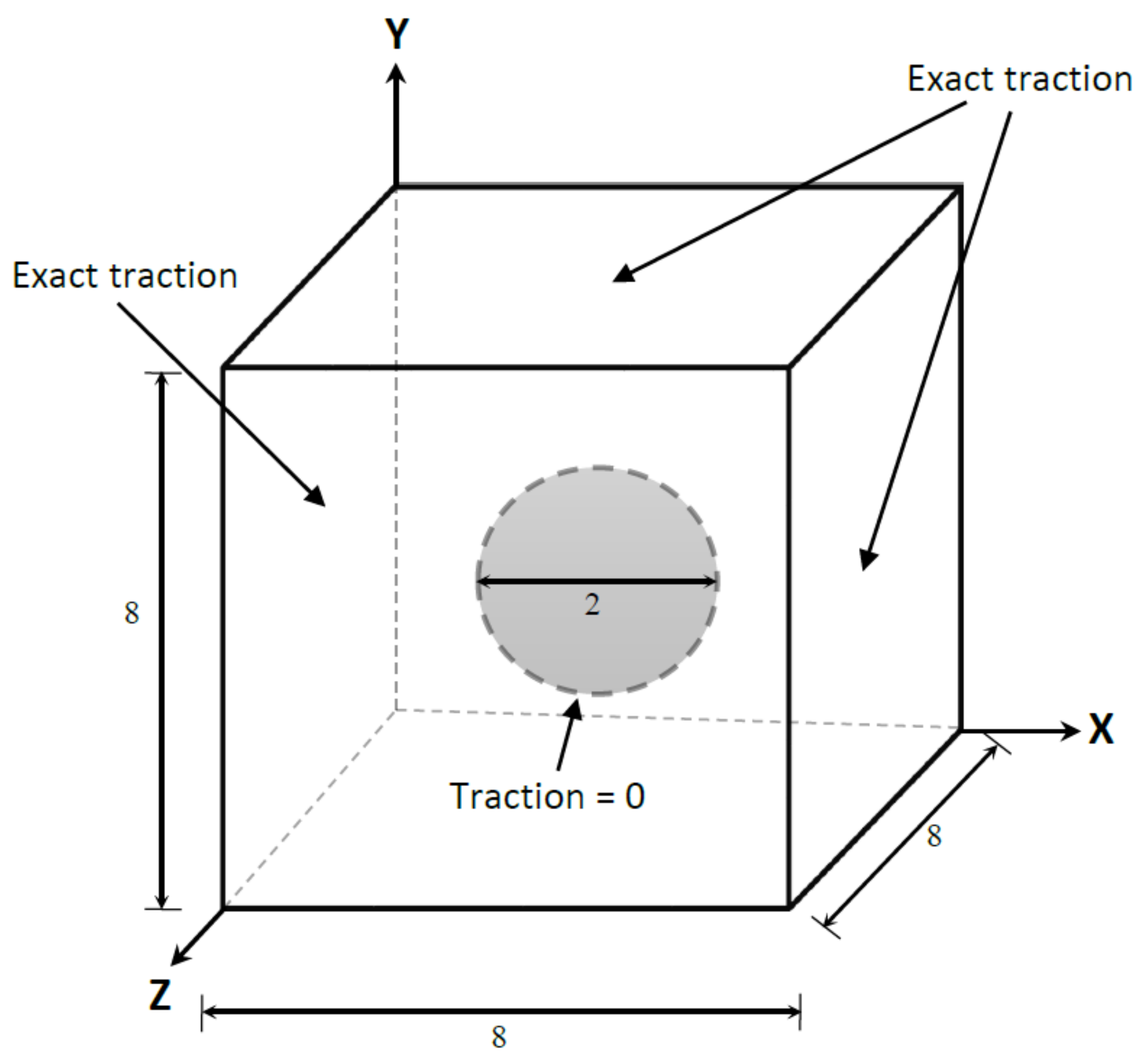}\hfill
		\caption{Cube with a spherical hole.}
	\end{subfigure}\hspace{5pt}
	\caption{Geometrical Setup and boundary condition for the application of linear elasticity on three-dimensional structure.}
	\label{fig:setup_hollowsphere}
\end{figure}

\begin{figure}
 \centering
 \begin{subfigure}[b] {0.4\textwidth}
   \centering
   \includegraphics[width=\textwidth]{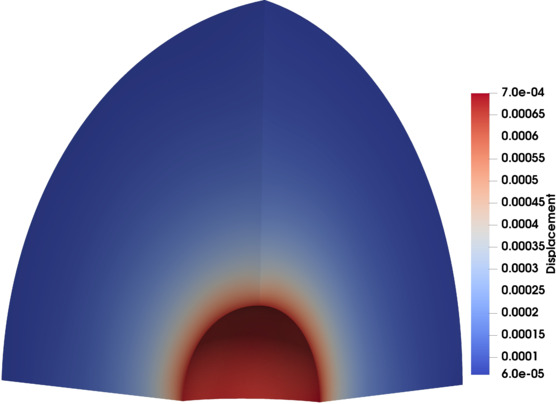}\hfill
   \caption{Displacement.}
 \end{subfigure}\hspace{5pt}
\begin{subfigure}[b] {0.4\textwidth}
   \centering
   \includegraphics[width=\textwidth]{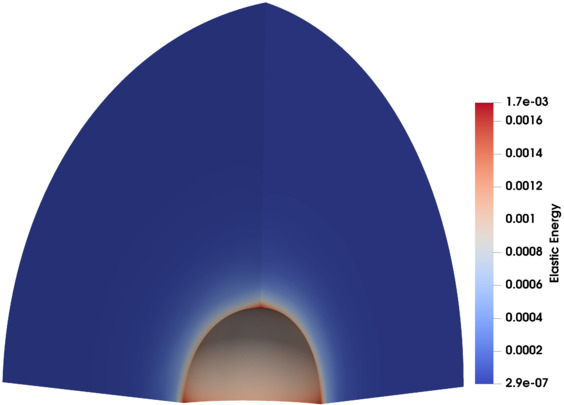}\hfill
   \caption{Strain energy.}
 \end{subfigure}\hspace{5pt}
 \caption{Computed solution for the hollow sphere subjected to internal pressure}
 \label{fig:disp_hollowsphere}
 \end{figure}

In this example, a hollow sphere is subjected to internal pressure is considered. Owing to the symmetrical structure, we analyze only one-eighth of the hollow sphere. The problem domain and its corresponding boundary conditions are shown in \autoref{fig:setup_hollowsphere}(a). The analytical solution of the problem is given by \cite{schillinger2015collocated}
\begin{equation}\label{eq:exact_hollowSphere}
    \begin{split}
        u_{r} &= \frac{PR_b^3r}{E(R_a^3-R_b^3)}\left(1-2\nu + \frac{(1+\nu)R_a^3}{2r^3}\right),\\
        \sigma_r &= \frac{PR_i^3(R_a^3-r^3)}{r^3(R_a^3-R_b^3)},\\
        \sigma_{\phi} &= \sigma_{\theta} = \frac{PR_i^3(R_a^3+2r^3)}{2r^3(R_a^3-R_b^3)},\\
    \end{split}
\end{equation}
where $r$ and $\theta$ denote the radial distance and the angle with respect to the origin to locate any co-ordinatewithin the domain and $\phi$ represents the azimuthal angular in the $\it{x}-\it{y}$ plane from the $\it{x}$-axis. In \autoref{eq:exact_hollowSphere}, $R_a$ and $R_b$ denote the outer and the inner radius of the hollow sphere, respectively, where $R_a = 4$ and $R_b = 1$. For this problem, the material properties considered are $E = 1\times10^3$ and $\nu = 0.3$. The hollow sphere is subjected to an internal pressure, $P = 1$. The Dirichlet boundary conditions are:
\begin{equation}
    u(x,y,0) = v(x,y,0) = w(x,y,0) = 0,  
\end{equation}
where $u$, $v$ and $w$ are the solutions of the elastic field in \textit{x}, \textit{y}, and \textit{z}-axis, respectively. To ensure that the boundary conditions are exactly satisfied, we have set
\begin{equation}\label{eq:trial_3d}
\begin{split}
    u &= x\hat{u},\\
    v &= y\hat{v},\\
    w & = z\hat{w},\\
\end{split}
\end{equation}
where $\hat{u}$, $\hat{v}$ and $\hat{w}$ are obtained from the neural network. To obtain the displacement field, we have used a network with 3 hidden layers of $50$ neurons each and $N_x\times N_y\times N_z$ uniformly spaced points in the interior of the domain, $N_x = N_y = 80$ and $N_z = 10$. The boundary face on which the internal pressure, in terms of traction force, is applied is discretized into $N_{Bound}$ points, with $N_{Bound} = 1600$. For the first two layers, we have considered rectified linear units, \texttt{ReLU}$^2$ activation function; whereas for the last layer, linear activation function has been considered. For a three-dimensional problem, the loss function is computed as 
\begin{equation}\label{eq:loss_elasticity_3D}
\begin{split}
    \mathcal{L}_{elastic} &= \mathcal{L}_{int} - \mathcal{L}_{neu},\\
    \mathcal{L}_{int} & = \frac{V_{\Omega}}{N_x\times N_y\times N_z}\sum_{i = 1}^{N_x\times N_y \times N_z}f(\bm{x}_i),\\
    \mathcal{L}_{neu} & = \frac{V_{\partial \Omega}}{N_{bound}}\sum_{i = 1}^{N_{Bound}}f_{neu}(\bm{x}_i),\\
    f(\bm{x}) &= \frac{1}{2}\bm{\epsilon}:\mathbb{C}:\bm{\epsilon},\\
    f_{neu}(\bm{x}) &= t_{N,x}(\bm{x}) + t_{N,y}(\bm{x}) + t_{N,z}(\bm{x}),\\
\end{split}
\end{equation}
where $\mathcal{L}_{int}$ and $\mathcal{L}_{neu}$ defines the losses in the elastic strain energy and the Neumann boundary loss, respectively and $V_{\Omega}$ is the volume of the analyzed domain. The predicted displacement field and the strain energy, using the DEM approach is shown in \autoref{fig:disp_hollowsphere}. 

Similar to the previous examples, the $\mathcal{L}_2$ error is computed to check the accuracy of the solution. Corresponding to $u$ and strain energy, a prediction error of 1.05\% and 4.44\%, respectively have been observed. 

\subsubsection{Cube with a spherical hole subject to uniform tension}
\label{subsec2:LE_prob4}
\begin{figure}
 \centering
 \begin{subfigure}[b] {0.4\textwidth}
   \centering
   \includegraphics[width=\textwidth]{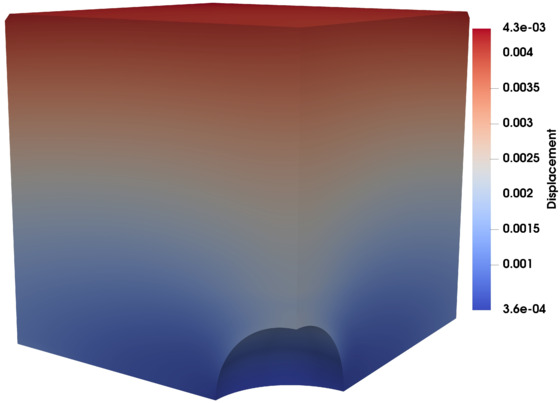}\hfill
   \caption{Displacement}
 \end{subfigure}\hspace{5pt}
\begin{subfigure}[b] {0.4\textwidth}
   \centering
   \includegraphics[width=\textwidth]{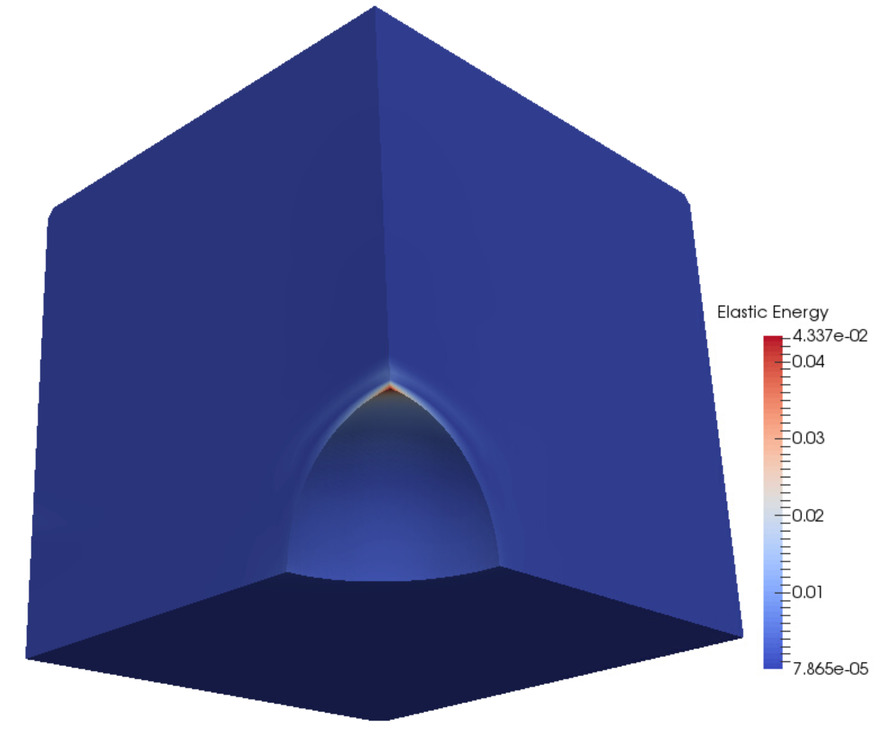}\hfill
   \caption{Strain energy}
 \end{subfigure}\hspace{5pt}
 \caption{Computed solution for the cube with a hole example}
 \label{fig:disp_cube}
 \end{figure}

As the last example of the linear elasticity section, we consider a spherical hole in a cube subjected to uniform tensile loading. Owing to symmetry, only one-eight of the spherical hole is analyzed as shown in \autoref{fig:setup_hollowsphere}(b). The analytical stresses in spherical coordinates for this are \cite{scott2013isogeometric}:
\begin{equation}
    \begin{split}
        \sigma_{rr} & = S\cos^2\theta +\frac{S}{7-5\nu}\left(\frac{a^3}{r^3}(6-5\cos^2\theta(5-\nu)) + \frac{6a^5}{r^5}(3\cos^2\theta -1)\right),\\
        \sigma_{\phi\phi} & = \frac{3S}{2(7-5\nu)}\left(\frac{a^3}{r^3}(5\nu -2+5\cos^2\theta(1-2\nu)) + \frac{a^5}{r^5}(1-5\cos^2\theta)\right),\\
        \sigma_{\theta\theta} & = S\sin^2\theta +\frac{S}{2(7-5\nu)}\left(\frac{a^3}{r^3}\left(4-5\nu +5\cos^2\theta\left(1-2\nu\right)\right) + \frac{3a^5}{r^5}(3-7\cos^2\theta)\right),\\
        \sigma_{r\theta} & = S\left(-1+\frac{1}{7-5\nu}\left(\frac{12a^5}{r^5}-\frac{5a^3(1+\nu)}{r^3}\right)\right),\\
    \end{split}
\end{equation}
where $S$ denotes the applied uniaxial tension and $a$ is the radius of the spherical hole. The material properties considered for this example are $E = 1\times 10^{3}$ and $\nu = 0.3$. The Dirichlet boundary conditions are:
\begin{equation}
    u(x,y,0) = v(x,y,0) = w(x,y,0) = 0,  
\end{equation}
where $u$, $v$ and $w$ are the solutions of the elastic field in \textit{x}, \textit{y}, and \textit{z}-axis, respectively. To ensure that the boundary conditions are exactly satisfied, we use the same trial function for the displacement field as stated in \autoref{eq:trial_3d}. For obtaining the displacement, we have used a fully connected neural network with 3 hidden layers of $50$ neurons each. Similar to previous example, for the first two layers \texttt{ReLU}$^2$ activation function and for the last layer linear activation function have been used. The domain is discretized into $N_{Int}$ points in the interior of the domain, with $N_{Int} = 32000$. The boundary face on which the internal pressure in terms of traction force is applied is discretized into $N_{Bound}$ points, with $N_{Bound} = 1600$. For a three-dimensional problem, the loss function is computed as 
\begin{equation}\label{eq:loss_elasticity_cube}
\begin{split}
    \mathcal{L}_{elastic} &= \mathcal{L}_{int} - \mathcal{L}_{neu},\\
    \mathcal{L}_{int} & = \frac{V_{\Omega}}{N_{Int}}\sum_{i = 1}^{N_{Int}}f(\bm{x}_i),\\
    \mathcal{L}_{neu} & = \frac{V_{\partial \Omega}}{N_{bound}}\sum_{i = 1}^{N_{Bound}}f_{neu}(\bm{x}_i),\\
    f(\bm{x}) &= \frac{1}{2}\bm{\epsilon}:\mathbb{C}:\bm{\epsilon},\\
    f_{neu}(\bm{x}) &= t_{N,x}(\bm{x}) + t_{N,y}(\bm{x}) + t_{N,z}(\bm{x}),\\
\end{split}
\end{equation}
where $\mathcal{L}_{int}$ and $\mathcal{L}_{neu}$ defines the losses in the elastic strain energy and the Neumann boundary loss, respectively and $V_{\Omega}$ is the volume of the analyzed domain. The predicted displacement field and the strain energy are shown in \autoref{fig:disp_cube}. 

Similar to the previous examples, the $\mathcal{L}_2$ error is computed for the strain energy to check thee accuracy of the solution, and a prediction error of 5.3\% is observed. In the next section, we would analyze the PDEs involved in solving problems of hyperelasticity.

\subsection{Elastodynamics}

To investigate the approximation properties of neural networks for time-dependent problems, we consider a wave propagation example. In one dimension, the governing (equilibrium) equation is of the form:
\begin{equation} \frac{\partial^2 u}{\partial t^2} = c^2 \frac{\partial^2 u}{\partial x^2} \mbox{ for } x \in \Omega:=(a,b) \mbox{ and } t\in(0,T) \end{equation}
together with the initial conditions
\begin{equation} u(x,0) = u_0(x)  \mbox{ and } u_t(x) = v_0(x) \mbox{ for } x\in\Omega \end{equation}
and boundary conditions 
\begin{align} u(x,t) &= \bar{u}(t) \mbox{ for } x\in\Gamma_D \mbox { and } \\
			  u_x(x,t) &= g(t) \mbox{ for } x\in\Gamma_N. \end{align}			  
Here $u(x,t):\Omega \times [0,T] \rightarrow \mathbb{R}$ is the displacement at point $x$ and time $t$, $c$ is a real positive constant representing the wave propagation speed, $u_0$ is the initial displacement, $v_0$ is the initial velocity, and $\bar{u}(t)$ and $g(t)$ are the prescribed data for the Dirichlet and Neumann boundaries $\Gamma_D$ and respectively $\Gamma_N$. 

In this example, we assume that $c=1$ and the left side of the domain is subject to a sinusoidal load while the right side is fixed:
\begin{equation} u_x(0,t) = \begin{cases} -\sin(\pi t), \mbox{ for } 0\leq t \leq 1 \\ 0 \mbox{ \quad otherwise } \end{cases} \mbox{ and } u(L,t) = 0, \end{equation}
where $L=4$. The initial displacement and velocity are set to zero:
\begin{equation} u_0(x)=v_0(x)=0. \end{equation}
\begin{figure}
	\centering
	\begin{subfigure}[b] {0.45\textwidth}
		\centering
		\includegraphics[width=\textwidth]{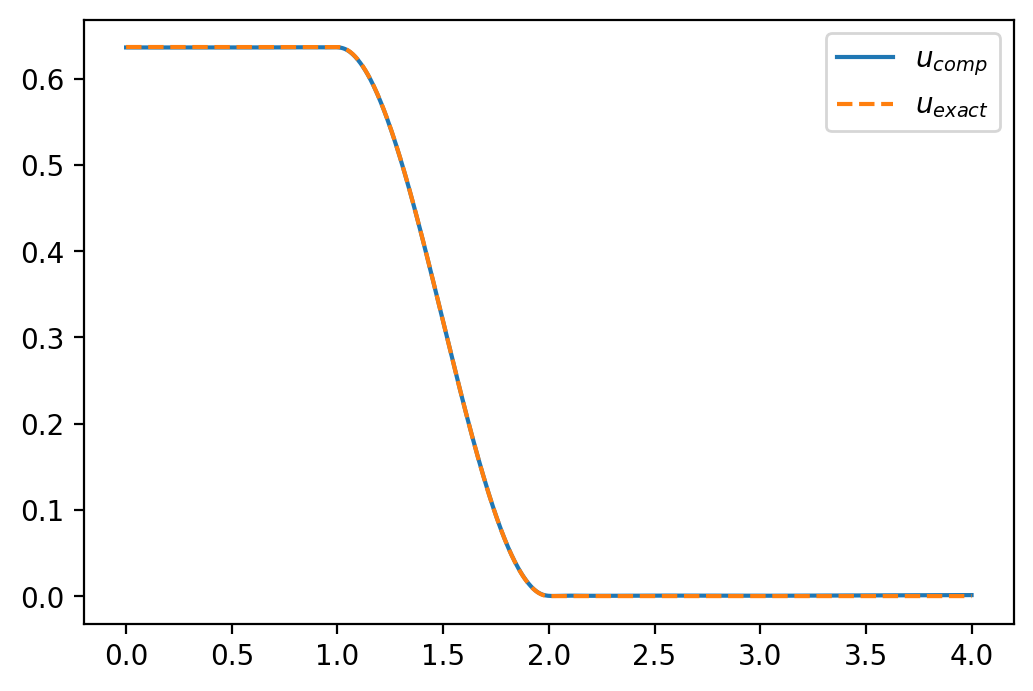}\hfill
		\caption{Computed and exact displacement}
	\end{subfigure}\hspace{5pt}
	\begin{subfigure}[b] {0.45\textwidth}
		\centering
		\includegraphics[width=\textwidth]{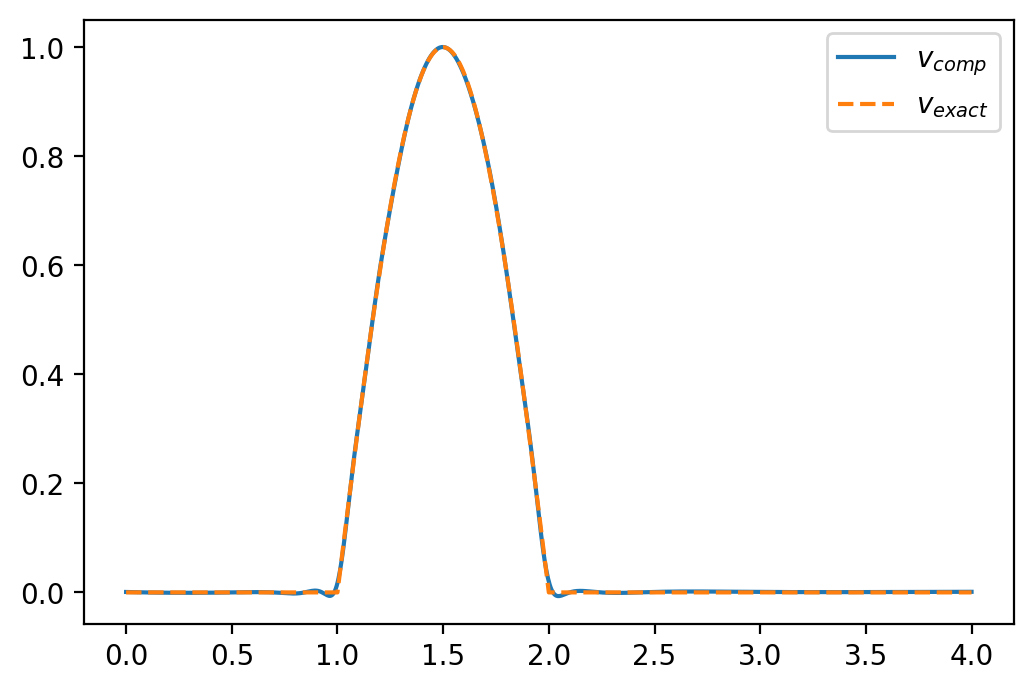}\hfill
		\caption{Computed and exact velocity}
	\end{subfigure}	
	\begin{subfigure}[b] {0.45\textwidth}
		\centering
		\includegraphics[width=\textwidth]{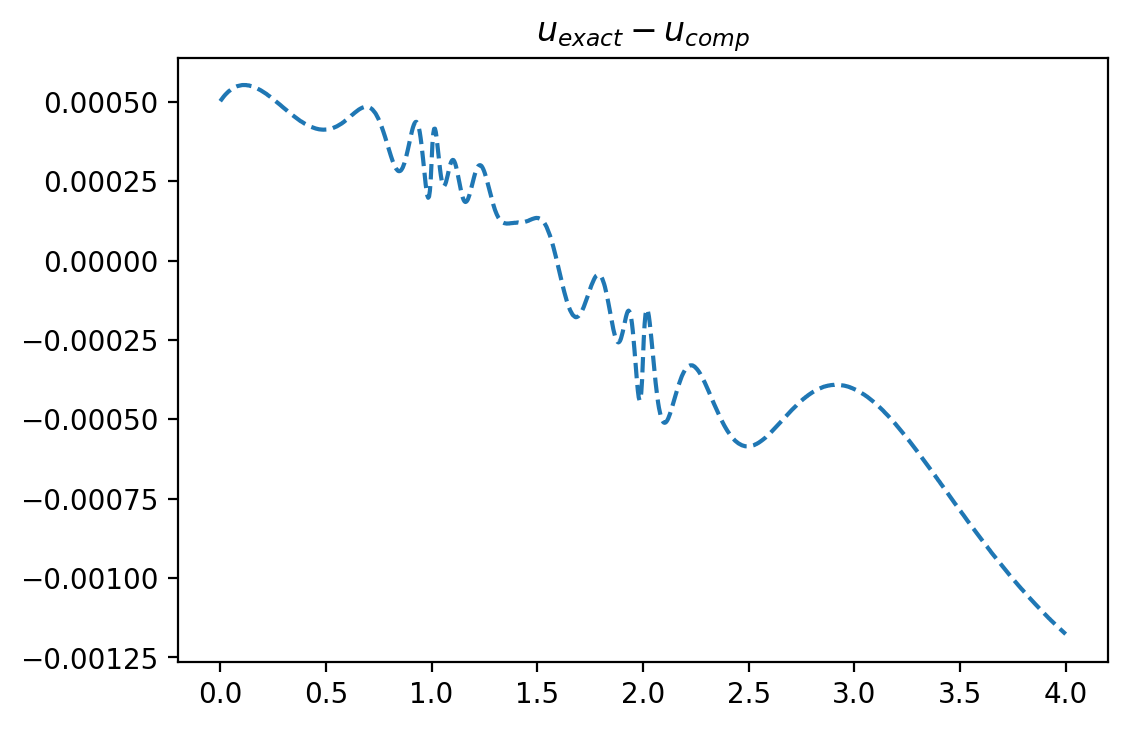}\hfill
		\caption{Error for displacement}
	\end{subfigure}\hspace{5pt}
	\begin{subfigure}[b] {0.45\textwidth}
		\centering
		\includegraphics[width=\textwidth]{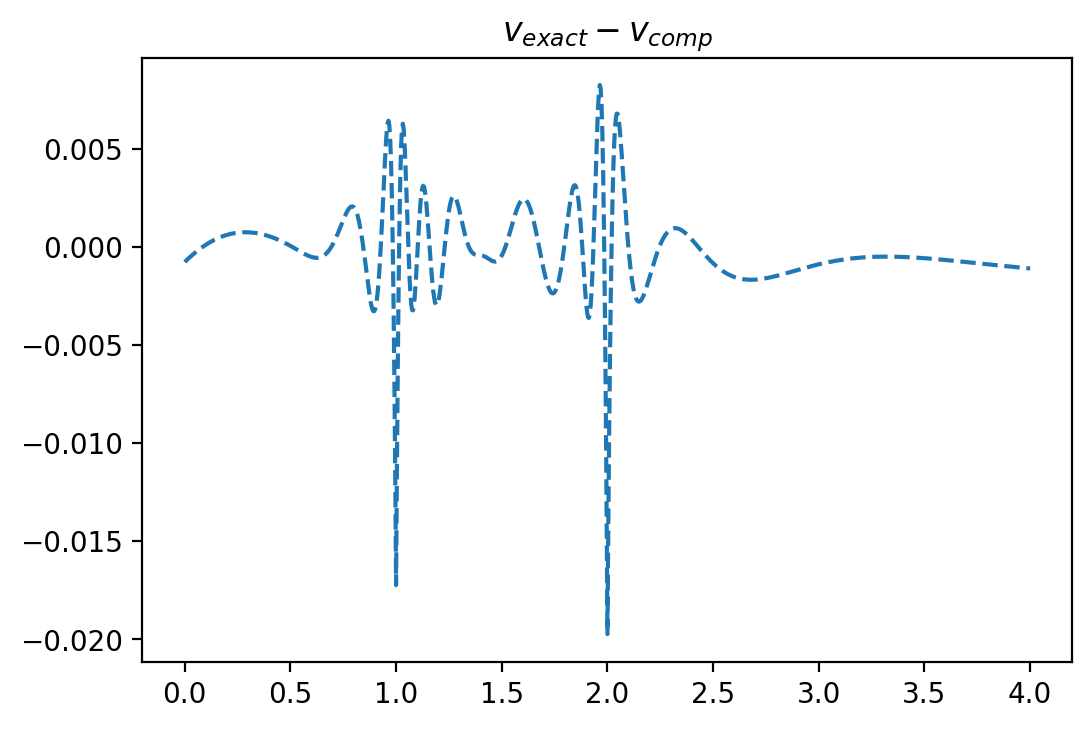}\hfill
		\caption{Error for velocity}
	\end{subfigure}
	\caption{Comparison between the computed and exact solutions for the 1D wave propagation example}
	\label{fig:1dwave}
\end{figure}
We compute the displacement up to time $T=2$ with a space-time discretization neural network with 3 hidden layers of 50 neurons each. The input layer has 2 neurons (for the space and time coordinates), while the output layer has 1 neuron for outputting the displacement. The loss function is defined based on a collocation approach which incorporates terms corresponding to the equilibrium equation together with the initial and boundary conditions as:
\begin{equation}
\mathcal{L}(\boldsymbol{p}) = \mathcal{L}_{resid}(\boldsymbol{p}) + \mathcal{L}_{init}(\boldsymbol{p}) + \mathcal{L}_{bnd}(\boldsymbol{p}), \end{equation}
where 
\begin{align}
&\mathcal{L}_{resid}(\boldsymbol{p}) = \frac{1}{N_{int}}\sum_{i=1}^{N_{int}}(\hat{u}_{tt}(x_i^{int},t_i^{int};\boldsymbol{p})-c^2\hat{u}_{xx}(x_i^{int},t_i^{int};\boldsymbol{p}))^2 \\
&\mathcal{L}_{init}(\boldsymbol{p}) = \frac{1}{N_{init}}\sum_{j=1}^{N_{init}}\left[(\hat{u}(x_j^{init},0;\boldsymbol{p})-u_0(x_j^{init}))^2+(\hat{u}_t(x_j^{init},0;\boldsymbol{p})-v_0(x_j^{init}))^2\right] \\
&\mathcal{L}_{bnd}(\boldsymbol{p})  =\frac{1}{N_{bnd}}\sum_{k=1}^{N_{bnd}}\left[(\hat{u}_x(0,t_k^{bnd};\boldsymbol{p})-\bar{u}(t_k^{bnd}))^2+(\hat{u}(L,t_k^{bnd};\boldsymbol{p})-g(t_k^{bnd}))^2\right].
\end{align}
Here $\hat{u}(x,t;\boldsymbol{p})$ is the neural network approximation to $u(x,t)$ which is determined by the parameters $\boldsymbol{p}$ obtained by minimizing the loss function. Moreover, $(x_i^{int}, t_i^{int})$ for $i=1, \ldots, N_{int}$ are interior collocation points, $x_j^{init}$ with $j=1,\ldots,N_{init}$ are the points in the space domain and $t_k^{bnd}$, $k=1,\ldots,N_{bnd}$ are points in time corresponding to the initial and boundary conditions respectively.

The approximation obtained with $N_{int}=199^2$,and $N_{init}=199$, and $N_{bnd}=200$ collocation points as well as the errors in the computed displacement and velocity are shown in Figure \ref{fig:1dwave}. The relative errors in $L^2$ norm for the displacement and velocity are $1.422382\cdot 10^{-3}$ and $5.954601\cdot 10^{-3}$ respectively. 

\subsection{Hyperelasticity}
\begin{figure}
	\centering
	\def\svgwidth{0.8\textwidth}
	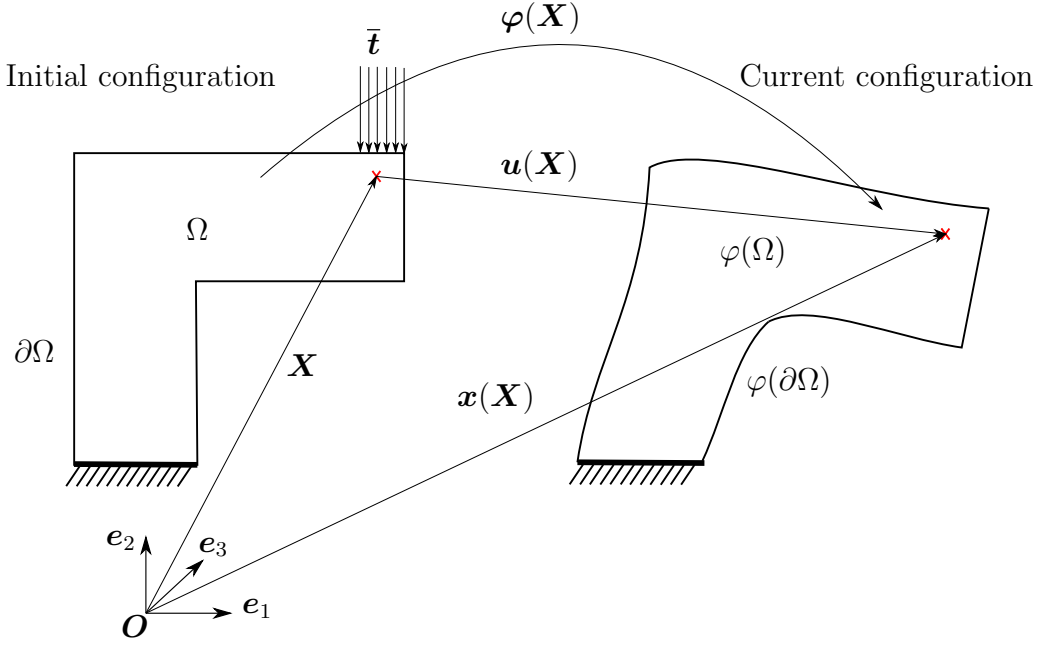
	\caption{Motion of body \textit{B}.}
	\label{pic:equi}
\end{figure}

In the context of the elastostatics at finite deformation (shown in Figure \ref{pic:equi}), the equilibrium equation along with the boundary conditions, for an initial configuration, is given as:

\begin{align}
& \text{Equilibrium:} \quad \nabla \cdot \tsr{P} + \tsr{f}_b = \mathbf{0}, \\
& \text{Dirichlet boundary}: \tsr{u} = \overline{\tsr{u}} \quad \text{on} \,\, \partial {\Omega}_{D} ,  \\
& \text{Neumann boundary}: \tsr{P} \cdot n = \overline{\tsr{t}} \quad \text{on} \,\, \partial {\Omega}_{N}, 
\end{align}
in which $\overline{\tsr{u}}$ and $\overline{\tsr{t}}$ are prescribed values on Dirichlet boundary and Neumann boundary, respectively. Therein, the boundaries have to fulfill $\partial \Omega_{D} \cup \partial \Omega_{N} = \partial \Omega$, $\partial \Omega_{D} \cap \partial \Omega_{N} = \emptyset$. The outward normal unit vector is denoted by $n$. The $1^{st}$ Piola-Kirchhoff stress tensor $\tsr{P}$ is related to its power conjugate $\tsr{F}$, so-called deformation gradient tensor, by a constitutive equation $\tsr{P} = {\partial \Psi} / {\partial \tsr{F}}$. The deformation gradient is defined as follows
\begin{equation}
\tsr{F} = \text{Grad }\tsr{\varphi}(\tsr{X}),
\end{equation}
where $\tsr{\varphi}$ denotes the mapping of material points in the initial configuration to the current configuration. It is defined as: \[\tsr{\varphi}(\tsr{X}) := \tsr{x} = \tsr{X} + \tsr{u}.
\]
Since the strain energy density $\Psi$ describing the elastic energy stored in the body is explicitly known, it is more convenient to find the possible deformations in a form of potential energy rather than the strong formulation. The potential functional is written as
\begin{equation} \label{eq: potential1}
\mathcal{E}(\tsr{\varphi}) = \int_{\Omega} \Psi	\,dV - \int_{\Omega} \tsr{f}_b \cdot \tsr{\varphi}\,dV - \int_{\partial \Omega_{N}} \overline{\tsr{t}} \cdot \tsr{\varphi} \, dA.
\end{equation}
In order to obtain the solution we minimize the potential energy
\begin{equation}
\min\limits_{\tsr{\varphi} \in H} \, \, \mathcal{E}(\tsr{\varphi}),
\end{equation}
where $H$ is the set of admissible functions (trial functions). The method is named \textit{the principle of minimum total potential energy}.

Now that our objective is to minimize the potential energy with the aid of neural networks, we need to cast the minimization into the optimization problem in the context of machine learning. Therefore, the definition of a loss function is essential. In this case we exploit the potential energy as the loss function. It reads
\begin{equation}
\mathcal{L}(\boldsymbol{p}) = \int_{\Omega} \Psi ({\tsr{\varphi}}(\tsr{X}; \boldsymbol{p}))	\,dV - \int_{\Omega} \tsr{f}_b \cdot {\tsr{\varphi}}(\tsr{X}; \boldsymbol{p})\,dV - \int_{\partial \Omega_{N}} \overline{\tsr{t}} \cdot \tsr{\varphi}(\tsr{X}; \boldsymbol{p}) \, dA,
\end{equation}
where the trial function now reads 
\begin{equation}
	{\tsr{\varphi}}_p(\tsr{X}) = {\tsr{u}}_p(\tsr{X}) + \tsr{X},
\end{equation}
and ${\tsr{u}}_p(\tsr{X})$ has to fulfill boundary conditions prior to being applied any operators. We use a short notation $f_p(\tsr{X})$ instead of $f(\tsr{X}; \tsr{p})$. Here, a feedforward neural network constructed by hyperparameters $\boldsymbol{p}$ correspond to the weights and biases of the neural architecture is employed to establish the trial solution.
In this context the loss function now is written by means of an approximation as follows
\begin{equation}
\mathcal{L}(\boldsymbol{p}) \approx  \frac{V_\Omega}{N_\Omega} \sum_{i=1}^{N_\Omega} {\Psi}((\tsr{\varphi}_p)_i) - \frac{V_\Omega}{N_\Omega} \sum_{i=1}^{N_\Omega} (\mathbf{f}_b)_i \cdot (\ensuremath{{\boldsymbol{\varphi}}}_p)_i - \frac{A_{\partial \Omega_N}}{N_{\partial \Omega_N}} \sum_{i=1}^{N_{\partial \Omega_N}} \overline{\mathbf{t}}_i \cdot (\tsr{\varphi}_p)_i,
\end{equation} 
where $N_\Omega$ is the total number of data of the solid; $N_{\partial \Omega_N}$ is the number of data of the surface having force. $V_\Omega$ is the volume of the solid and $A_{\partial \Omega_{N}}$ is the area of the surface. If a function $f$ is evaluated at a material point $\mathbf{X}_i$, we denote this by $f_i$.

\begin{figure}
	\centering
	\def\svgwidth{0.7\textwidth}
	\hspace{30mm}
	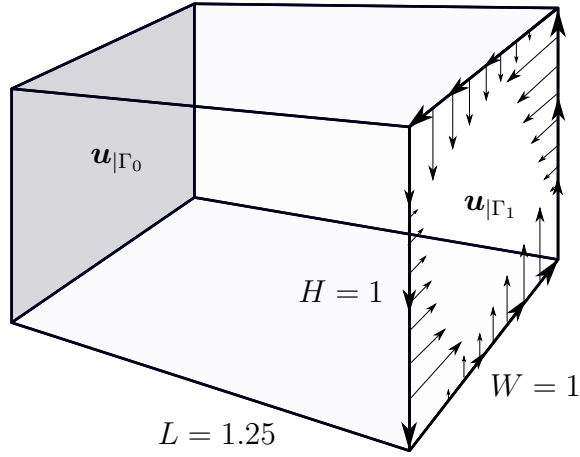
	\caption{{A hyperelastic 3D Cuboid is twisted an angle of $60^o$.}}
	\label{pic:TwistingCuboid}
\end{figure}

An example about torsion inspired from a Fenics application \cite{LoggMardalEtAl2012} is chosen to examine in order to verify the robustness of DEM in hyperelasticity problems.  Let us consider a 3D Cuboid made of an isotropic, homogeneous, hyperelastic material with length $L = 1.25$, width $W=1.0$ and depth $L=1.0$ (we drop out all units for the sake of simplicity). The solid is clamped at the left surface and is twisted an angle of $60^o$ counterclockwise by prescribed displacement boundary conditions $\tsr{u}_{|\Gamma_1} $ applied at the right-end surface. In addition, the Cuboid is subjected to a body force $\tsr{f}_b = [0, -0.5, 0]^T$ and traction forces $\overline{\tsr{t}} = [1, 0, 0]^T$ at the bottom, back, top and front surfaces (Figure \ref{pic:TwistingCuboid}).
The Dirichlet boundary conditions are prescribed as follows
\begin{align} \label{eq:DirBCCuboic}
& \tsr{u}_{|\Gamma_0} = \left[0, 0, 0\right]^T, \notag \\
& \tsr{u}_{|\Gamma_1} = \begin{bmatrix}
0 \\
0.5 \left[0.5 + (X_2 - 0.5)\cos({\pi}/{3}) - (X_3-0.5)\sin({\pi}/{3}) - X_2\right] \\
0.5 \left[0.5 + (X_2 - 0.5)\sin({\pi}/{3}) + (X_3-0.5)\cos({\pi}/{3}) - X_3\right] 
\end{bmatrix}. 
\end{align}
In this problem, the Neo-Hookean model is considered and it has the form
\begin{equation}
\Psi(I_1, J) = \frac{1}{2} \lambda [\log(J)]^2 - \mu \log(J) + \frac{1}{2} \mu (I_1-3),
\end{equation}
where two invariants are given by 
\begin{equation}
I_1 = \text{trace}(\tsr{C}), \quad  \quad J = {\det(\tsr{F})}.
\end{equation}
The right Cauchy-Green tensor is expressed as $\tsr{C} = \tsr{F}^T \cdot \tsr{F}$. In this example, material properties are prescribed in Table \ref{table:cuboid},
\begin{table}
	\begin{center}
		\scalebox{1.0}{
			\def\arraystretch{1.65}
			\begin{tabular}{|p{4.5cm}||p{2cm}|}
				\hline
				Description & Value \\
				\hline
				$E$ - Young modulus   & $10^6$   \\
				$\nu$ - Poisson ratio &   0.3  \\
				$\mu$ - Lame' parameter  & $\frac{E}{2(1+\nu)}$ \\
				$\lambda$ - Lame' parameter & $\frac{E\nu}{(1+\nu)(1-2\nu)}$  \\
				\hline
		\end{tabular}}
	\end{center}
	\caption{Material parameters for 3D hyperelastic Cuboid.}
	\label{table:cuboid}
\end{table} \\
With the prescribed boundary conditions, the displacement has the form
\begin{equation}
\tsr{{u}}_p(\tsr{X}) = \mathbf{u}_{|\Gamma_1}\, \frac{X_1}{1.25} + (X_1 - 1.25)\, X_1 \, \wh{\tsr{u}}(\mathbf{X}; \boldsymbol{p}).
\end{equation}

\begin{figure}[H]
	\centering
	\includegraphics[width=1\textwidth]{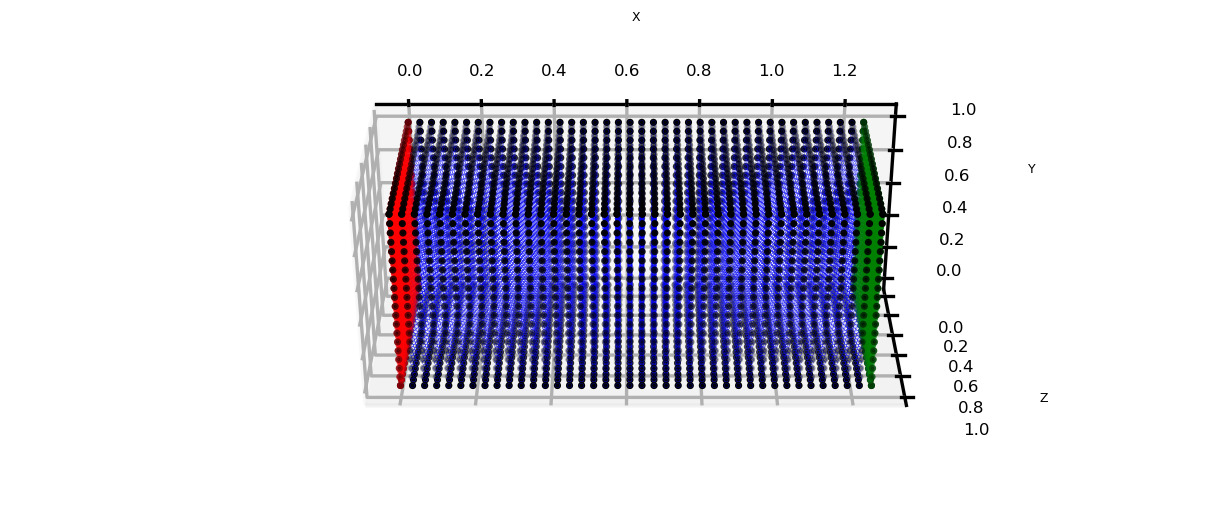}
	\caption{The training point distribution of a 3D hyperelastic cuboid in DEM. The red points correspond to the first Dirichlet boundary condition.  The points data are used to enforce/train the second Dirichlet boundary condition. The black points are used to train the surface force. The blue points are used for the interior domain integration and the body force.}
	\label{fig:dataCuboid}
\end{figure}

\paragraph{\textbf{Setup}}We generate 64000 equidistant grid points ($N_1 = 40, N_2 = 40, N_3 = 40$) over the entire domain as the feeding data to train our network (Figure \ref{fig:dataCuboid}). A 5-layer network ($3-30-30-30-3$) is constructed. Therein, we use 3 neurons in the input layer for the nodal coordinates and 3 neurons in the output layer for the unconstrained solution. Moreover, in each hidden layer 30 neurons are used to enforce the learning behavior. We use \textit{tanh} activation function to evaluate neural values in the hidden layers. The learning rate $r = 0.5$ is used in the L-BFGS optimizer. The neural network is strained for 50 steps.

\begin{figure}[H]
	\begin{subfigure}[b]{.5\linewidth}
		\centering\large
		\includegraphics[width=60mm]{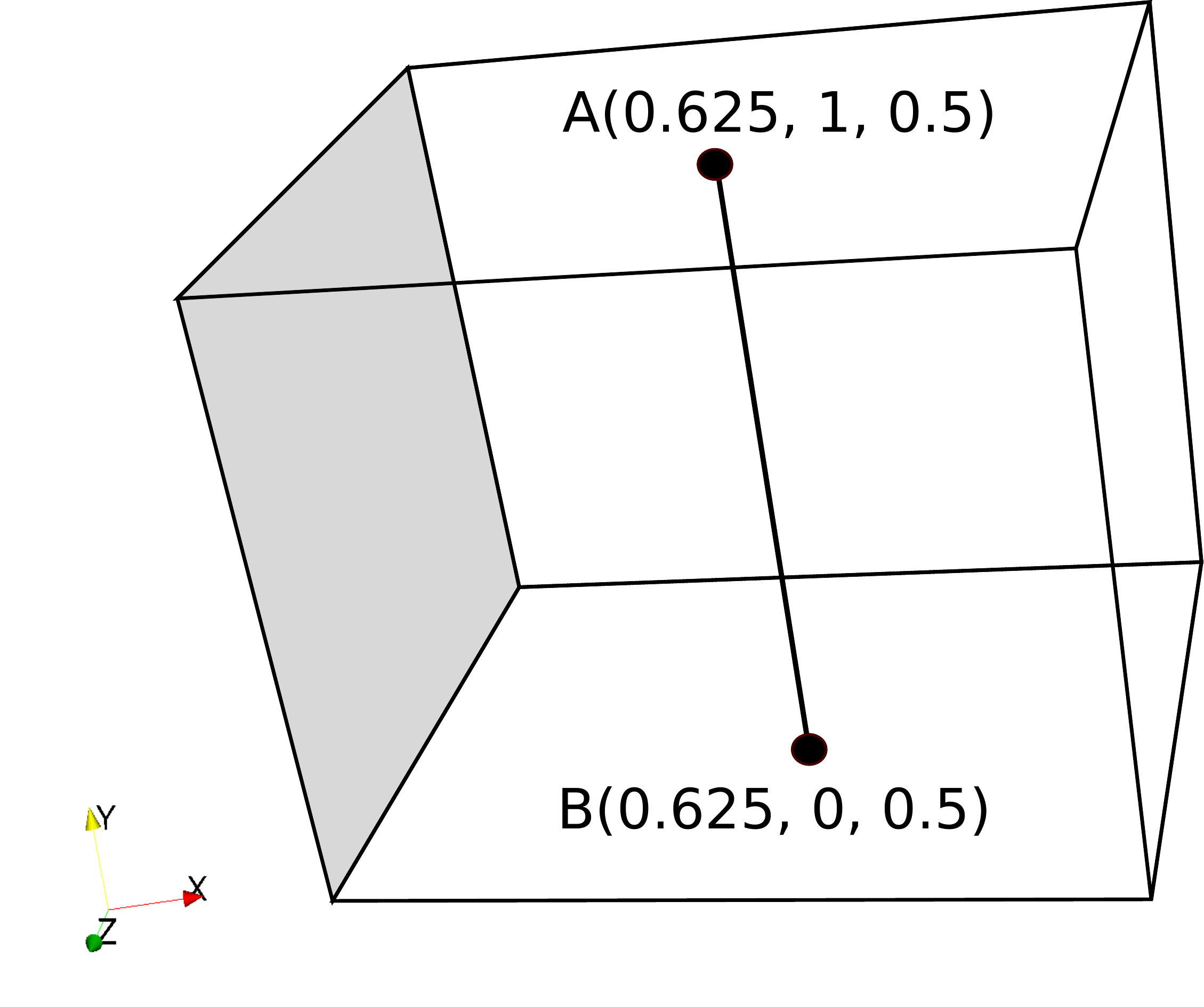}
		\centering\caption{AB line}
		\label{fig:ABline}
	\end{subfigure}
	\begin{subfigure}[b]{.5\linewidth}
		\centering\large
		\includegraphics[width=60mm]{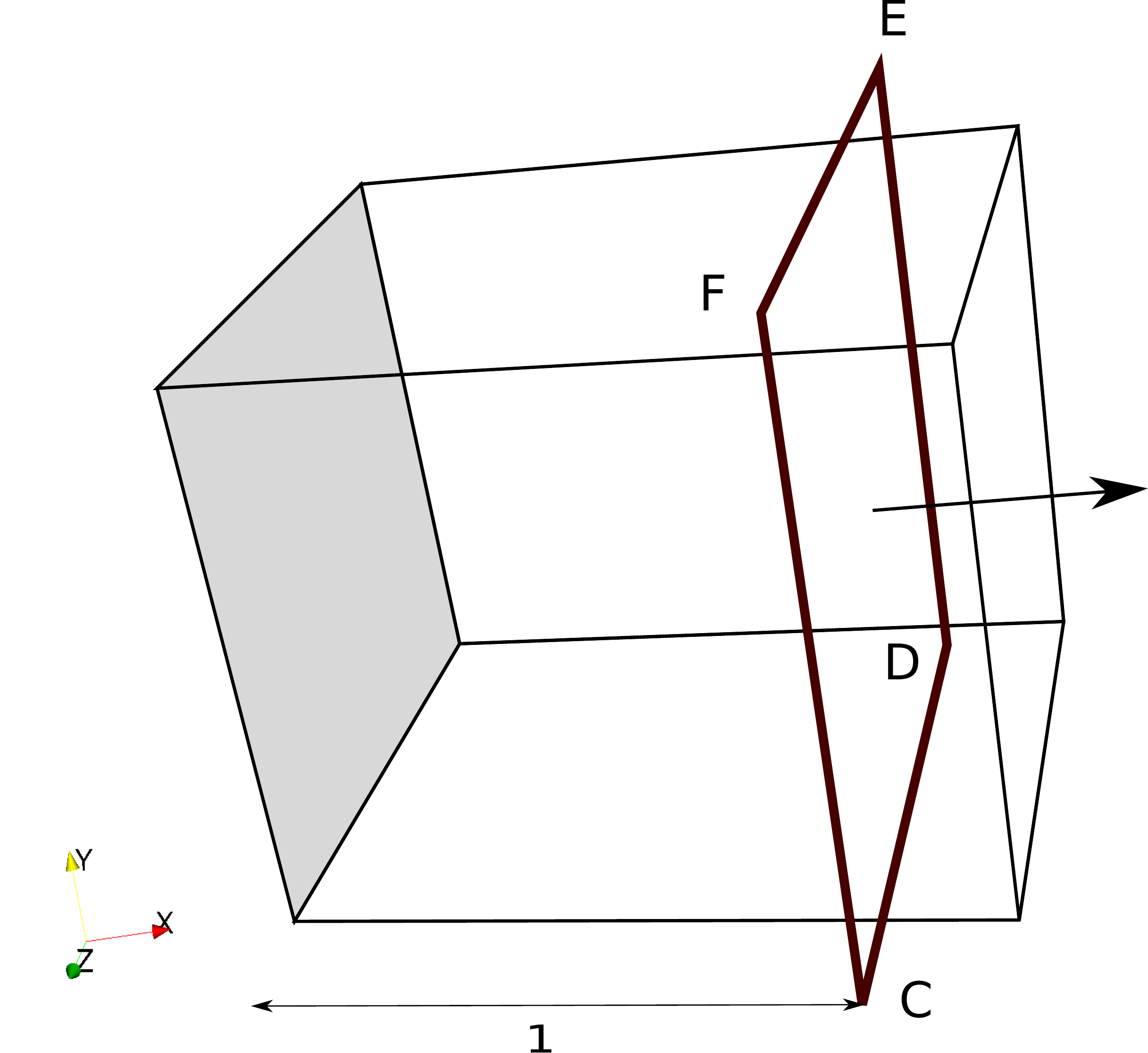}
		\centering\caption{CDEF plane}
		\label{fig:CDEFplane}
	\end{subfigure}\\
	\caption{Positions of AB line and CDEF plane.}
\end{figure}

\begin{figure}[H]
	\begin{subfigure}[b]{.5\linewidth}
		\centering\large
		\includegraphics[width=90mm]{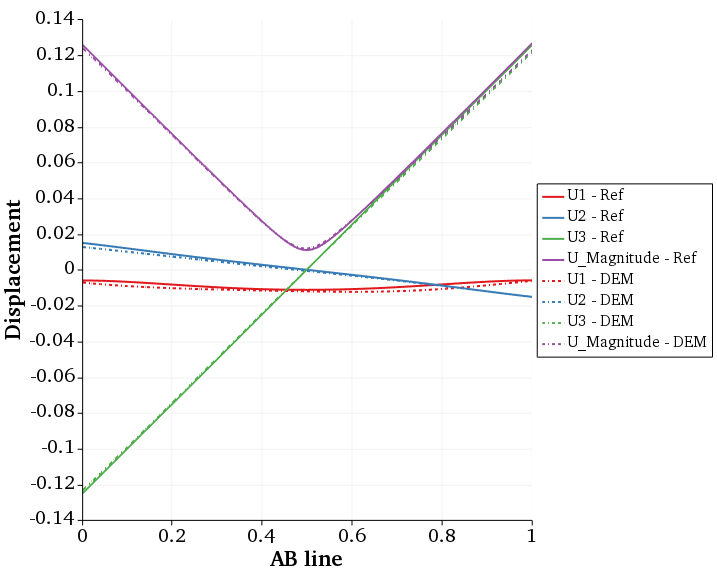}
		\centering\caption{Displacement comparison}
		\label{fig:displacemntAB}
	\end{subfigure}
	\begin{subfigure}[b]{.5\linewidth}
		\centering\large
		\includegraphics[width=90mm]{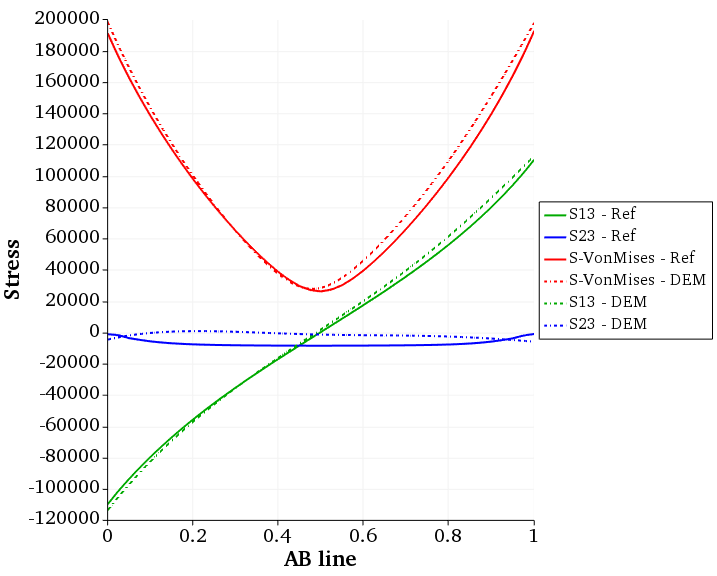}
		\centering\caption{Stress comparison}
		\label{fig:stressAB}
	\end{subfigure}\\
	\caption{The displacement and stress results of DEM measured at the AB line compared to the reference solution.}
	\label{fig:displacement_stress_AB}
\end{figure}

\begin{figure}[H]
	\begin{subfigure}[b]{.5\linewidth}
		\centering\large
		\includegraphics[width=85mm]{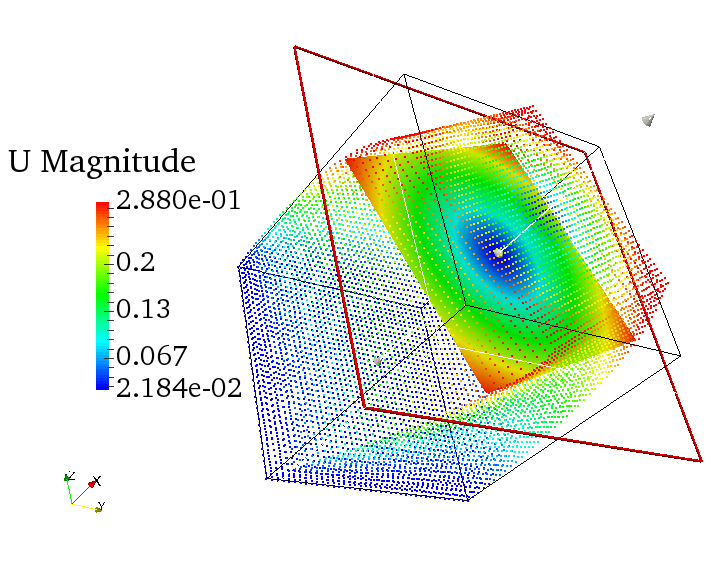}
		\caption{Displacement}
	\end{subfigure}
	\begin{subfigure}[b]{.5\linewidth}
		\centering\large
		\includegraphics[width=85mm]{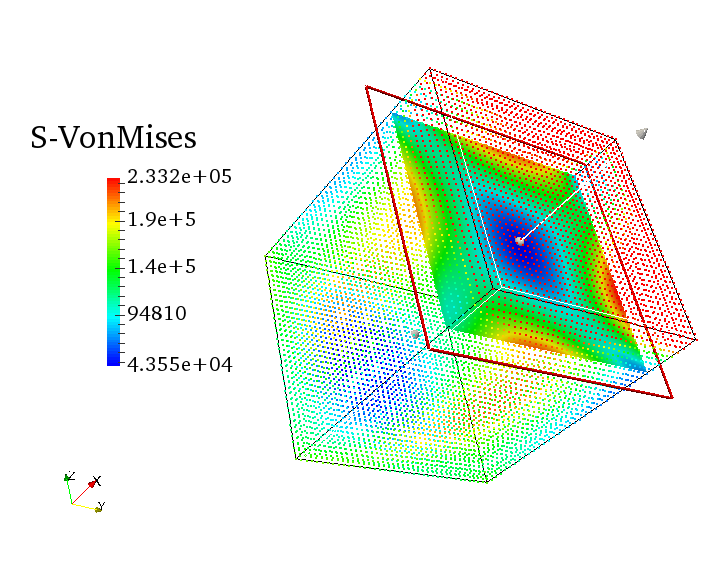}
		\caption{VonMises stress}
	\end{subfigure}
	\caption{Displacement and VonMises stress at the CDEF plane in DEM of a twisted Neo-Hookean 3D Cuboid.}
	\label{fig:Stress_CDEFplane}
\end{figure}

\paragraph{\textbf{Result}} 	
Since the analytical solutions are not available in this example, we use a finite element program to simulate the Cuboid with a fine mesh to obtain the reference solution. We first measure the displacements and stresses along the AB line depicted in the Figure \ref{fig:ABline}. The result in general reveals a good agreement between DEM solution and the reference solution in Figure \ref{fig:displacement_stress_AB}. It is found that the deep energy method exhibits good solutions in terms of displacements. However, there has small difference in terms of stress because the outputs of our neural network are the displacements and the gradient fields are then given in the post-processing procedure based on the displacement field.\\
We observe the surface displacement and surface stress at the CDEF plane as sketched in Figure \ref{fig:CDEFplane}. Figure \ref{fig:Stress_CDEFplane} shows the predicted displacement in terms of magnitude and the predicted stress in terms of VonMises. Then we compare the norms of the DEM solution and reference solution in terms of $L^2$ norm and $H^1$ seminorm in two setups. Setup 1: We use 8000 equidistant data points ($N_1 = 20, N_2 = 20, N_3 = 20$) as the feeding data to train our network; the network is trained for 50 steps. Setup 2: We use 64000 equidistant data points ($N_1 = 40, N_2 = 40, N_3 = 40$) as the feeding data to train our network; the neural network is trained for 25 steps. The $L^2$ norm and $H^1$ seminorm of the reference solution are given by a Fenics program with a fine mesh as: $||u||_{L^2}^{FE} = 0.13275$ and $||u||_{H^1}^{FE} = 0.51407$. As shown in Table \ref{table:table2}, the DEM obtain good results in terms of $L^2$ norm and $H^1$ seminorm. 

\begin{table}[H]
	\begin{center}
		\scalebox{0.9}{
			\def\arraystretch{1.0}
			\begin{tabular}{|l|c|c|}
				\hline
				\multicolumn{1}{|c|}{\multirow{2}{*}{\textbf{Data}}} & \multicolumn{2}{c|}{\textbf{DEMM}}                \\
				\cline{2-3} 
				\multicolumn{1}{|c|}{}                               & \multicolumn{1}{c|}{$||u||_{L^2}$} & \multicolumn{1}{c|}{$||u||_{H^1}$} \\
				\hline
				20x20x20 (50 steps)                                  & 0.12921                 & 0.49929                 \\
				\hline
				40x40x40 (25 steps)                                  & 0.13210                 & 0.51001                 \\
				\hline
			\end{tabular}
		}
	\end{center}
	\caption{The $L^2$ norm and $H^1$ seminorm of DEM in 2 setups. The corresponding norms of the reference solution are given by Fenics program with a fine mesh as: $||u||_{L^2}^{FE}=0.13275$, $||u||_{H^1}^{FE}=0.51407$.}
	\label{table:table2}
\end{table}

For the next three sections, we apply DEM on problems involving more than one field. We start with the modeling of fracture using phase field approach.

\subsection{Phase field modeling of fracture}
\label{subsec:fracture}
The prevention of fracture-induced failure is a major constraint in engineering designs. The phase field model for fracture is an effective way to model fracture by assuming the process zone has a finite width which is controlled by a length scale parameter ($l_0$). A sharp crack topology is recovered in the limit as $l_0 \to 0$ \cite{Bourdin2000}. In this approach, the effects associated with crack formation such as stress release are incorporated into the constitutive model. A continuous scalar parameter ($\phi$) is used to track the fracture pattern. The cracked region is represented by $\phi = 1$ while the undamaged portion is given by $\phi = 0$. The phase field approach aims to simultaneously solve for the displacement field and fracture region by minimizing the total potential energy of system, as postulated by the Griffith theory for brittle fracture \cite{A.Griffith1921}.

In this section, physics informed neural networks are developed to solve the governing partial differential equations for fracture analysis using DEM. The neural networks are trained to approximate the displacement field and the damage fields which satisfy the governing equations used to describe the physical phenomena. Modeling fracture using the phase field method involves the solving for the vector-valued elastic field, $\bm{u}$ and the scalar-valued phase field, $\phi$. In DEM, the solution is obtained by minimization of the total energy of the system, $\mathcal{E}$ \cite{Borden2014}. The problem statement can be written as:
\begin{equation}\label{eq:totalenergy}
\begin{split}
    \text{Minimize:}\;\;\;\; \mathcal{E} &= \mathcal{E}_e + \mathcal{E}_c,\\
    \text{subject to:}\;\;\;\; \bm{u} &= \bm{\overline u} \text{ on } \partial \Omega_{D},
    \end{split}
\end{equation}
where $\mathcal{E}_e$ is the stored elastic strain energy, $\mathcal{E}_c$ is the fracture energy and $\bm{\overline u}$ is the prescribed displacement on the Dirichlet boundary, $\partial \Omega_{D}$. In \autoref{eq:totalenergy}, $\mathcal{E}_e$ and $\mathcal{E}_c$ are defined as:
\begin{equation}\label{eq:energyterms}
    \begin{split}
       \mathcal{E}_e &=  \int_{\Omega}g(\phi) \Psi_{0}(\bm{\epsilon}) d\Omega,\\
       \mathcal{E}_c &=  \frac{G_c}{2l_0} \int\limits_\Omega{\left(\phi^2 +l_0^2 |\nabla\phi|^2 \right)}d\Omega + \int_{\Omega}g(\phi)H(\bm{x},t)d\Omega,
    \end{split}
\end{equation}
where $g(\phi)$ represents the monotonically decreasing stress-degradation function, $G_c$ is the critical energy release rate, $\Psi_0(\bm{\epsilon},\phi)$ is the initial strain energy density functional expressed in terms of linearised strain tensor, $\bm{\epsilon(u)}$. A common form of the degradation function, as used in literature, for isotropic solids is \cite{Miehe2010}:
\begin{equation}\label{eq:degradation_func}
    g(\phi) = (1-\phi)^2.
\end{equation}
$H(\bm{x},t)$ contains the maximum positive tensile energy ($\Psi^{+}_{0}$) in the history of deformation of the system and is defined as:
\begin{equation}\label{eq:history_field}
    H(\bm{x},t) = {\max_{s \in [0,t]}}\Psi^{+}_{0}(\bm{\epsilon}(\bm{x},s)),
\end{equation}
where $\bm{x}$ is a point in the domain. The strain-history functional enforces irreversibility condition on the crack \cite{Miehe2010}. $\Psi_0^{+}$ is obtained as:
\begin{equation}\label{eq:tensileEnergy}
    {\Psi_0^{+}}(\bm{\epsilon}) = \frac{\lambda}{2}\left\langle {\text{tr}(\bm{\epsilon})}\right\rangle ^{2}_{+}  + \mu \text{tr}(\bm{\epsilon}^2)_{+},
\end{equation}
where $\lambda$ and $\mu$ are Lam\'e constants. The tensile strain, $\bm{\epsilon}_{+}$ is computed using the spectral decomposition of the strain tensor. The strain-history functional can be used to define initial cracks in the system \cite{Miehe2010a}. The initial strain history function ($H(\bm{x},0)$) could be defined as a function of the closest distance from $\bm{x}$ to the line ($l$), which represents the discrete crack \cite{Borden2012}. In particular, we set
\begin{equation}\label{eq:initial_history_field}
    H(\bm{x},0) = \left\{ {\begin{array}{l l}
  {\frac{BG_c}{2l_0}(1 - \frac{2d(\bm{x},l)}{l_0})}&{d(\bm{x},l) \leqslant \frac{l_0}{2}}\\ 
  0&{d(\bm{x},l) > \frac{l_0}{2}}
\end{array}} \right.,
\end{equation}
where $B$ is a scalar parameter that controls the magnitude of the scalar history field and is calculated as:
\begin{equation}\label{eq:scalarB}
    B = \frac{1}{1-\phi}  \; \; \; \text{for} \;\;\phi < 1.
\end{equation} 

We shall now present the results obtained using the physics informed neural networks to study phase field modeling of fracture.
\subsubsection{One-dimensional phase field model}
\label{subsec4:Frac_prob1}
The first example is to approximate the phase field in a one-dimensional bar of length, $L= 50$ units and has a crack located at 25 units. An analytical solution for a one-dimensional phase field model is available in \cite{Miehe2010}. Hence, it is possible to validate the results obtained from the proposed approach. The analytical solution for $\phi(x)$ is:
\begin{equation}\label{eq:crack_diffusion_2nd}
    \phi_{ex} (x) = \exp\left(\frac{-|x-a|}{l_0}\right),
\end{equation}
where the crack is located at $a$ units. We consider $l_0 = 1$ in this example, which is solved using DCM and DEM to compare the accuracy of both methods with the analytical solution.

\begin{figure}
 \centering
 \begin{subfigure}[b] {0.4\textwidth}
   \centering
   \includegraphics[width=0.9\textwidth]{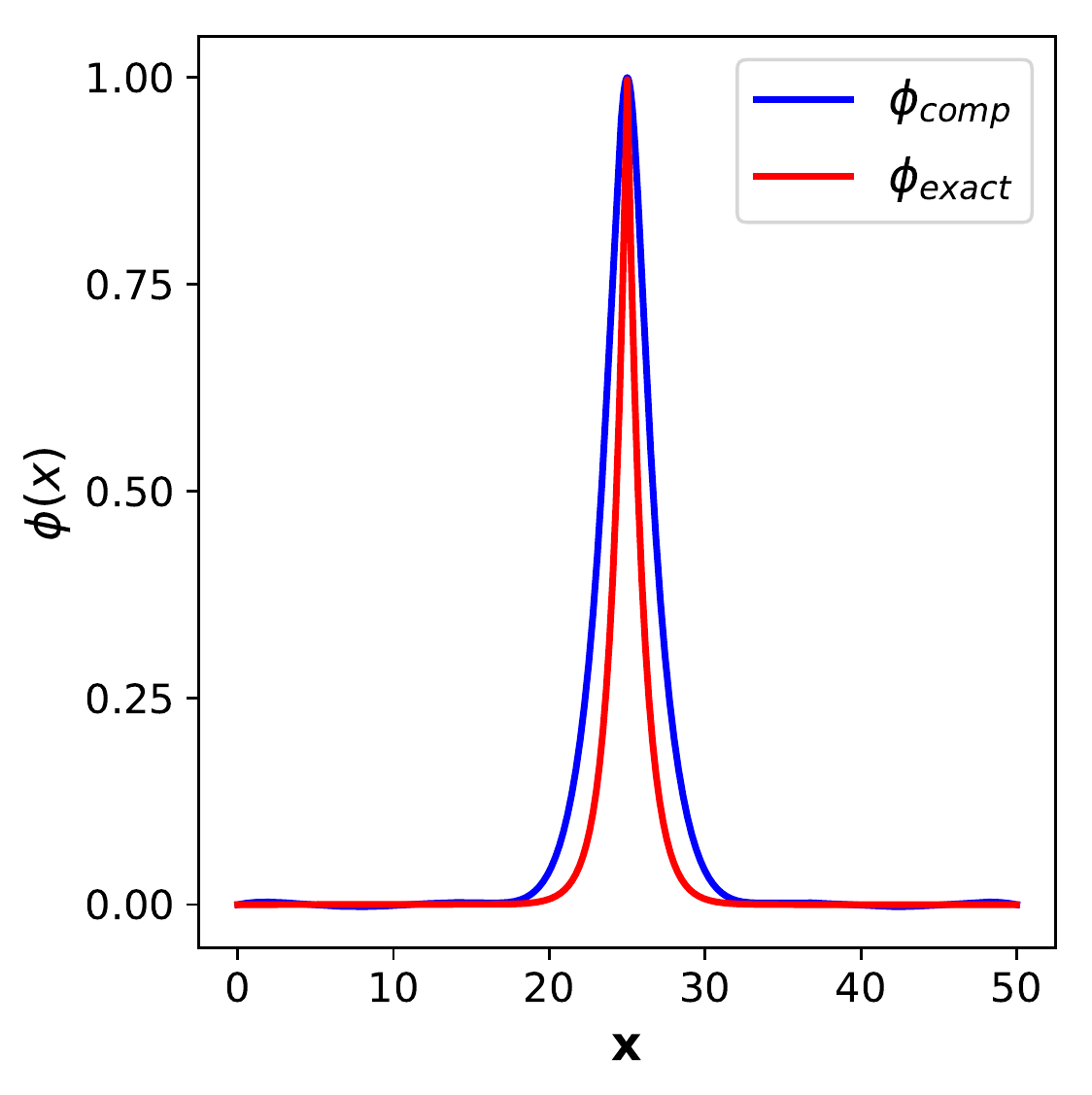}\hfill
   \caption{Comparison of $\phi_{exact}$ and $\phi_{comp}$.}
 \end{subfigure}\hspace{5pt}
 \begin{subfigure}[b] {0.4\textwidth}
   \centering
   \includegraphics[width=0.9\textwidth]{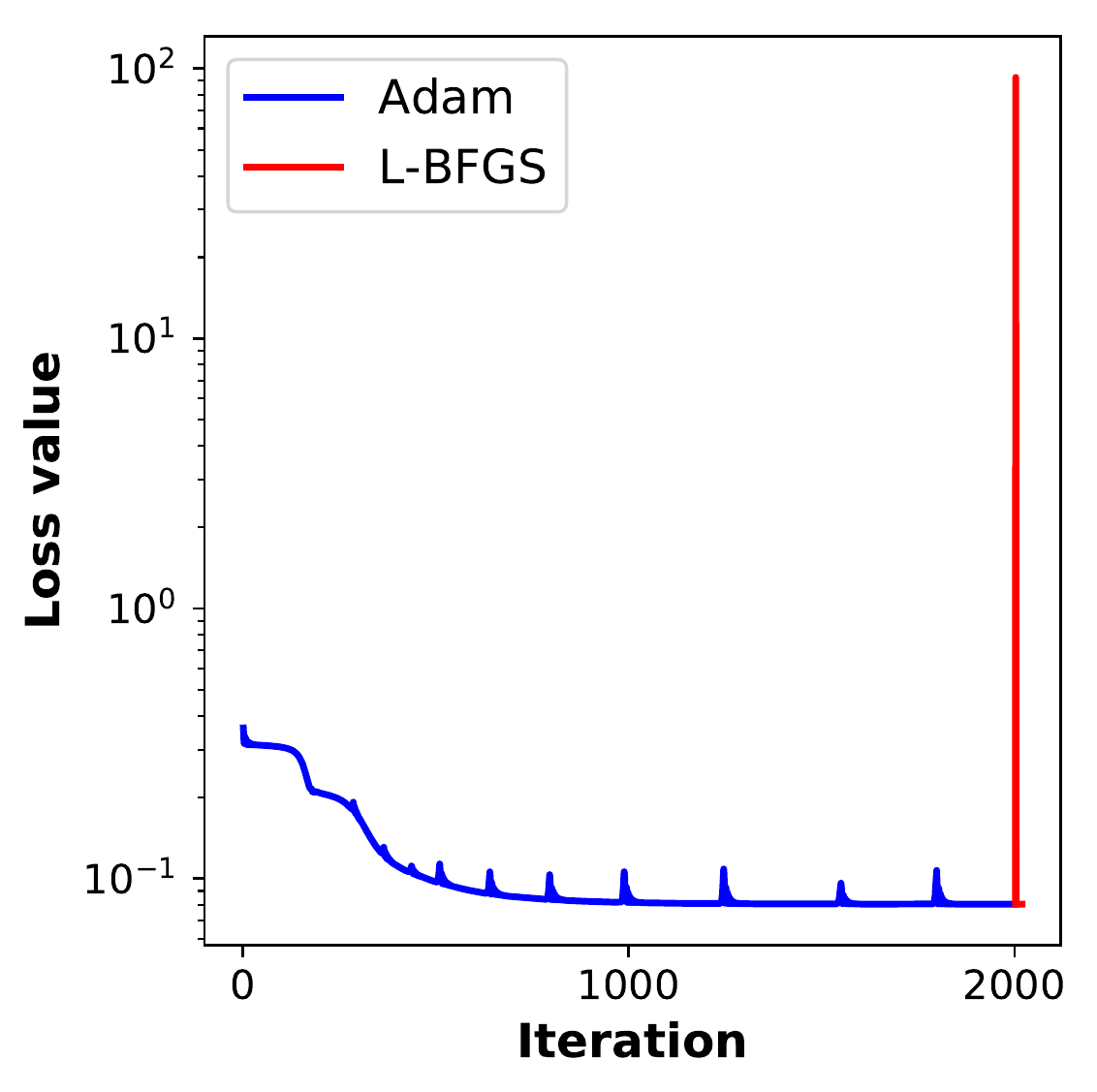}\hfill
   \caption{Convergence of the loss function.}
 \end{subfigure}
 \caption{One-dimensional phase field model using the collocation method.}
 \label{fig:Collo_1D}
 \end{figure}

The formulation of the collocation method is to minimize the function $f$, defined by
\begin{equation}\label{eq:1d_ODE_2nd}
    f(x) = \phi''(x) - \frac{1}{l_0^2} \phi (x) \text{ in } \Omega,
\end{equation}
at a set of chosen collocation points, $x$ subject to the Dirichlet-type boundary conditions: 
\begin{equation}\label{eq:1D-dirichlet}
\begin{split}
    \phi\left(a\right) &= 1 \text{  , } \phi'\left(a\right) = 0 \text{  , }\\ \lim\limits_{x\to\infty}\phi\left(x\right) &=  \lim\limits_{x\to-\infty}\phi\left(x\right)= 0, \text{  and  } \\
      \lim\limits_{x\to\infty}\phi'\left(x\right) &=  \lim\limits_{x\to-\infty}\phi'\left(x\right)= 0.
    \end{split}
\end{equation}
We have used a network with 3 hidden layers of $50$ neurons each and $N_{Col} = 8000$ uniformly spaced points within the domain. To find the solution of $\phi$ in the domain $[0,50]$, the trial solution is defined as
\begin{equation}\label{eq:boundary1D}
    \phi(x) = -\frac{x(x-50)}{625}\left[(x-25)\hat\phi + 1\right],
\end{equation}
where $\hat\phi$ is given by the neural network. The solution of $\phi$ is chosen in such a way that it satisfies all the boundary conditions as in \cite{avrutskiy2017enhancing, berg2018unified}. This ensures that the boundary conditions are satisfied during the training of the network. In the collocation method, the mean squared error of the function, $f$ at the collocation points is minimized. The loss function, $\mathcal{L}_{Col}$ for the collocation method is defined as:
\begin{equation}\label{eq:loss1DCollo}
    \mathcal{L}_{Col} = \frac{1}{N_{Col}}\sum_{i=1}^{N_{Col}}|f(x_i)|^2.
\end{equation}
For optimization, we use the Adam (adaptive momentum) optimizer followed by a quasi-Newton method (L-BFGS). The boundary conditions are satisfied exactly as could be seen in \autoref{fig:Collo_1D}(a), where $\phi_{comp}$ represents the value of $\phi$ obtained using the neural network and $\phi_{exact}$ is obtained using \autoref{eq:crack_diffusion_2nd}. \autoref{fig:Collo_1D}(b) shows the convergence of the loss function first with the Adam optimizer and later using the L-BFGS optimizer. To measure the accuracy of the method, the relative $\mathcal{L}_2$ error, $\mathcal{L}_2^{rel}$ is computed using using the formula:  
\begin{equation}\label{eq:L2norm}
    \mathcal{L}_2^{rel} = \frac{\sqrt{\sum_{i = 1}^{N_{pred}}{(\phi_{comp} - \phi_{ex})^2}dx}}{\sqrt{\sum_{i = 1}^{N_{pred}}{\phi_{ex}^2}dx}}.
\end{equation}
For the collocation method, $\mathcal{L}_2^{rel} = 70.6\%$ due to the inability of the collocation to capture the discontinuity in the derivative at $x=25$.

\begin{figure}
 \centering
 \begin{subfigure}[b] {0.4\textwidth}
   \centering
   \includegraphics[width=\textwidth]{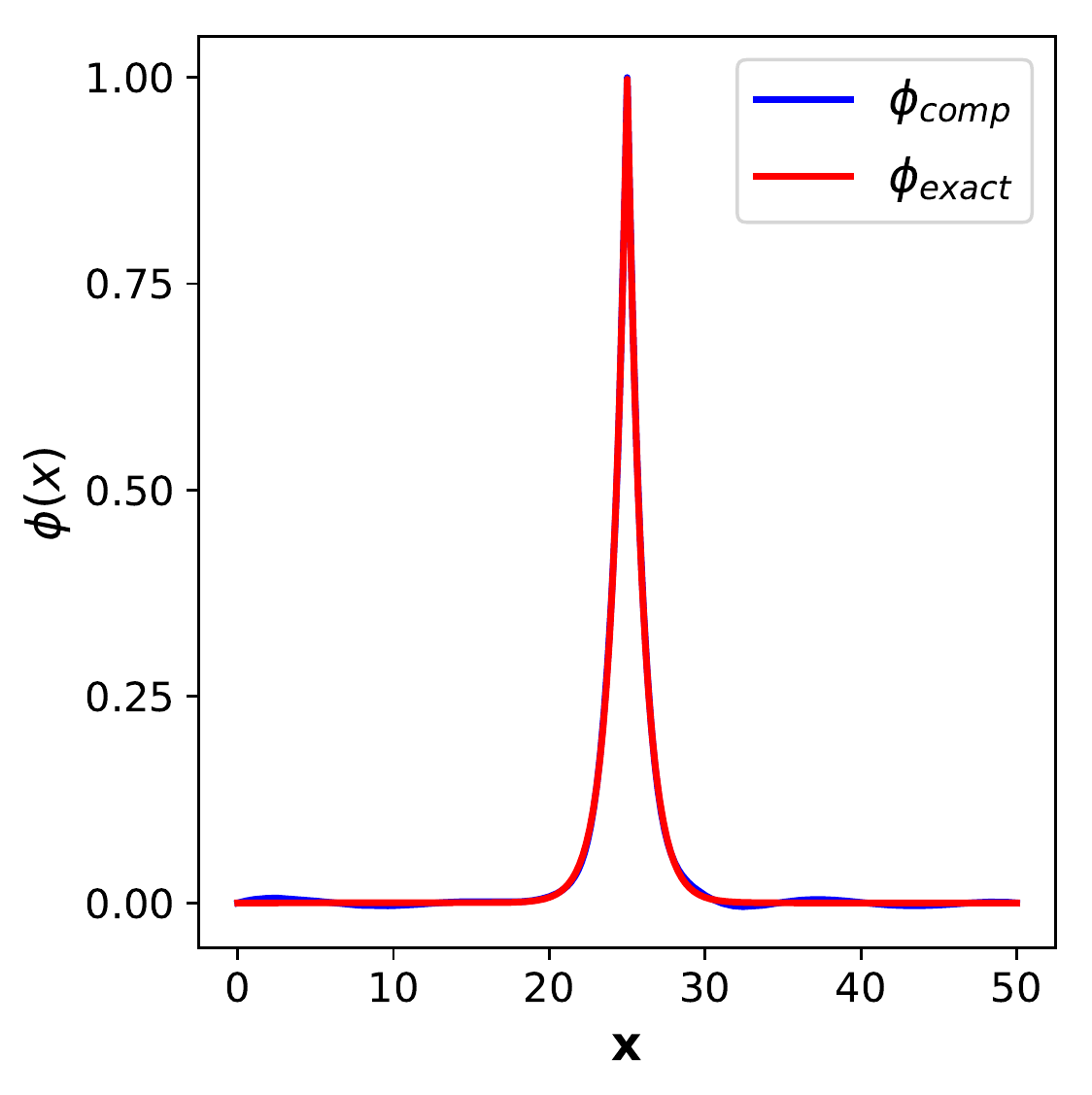}\hfill
   \caption{Comparison of $\phi_{exact}$ and $\phi_{comp}$.}
 \end{subfigure}\hspace{5pt}
 \begin{subfigure}[b] {0.4\textwidth}
   \centering
   \includegraphics[width=\textwidth]{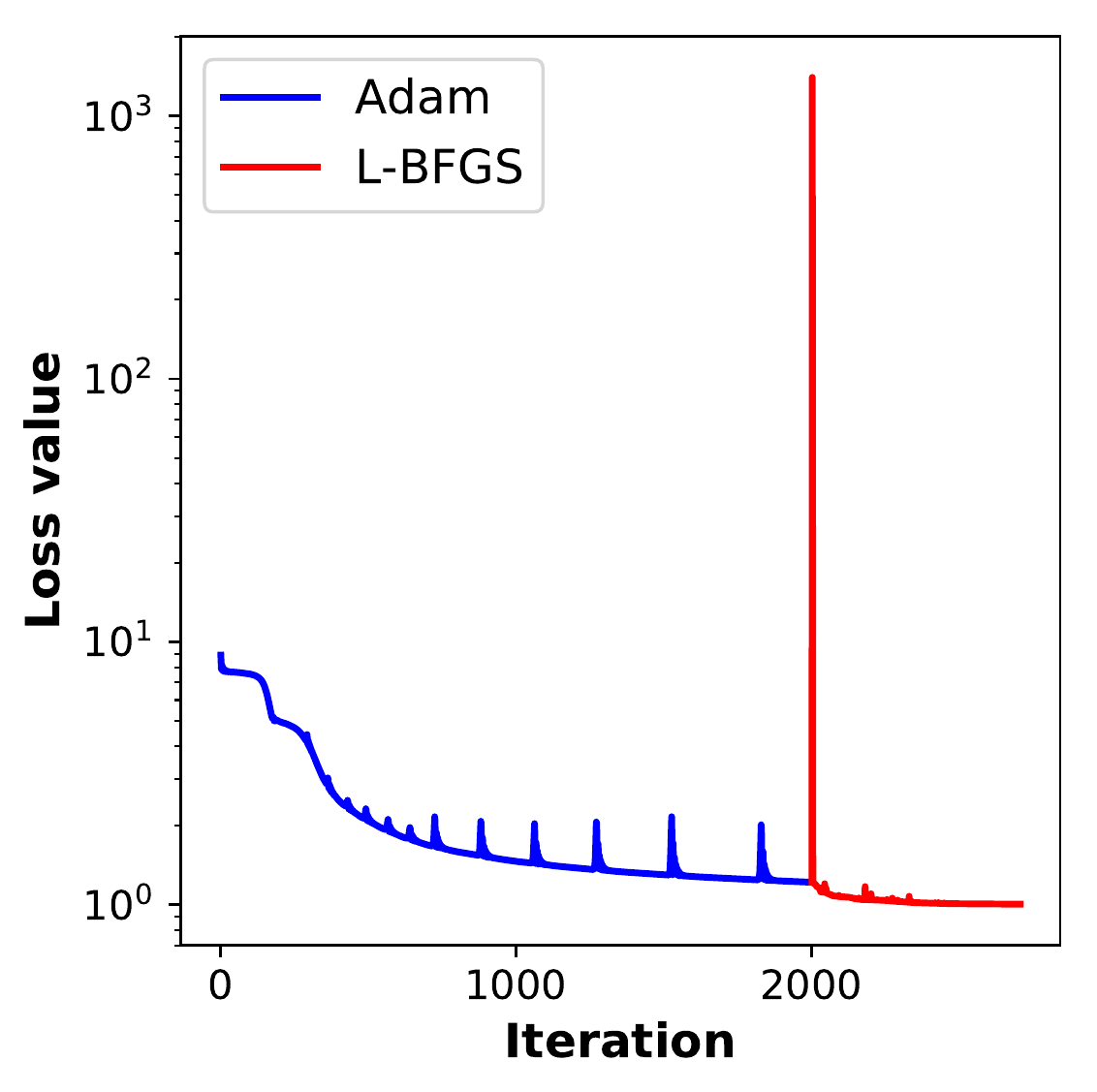}\hfill
   \caption{Convergence of the loss function.}
 \end{subfigure}\hspace{5pt}
 \caption{One-dimensional phase field model using the energy method.}
 \label{fig:Energy_1D}
 \end{figure}

Using the same network architecture and the same number of integration points, $N_{Int} = 8000$ in the interior domain, we solve the one-dimensional problem using DEM. For DEM, the problem is written as:
\begin{equation}\label{eq:1d_energy}
        \text{Minimize:}\;\;\;\; I\left(\phi\right) = \frac{1}{2}\int\limits_\Omega{\left(\phi(x) - l_0^2|\nabla\phi|^2 \right)}dx, 
\end{equation}
subject to the same boundary conditions as in \autoref{eq:1D-dirichlet}. The form of the solution used for DEM is as stated in \autoref{eq:boundary1D}. The loss function, $\mathcal{L}_{Ener}$ for DEM is defined as
\begin{equation}\label{eq:loss1DEnergy}
    \mathcal{L}_{Ener} = \frac{L}{N_{Int}}\sum_{i=1}^{N_{Int}}\left(\phi(x_i) - l_0^2|\nabla \phi (x_i)|^2\right).
\end{equation}
\autoref{fig:Energy_1D}(a) presents the solution of $\phi$ obtained using DEM, while \autoref{fig:Collo_1D}(b) shows the convergence of the loss function. For DEM, $\mathcal{L}_2^{rel} = 2.88\%$.

\begin{figure}
 \centering
 \begin{subfigure}[b] {0.4\textwidth}
   \centering
   \includegraphics[width=\textwidth]{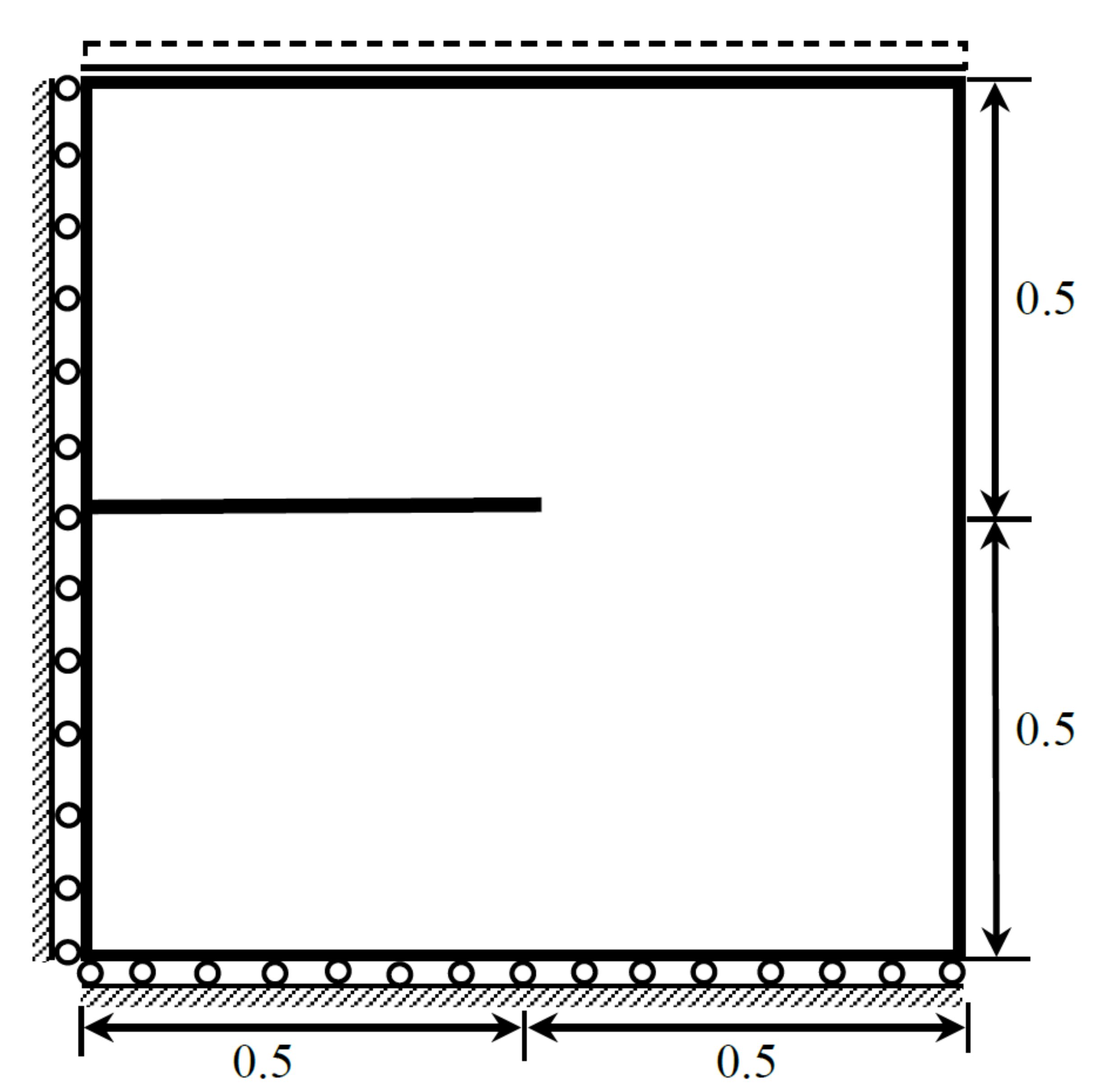}\hfill
   \caption{Geometrical setup and boundary conditions.}
 \end{subfigure}\hspace{5pt}
 \begin{subfigure}[b] {0.4\textwidth}
   \centering
   \includegraphics[width=\textwidth]{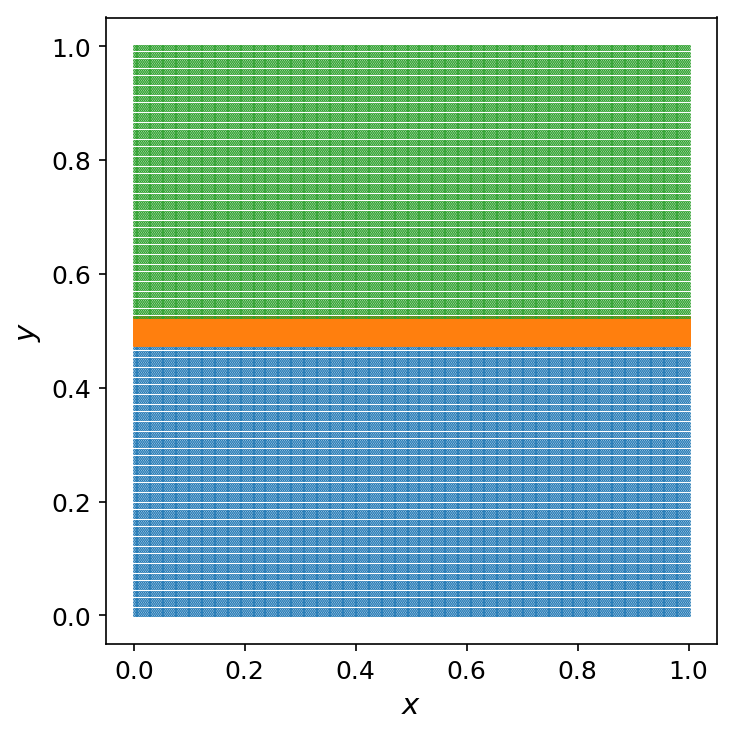}\hfill
   \caption{Training grid.}
 \end{subfigure}\hspace{5pt}
 \caption{Single-edge notch tension example.}
 \label{fig:setup}
 \end{figure}
 
\subsubsection{Single-edge notched tension example}
\label{subsec4:Frac_prob2}
\begin{figure}
 \centering
 \begin{subfigure}[b] {0.4\textwidth}
   \centering
   \includegraphics[width=\textwidth]{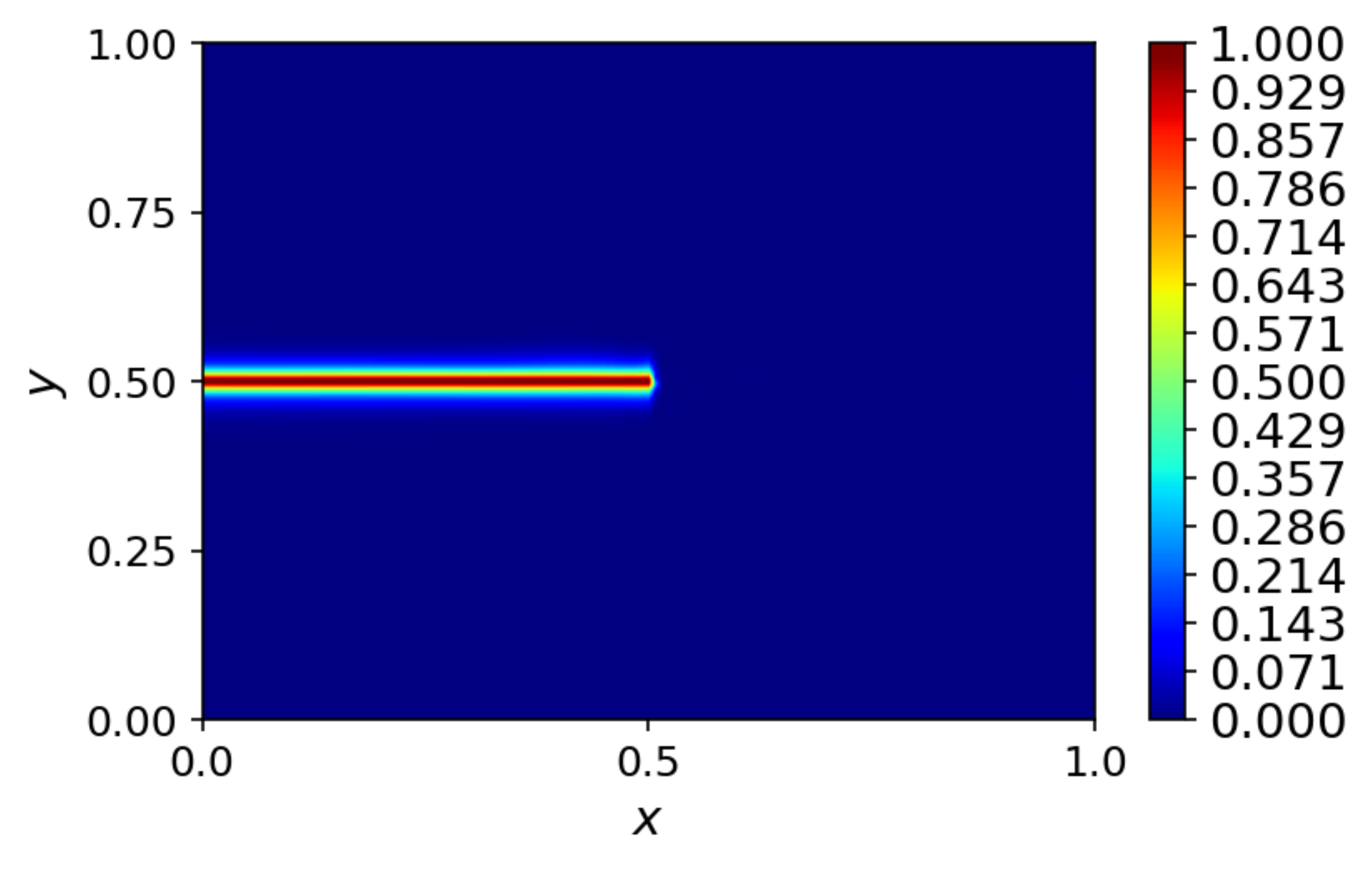}\hfill
   \caption{Initial crack using DEM.}
 \end{subfigure}\hspace{5pt}
 \begin{subfigure}[b] {0.4\textwidth}
   \centering
   \includegraphics[width=\textwidth]{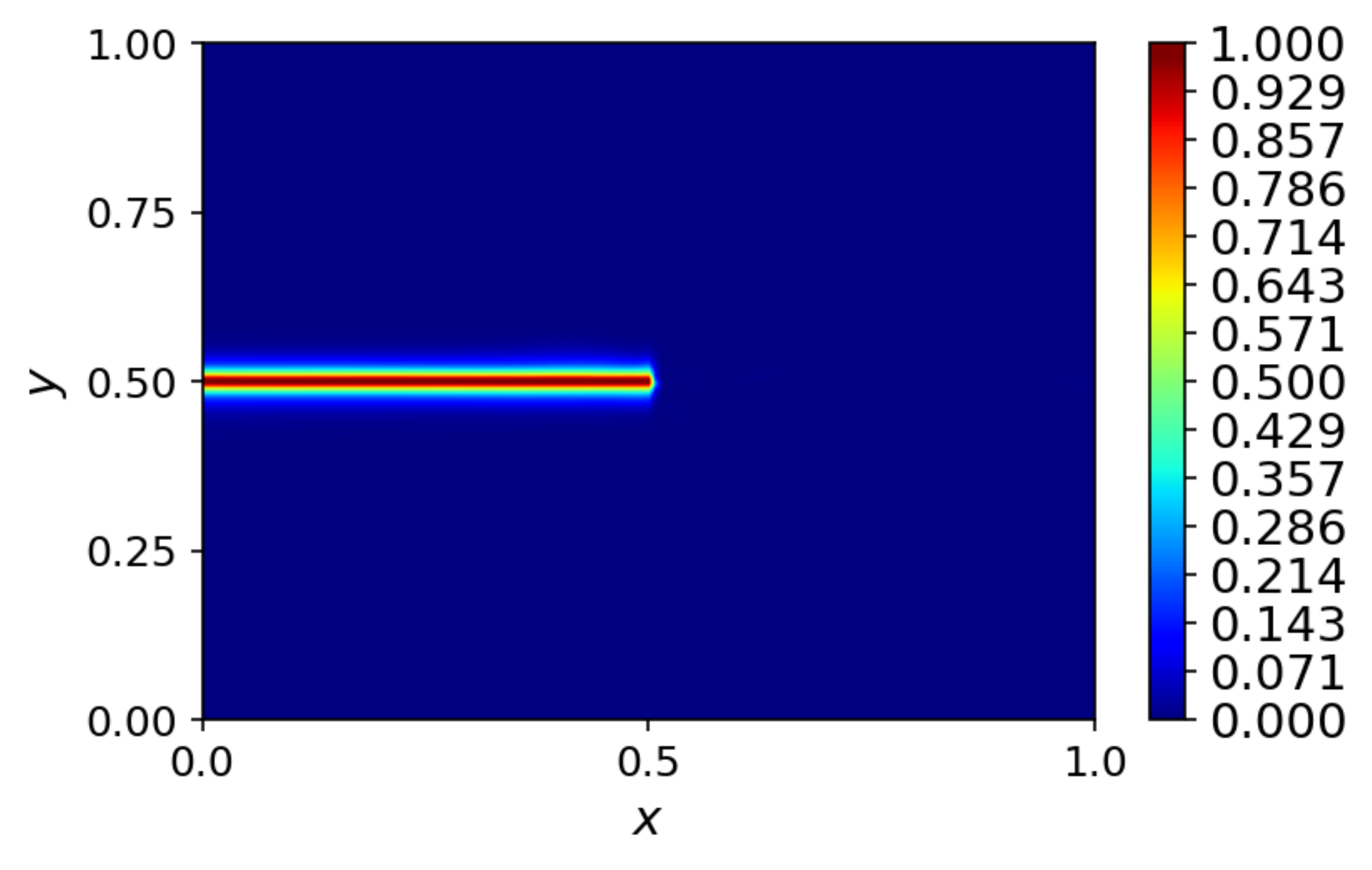}\hfill
   \caption{Initial crack using DCM.}
 \end{subfigure}\hspace{5pt}
  \begin{subfigure}[b] {0.4\textwidth}
   \centering
   \includegraphics[width=\textwidth]{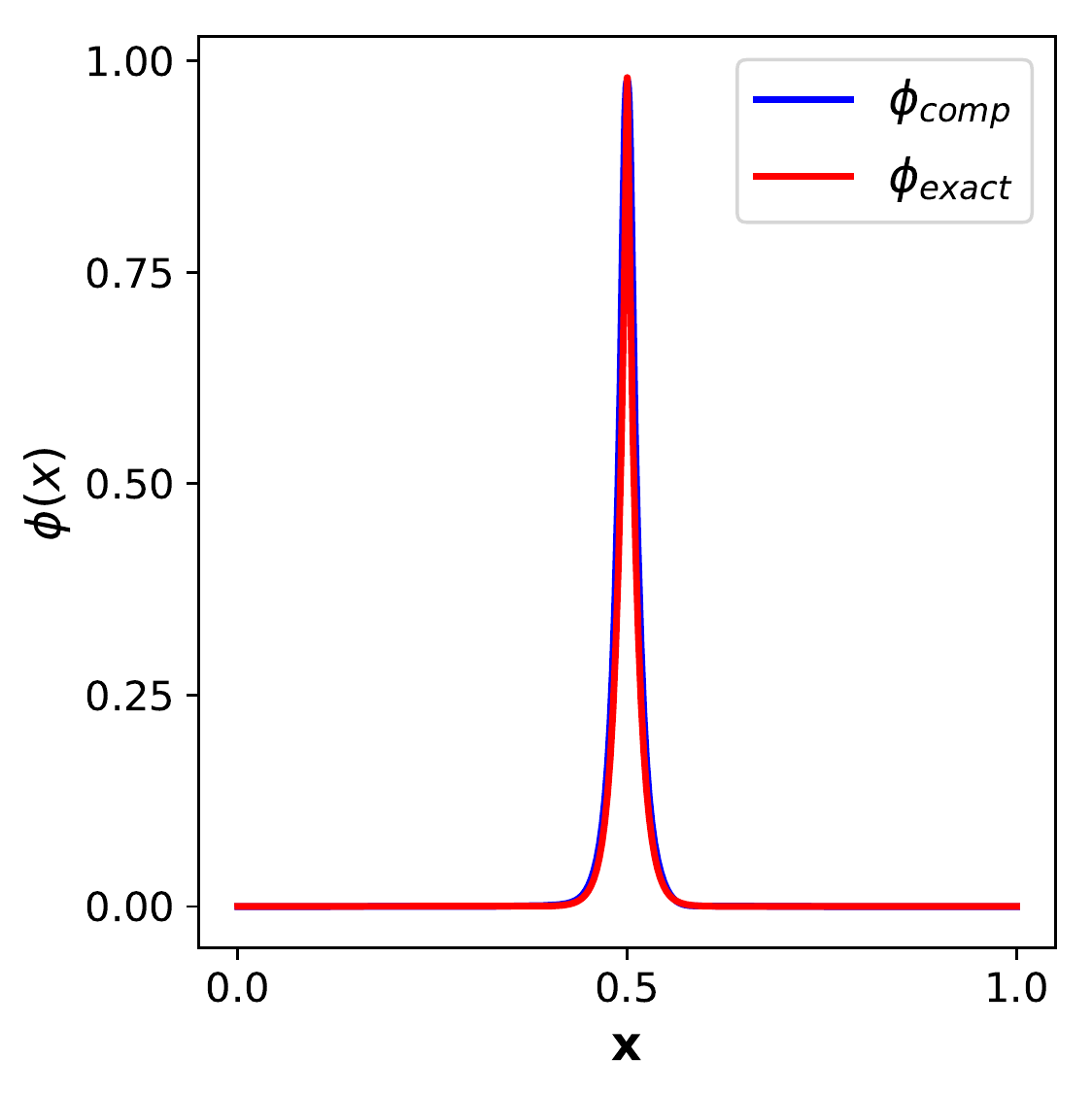}\hfill
   \caption{Comparison of $\phi_{exact}$ and $\phi_{comp}$ using DEM.}
 \end{subfigure}\hspace{5pt}
  \begin{subfigure}[b] {0.4\textwidth}
   \centering
   \includegraphics[width=\textwidth]{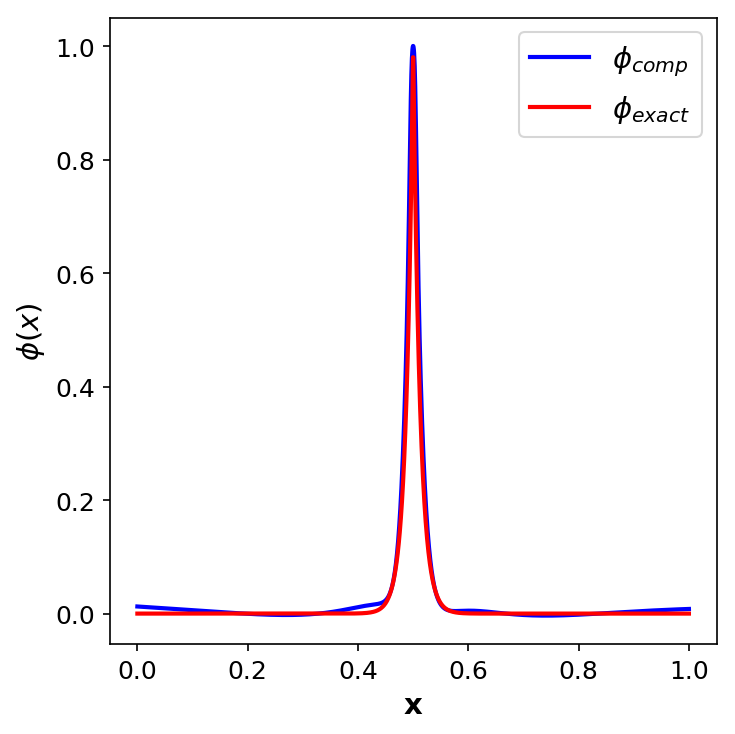}\hfill
   \caption{Comparison of $\phi_{exact}$ and $\phi_{comp}$ using DCM.}
 \end{subfigure}\hspace{5pt}
  \begin{subfigure}[b] {0.4\textwidth}
   \centering
   \includegraphics[width=\textwidth]{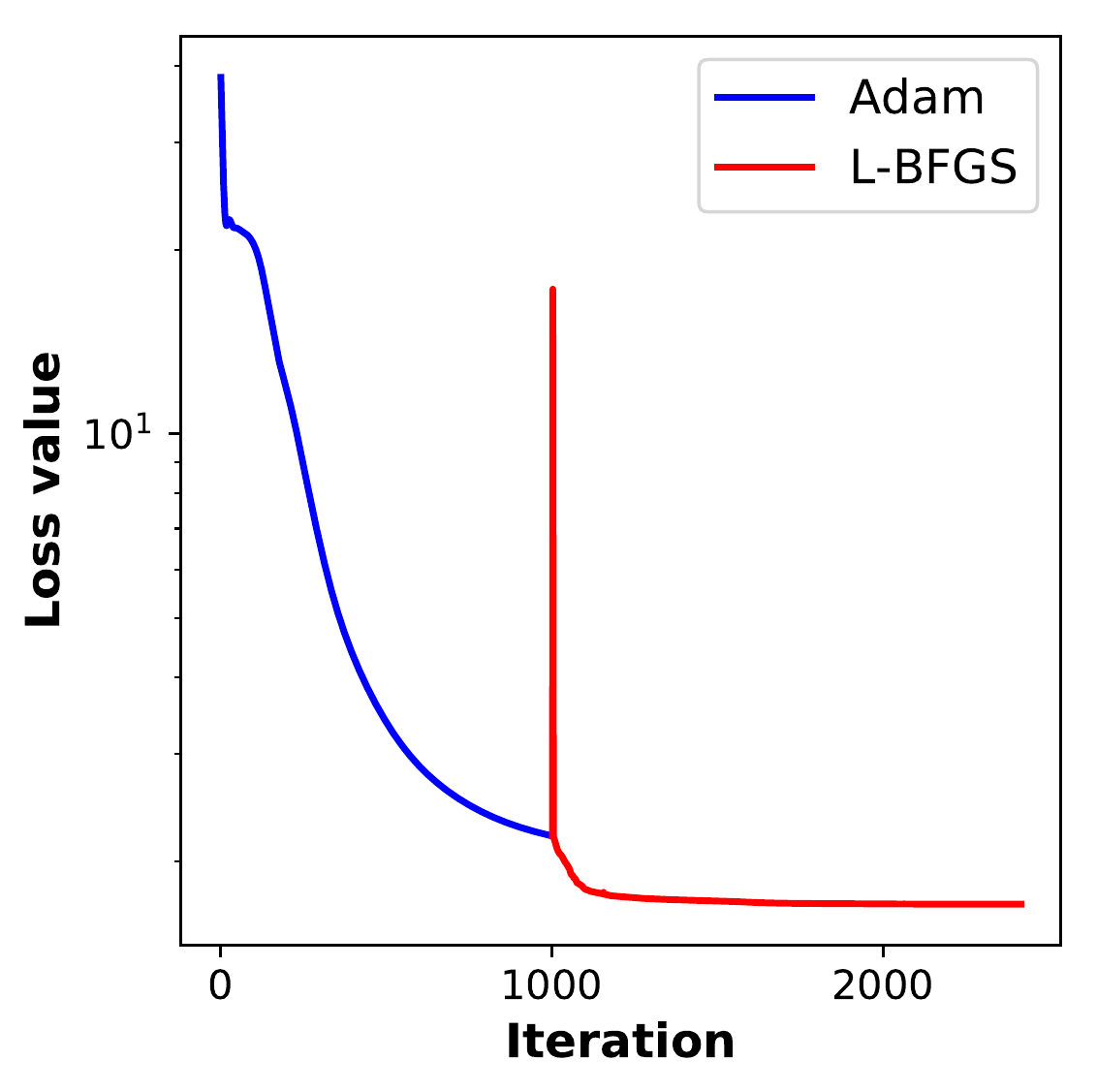}\hfill
   \caption{Convergence of the loss function using DEM.}
 \end{subfigure}\hspace{5pt}
  \begin{subfigure}[b] {0.4\textwidth}
   \centering
   \includegraphics[width=\textwidth]{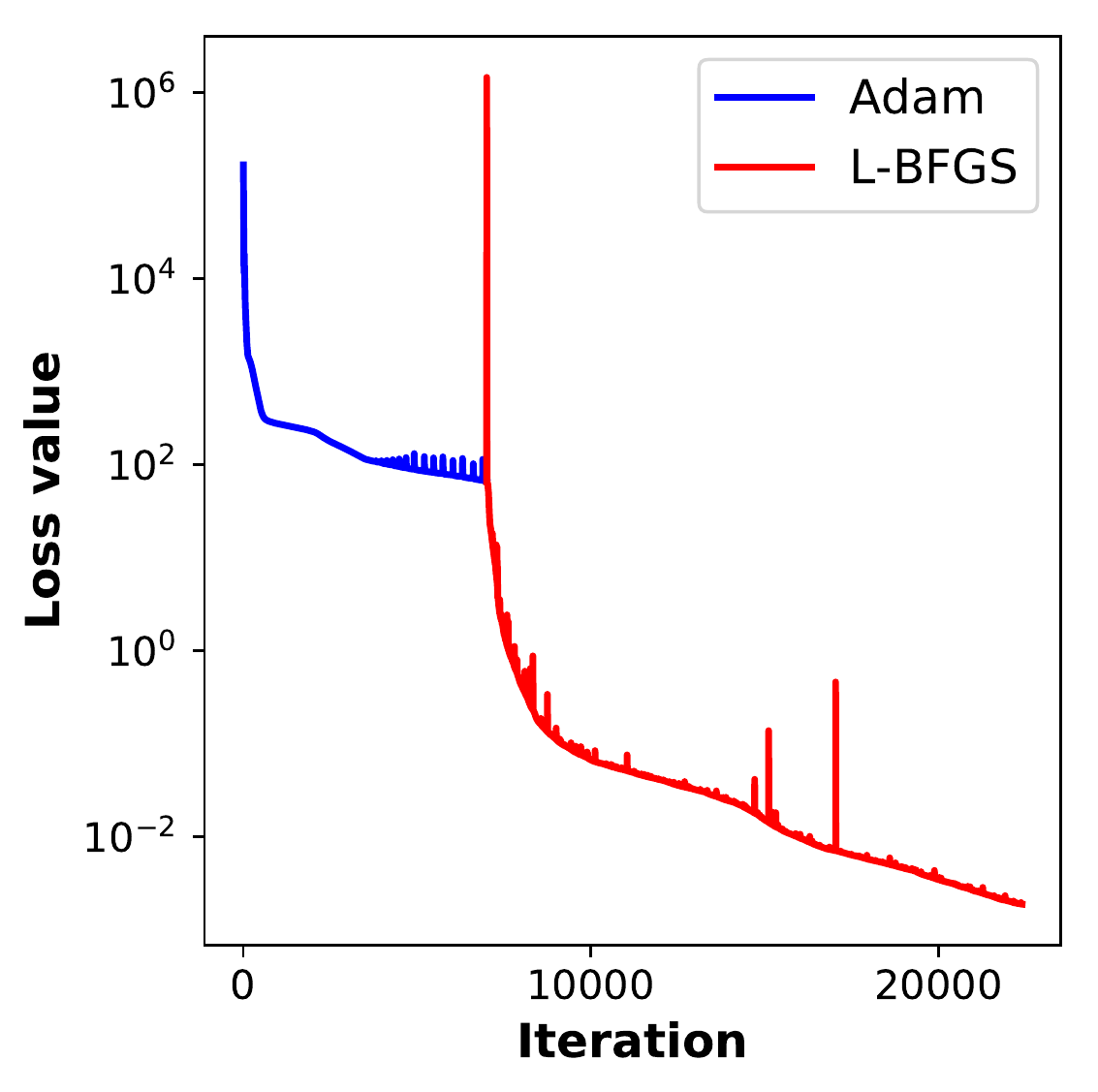}\hfill
   \caption{Convergence of the loss function using DCM.}
 \end{subfigure}\hspace{5pt}
 \caption{Initialization of crack in the plate using the strain-history function.}
 \label{fig:phiInitial}
 \end{figure}

In this example, we consider a unit square plate with a horizontal crack from the midpoint of the left outer edge to the center of the plate. The geometric setup and boundary conditions of the problem are shown in \autoref{fig:setup}(a). In this example, we consider $l_0 = 0.0125$ and $G_c = 2.7 \times 10^{-3}$ kN/mm. The material properties of the plate are expressed in terms of L\'ame's constants, $\lambda = $ 121.15 kN/mm$^{2}$ and $\mu = $ 80.77 kN/mm$^{2}$. The domain is subdivided into 3 sections along the y-axis; the crack zone, the bottom and the top of the crack. The crack zone is between [($0.5-2l_0$), ($0.5+2l_0$)] on the y-axis. The training grid is shown in \autoref{fig:setup}(b). The strain history function is used to initialize the crack in the domain as in \autoref{eq:history_field}. We have used a network with 3 hidden layers of $20$ neurons each and $N_x\times N_y$ uniformly spaced points for each section in the interior of the plate with $N_x = 300$ and $N_y = 81$. The problem statement is defined as:
\begin{equation}\label{eq:2d_energy}
    \begin{split}
        \text{Minimize:}\;\;\;\; & I\left(\phi\right) = \int\limits_\Omega{f(\phi(\bm{x}))}dx,\\
        \text{where}\;\;\;\; &f(\bm{x}) = \frac{G_c}{2l_0}\left(\phi(\bm{x})^2 + l_0^2|\nabla\phi(\bm{x})|^2  + g(\phi(\bm{x}))H(\bm{x},0)\right).\\
    \end{split}
\end{equation}
The loss function, $\mathcal{L}_{0}$ is defined as:
\begin{equation}\label{eq:loss_phiInitial}
    \begin{split}
        A_{tc} &= A_{bc} = (0.5-2l_0),\\
        A_{c} &= 4l_0,\\
        \mathcal{L}_{0} &= \frac{1}{N_x\times N_y}\left(A_{tc}\sum_{i = 1}^{N_x\times N_y}f(\bm{x}_i^{tc}) + A_{c}\sum_{i = 1}^{N_x\times N_y}f(\bm{x}_i^{c}) + A_{bc}\sum_{i = 1}^{N_x\times N_y}f(\bm{x}_i^{bc})\right), 
    \end{split}
\end{equation}
where $A_{tc}$, $A_{bc}$, $A_{c}$ denote the areas of section on top and bottom of the crack and the crack zone, respectively. $\bm{x_i^{tc}}$, $\bm{x_i^{c}}$ and $\bm{x_i^{bc}}$ in \autoref{eq:loss_phiInitial} represent points in the three sub domains in the interior of the plate. To compare the performance of DEM, we also initialize the crack using the collocation method. The formulation of the collocation method is to minimize the function $f$, defined by
\begin{equation}\label{eq:2D_collo}
\begin{split}
    & f(x) = G_cl_0 \Delta \phi(\bm{x}) - \frac{G_c}{l_0} \phi (\bm{x}) - g'(\phi(\bm{x}))H(\bm{x},0)  \text{ in } \Omega,\\
    \text{and}\;\;\;\; &g'(\phi(\bm{x})) = -2(1-\phi(\bm{x}))
\end{split}
\end{equation}
subjected to homogeneous Neumann-type boundary conditions on the entire boundary defined as:
\begin{equation}\label{eq:Phase_boundary}
    \nabla\phi\cdot \bm{n} = 0 \text{ on } \partial\Omega.
\end{equation}
We have used a network with 5 hidden layers of $20$ neurons each and $N_x\times N_y$ uniformly spaced points for each section in the interior of the plate and $N_{Bound}$ uniformly spaced points on each edge of the plate, with $N_x = 300$ $N_y = 81$ and $N_{Bound} = 200$. The loss function, $\mathcal{L}_{Col}$ for the collocation method is defined as:
\begin{equation}\label{eq:loss2DCollo}
\begin{split}
    \mathcal{L}_{Col} &= \frac{1}{N_x\times N_y}\left(A_{tc}\sum_{i = 1}^{N_x\times N_y}|f(\bm{x}_i^{tc})|^2 + A_{c}\sum_{i = 1}^{N_x\times N_y}|f(\bm{x}_i^{c})|^2 + A_{bc}\sum_{i = 1}^{N_x\times N_y}|f(\bm{x}_i^{bc})|^2\right) + \\
    &\frac{1}{N_{Bound}}\sum_{i=1}^{N_{Bound}} \left(\left[\frac{\partial\phi(\bm{x}_i^L)}{\partial x}\right]^2 + \left[\frac{\partial\phi(\bm{x}_i^R)}{\partial x}\right]^2 + \left [\frac{\partial\phi(\bm{x}_i^T)}{\partial y}\right]^2 + \left[\frac{\partial\phi(\bm{x}_i^B)}{\partial y}\right]^2\right),
\end{split}
\end{equation}
where $\bm{x}_i^L, \bm{x}_i^R, \bm{x}_i^T, \bm{x}_i^B$ are the points on the left, right, top and bottom edges of the plate, respectively.
\autoref{fig:phiInitial}(a) and (b) show the initial crack in the plate using DEM and DCM, respectively. \autoref{fig:phiInitial}(c) and (d) present the one-dimensional approximation plots and  \autoref{fig:phiInitial}(e) and (f) show the convergence of the loss function for both the methods.

In \autoref{subsec4:Frac_prob1}, we observed that DEM performs better than DCM for the same network parameters. From the results of the initilization of the crack, we can conclude that DEM requires a smaller network to predict results accurately. Moreover, it is also seen that DCM needs an explicit treatment of the Neumann boundary conditions even for traction-free boundaries, which requires the use of more collocation points. The convergence of the loss function is also slow, requiring $10$ times more iterations. Hence, we use DEM to study the growth of crack in a plate under tensile loading. 

\begin{figure}
 \centering
 \begin{subfigure}[b] {0.3\textwidth}
   \centering
   \includegraphics[width=\textwidth]{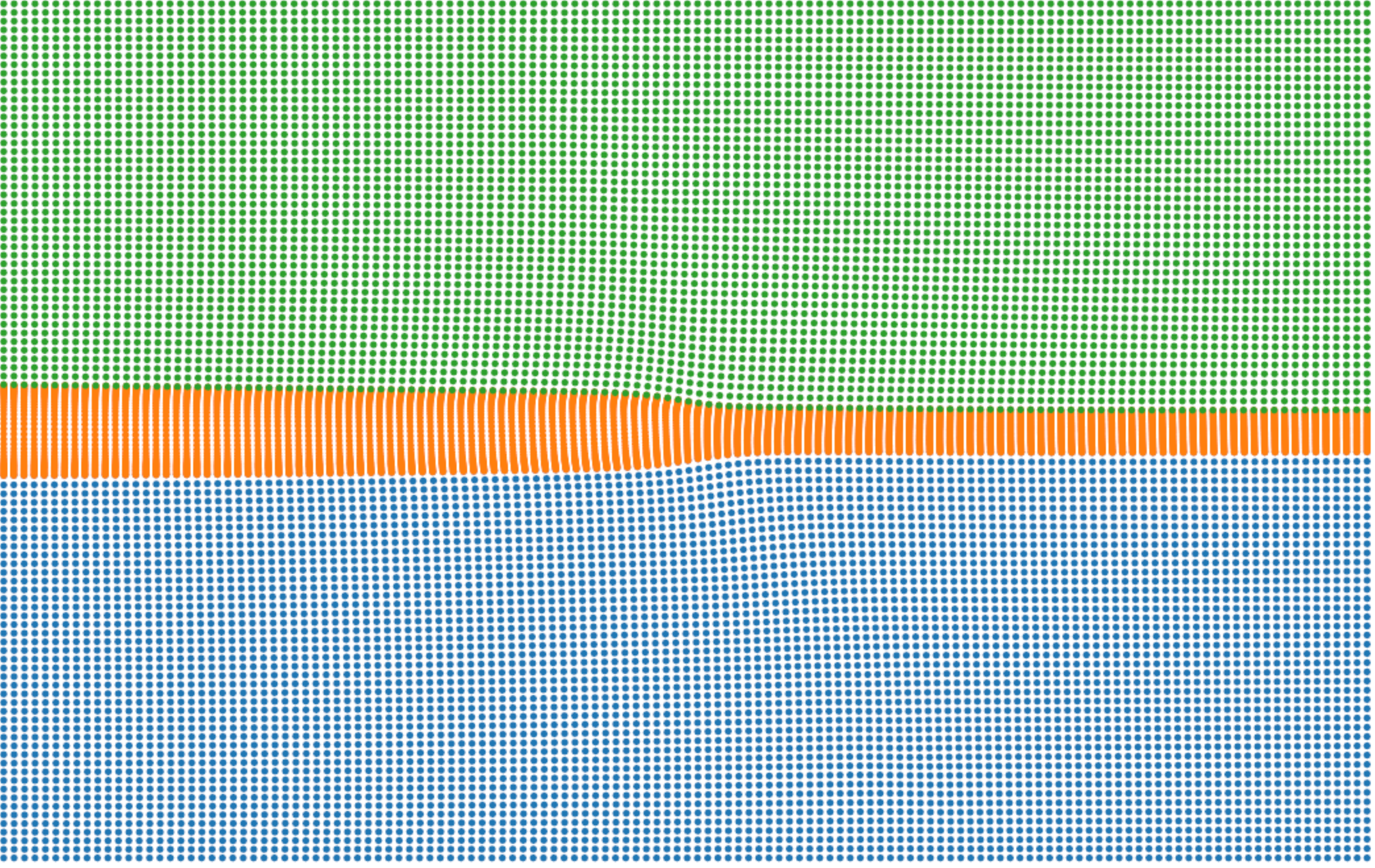}\hfill
   \caption{}
 \end{subfigure}\hspace{5pt}
 \begin{subfigure}[b] {0.3\textwidth}
   \centering
   \includegraphics[width=\textwidth]{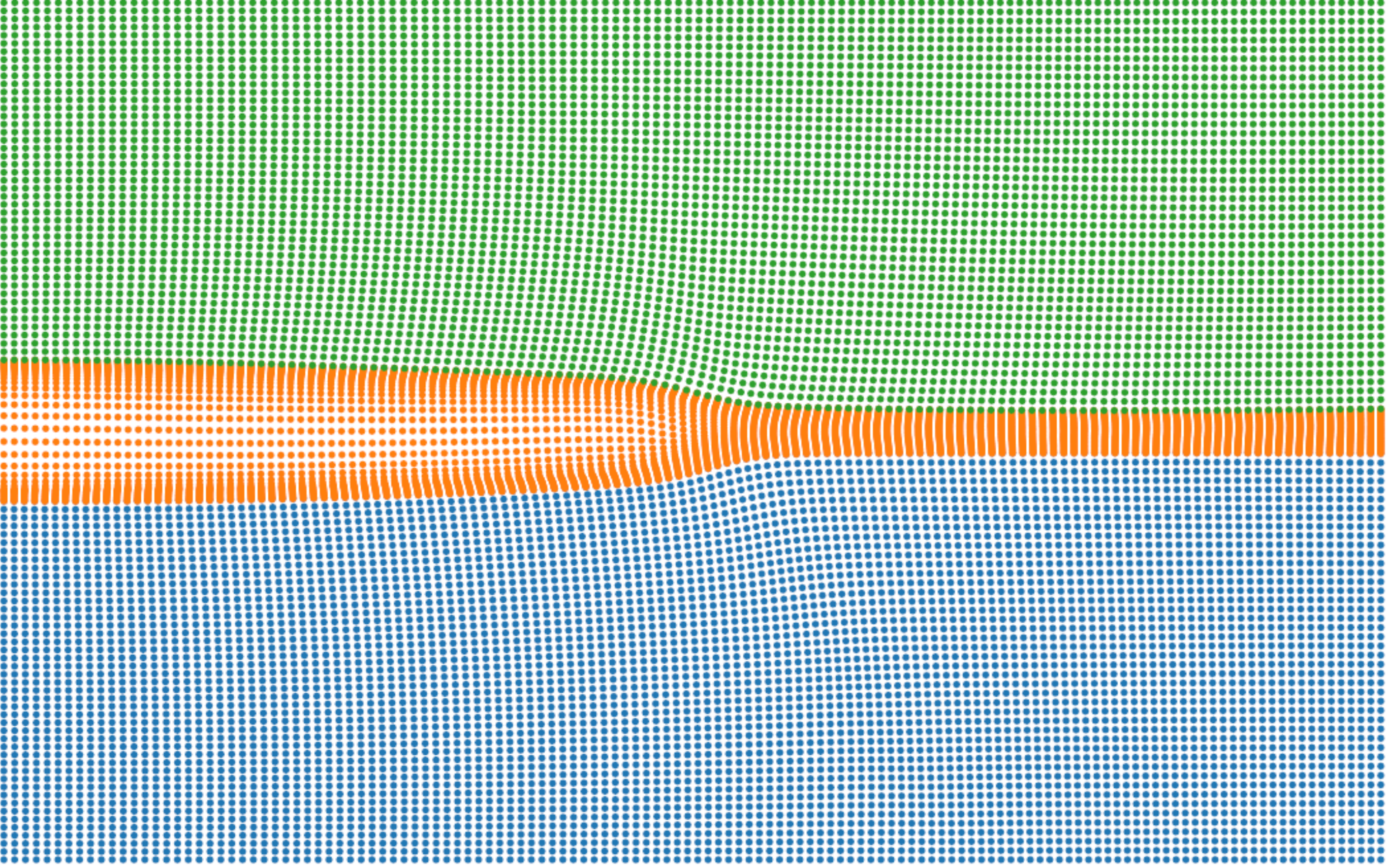}\hfill
   \caption{}
 \end{subfigure}\hspace{5pt}
  \begin{subfigure}[b] {0.3\textwidth}
   \centering
   \includegraphics[width=\textwidth]{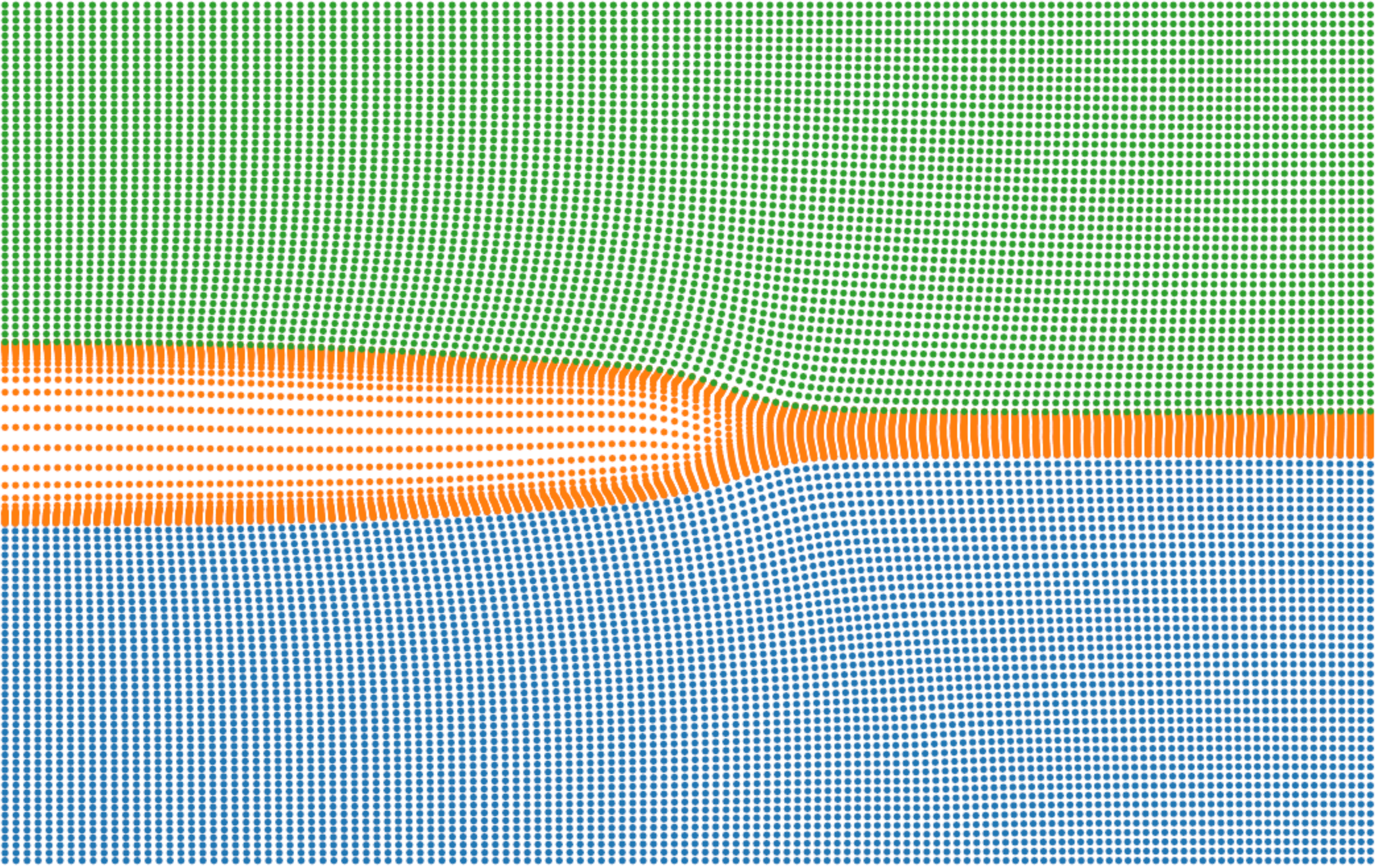}\hfill
   \caption{}
 \end{subfigure}\hspace{5pt}
  \begin{subfigure}[b] {0.3\textwidth}
   \centering
   \includegraphics[width=\textwidth]{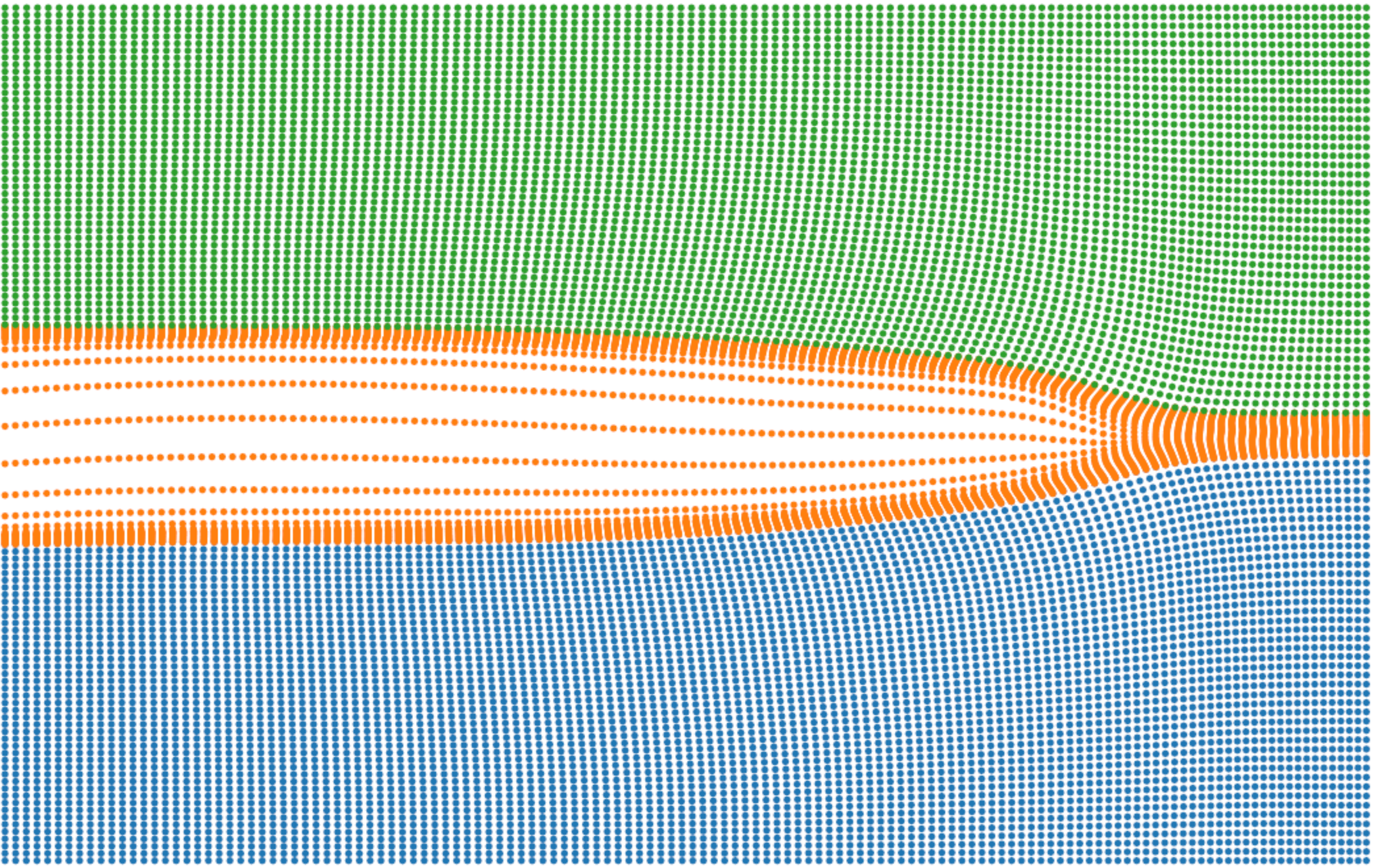}\hfill
   \caption{}
 \end{subfigure}\hspace{5pt}
  \begin{subfigure}[b] {0.3\textwidth}
   \centering
   \includegraphics[width=\textwidth]{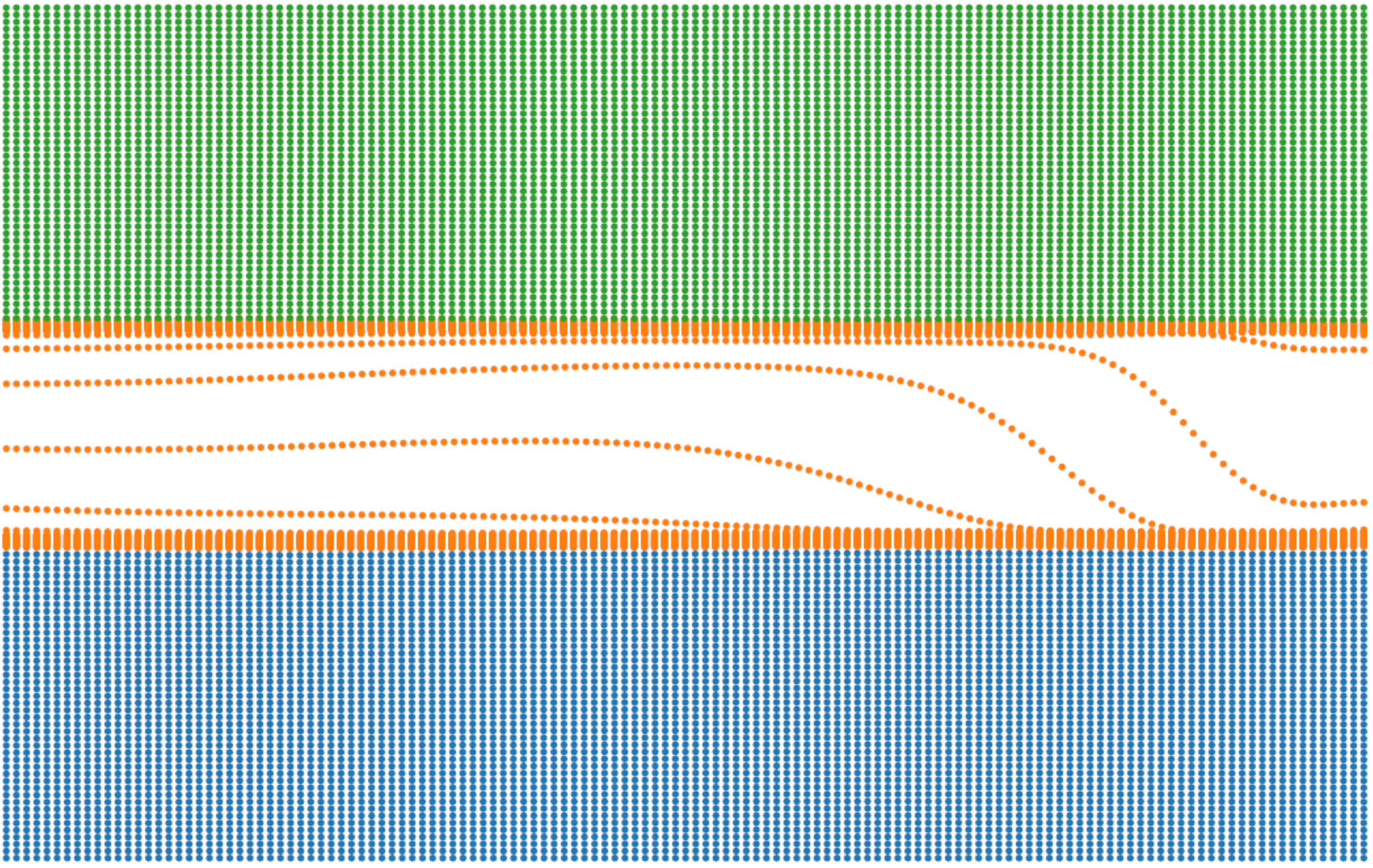}\hfill
   \caption{}
 \end{subfigure}\hspace{5pt}
 \caption{Scatter plots of the deformed configuration for prescribed displacement of (a) $1\times10^{-3}$, (b) $2\times10^{-3}$, (c) $3\times10^{-3}$, (d) $4\times10^{-3}$ and (e) $5\times10^{-3}$.}
 \label{fig:crackflow}
 \end{figure}

\begin{figure}
 \centering
 \begin{subfigure}[b] {0.3\textwidth}
   \centering
   \includegraphics[width=\textwidth]{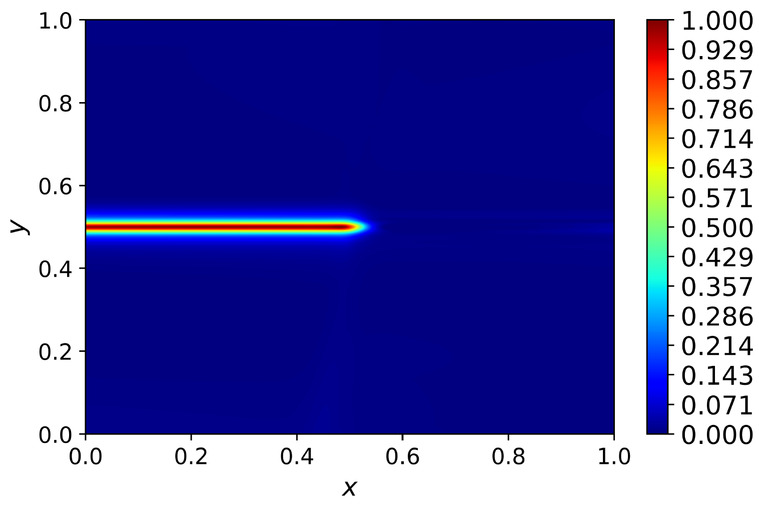}\hfill
   \caption{}
 \end{subfigure}\hspace{5pt}
 \begin{subfigure}[b] {0.3\textwidth}
   \centering
   \includegraphics[width=\textwidth]{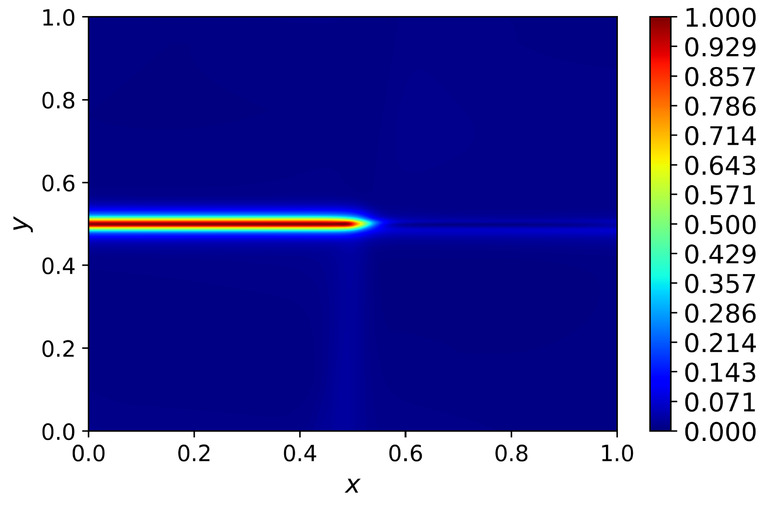}\hfill
   \caption{}
 \end{subfigure}\hspace{5pt}
  \begin{subfigure}[b] {0.3\textwidth}
   \centering
   \includegraphics[width=\textwidth]{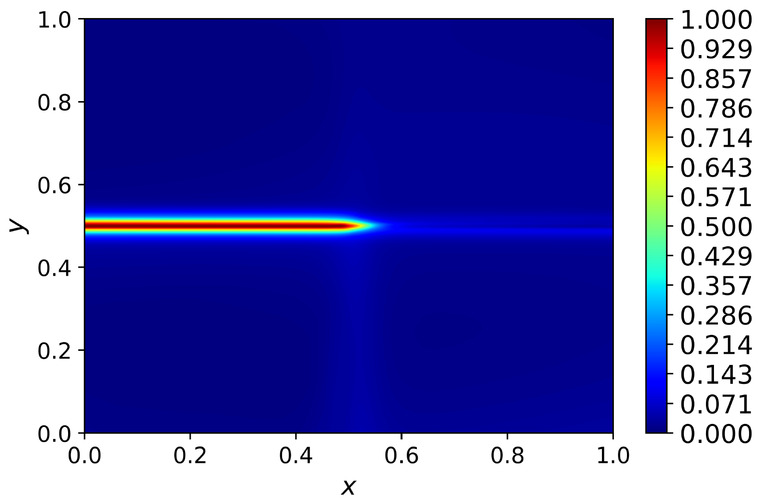}\hfill
   \caption{}
 \end{subfigure}\hspace{5pt}
  \begin{subfigure}[b] {0.3\textwidth}
   \centering
   \includegraphics[width=\textwidth]{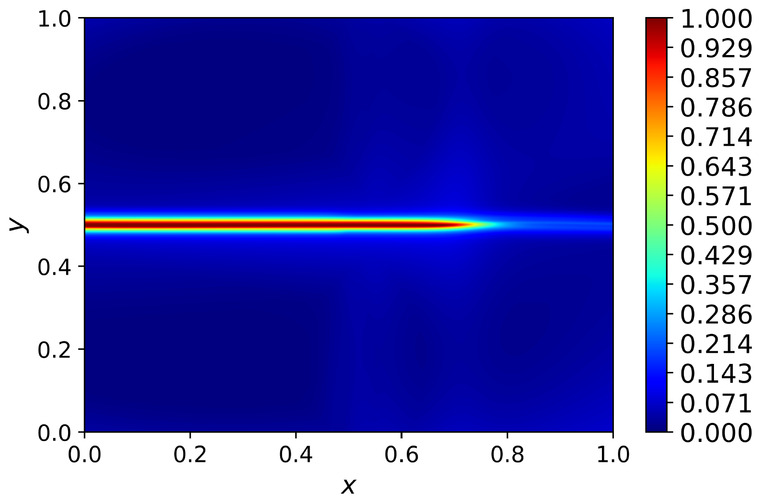}\hfill
   \caption{}
 \end{subfigure}\hspace{5pt}
  \begin{subfigure}[b] {0.3\textwidth}
   \centering
   \includegraphics[width=\textwidth]{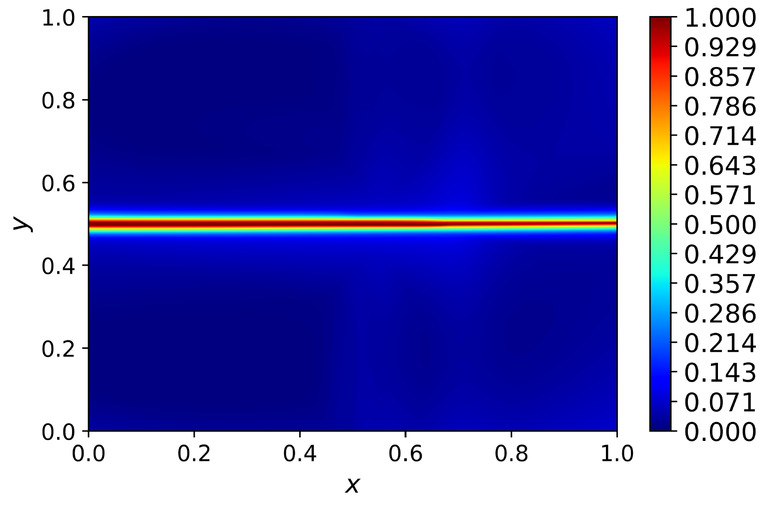}\hfill
   \caption{}
 \end{subfigure}\hspace{5pt}
 \caption{Crack pattern for prescribed displacement of (a) $1\times10^{-3}$, (b) $2\times10^{-3}$, (c) $3\times10^{-3}$, (d) $4\times10^{-3}$ and (e) $5\times10^{-3}$.}
 \label{fig:phasefieldPlots}
 \end{figure}

To observe the propagation of crack in the plate, we simultaneously solve the elastic field and the phase field using a monolithic solver. The minimization problem reads as stated in \autoref{eq:energyterms}. The Dirichlet boundary conditions are:
\begin{equation}
    u(0,y) = v(x,0) = 0, \;\;\; v(x,1)= \Delta v,
\end{equation}
where $u$ and $v$ are the solutions of the elastic field in $\it{x}$ and $\it{y}$-axis-axis. The computation is performed by applying constant displacement increments of $\Delta v$ = $1\times 10^{-3}$ mm. The model for the Dirichlet problem is:
\begin{equation}
\begin{split}
    u &= [x(1-x)]\hat{u},\\
    v &= [y(y-1)]\hat{v} + y\Delta v, 
\end{split}
\end{equation}
where $\hat{u}$ and $\hat{v}$ are obtained from the neural network. We have used a network with 5 hidden layers of $50$ neurons each. The loss function, $\mathcal{L}_{\mathcal{E}}$ is defined as:
\begin{equation}\label{eq:phasefield2D_loss}
\begin{split}
        \mathcal{L}_{\mathcal{E}} &= \mathcal{L}_{Elas} + \mathcal{L}_{PF},\\
        \mathcal{L}_{Elas} & = \frac{1}{N_x\times N_y}\left(A_{tc}\sum_{i = 1}^{N_x\times N_y}f_e(\bm{x}_i^{tc}) + A_{c}\sum_{i = 1}^{N_x\times N_y}f_e(\bm{x}_i^{c}) + A_{bc}\sum_{i = 1}^{N_x\times N_y}f_e(\bm{x}_i^{bc})\right),\\
        \mathcal{L}_{PF} & = \frac{1}{N_x\times N_y}\left(A_{tc}\sum_{i = 1}^{N_x\times N_y}f_c(\bm{x}_i^{tc}) + A_{c}\sum_{i = 1}^{N_x\times N_y}f_c(\bm{x}_i^{c}) + A_{bc}\sum_{i = 1}^{N_x\times N_y}f_c(\bm{x}_i^{bc})\right),\\
        f_e(\bm{x}) &= g(\phi(\bm{x}))\Psi_{0}(\bm{\epsilon(x)}),\\
        f_c(\bm{x}) &= \frac{G_c}{2l_0}\left(\phi(\bm{x})^2 + l_0^2|\nabla\phi(\bm{x})|^2\right)  + g'(\phi(\bm{x}))H(\bm{x},t),\\
\end{split}
\end{equation}
where $\mathcal{L}_{Elas}$ and $\mathcal{L}_{PF}$ define the losses in the elastic strain energy and the fracture energy, respectively. The scatter plots of the deformed configuration and the corresponding phase field plots for the simulation are shown in \autoref{fig:crackflow} and \autoref{fig:phasefieldPlots}, respectively.

In the next section, we apply DEM on an electro-mechanically coupled systems.

\subsection{Piezoelectricity}
\label{subsec:piezo}
The electro-mechanical coupling is generated in some crystal materials where an electrical polarization, $\textbf{P}$ is induced due to the existence of the mechanical strain or the existence of an electrical field can cause deformation of the geometry. 
The linear dependence between the electric polarization and the associated  mechanical strain refers to the piezoelectric effect which is common in non-centrosymmetric crystals.
The direct and the indirect piezoelectric effect can be expressed in mathematical form of as
\begin{linenomath}
\begin{align}
\textbf{P}&= \textbf{p}:\bm{\epsilon}, \nonumber \\
\bm{\sigma} &= \textbf{p}\cdot \bm{E},
\end{align}
\end{linenomath}
where $\textbf{p}$ refers to the third order piezoelectric tensor, $\bm{\epsilon}$ and $\bm{\sigma}$ are the second order strain and stress tensors, and $\bm{E}$ is the electric field vector with $E_i=-\theta_{,i}$ and $\theta$ is the electric potential.
It can be concluded that the piezoelectricity is a cross coupling between the elastic and the dielectric variables
Hence, the constitutive equations to identify the coupling between the mechanical stress and the strain with the related electric field and displacement are given by
\begin{linenomath}
\begin{align}
\bm{D} = \bm{e}:\bm{\epsilon}+ \bm{\kappa}\cdot \bm{E}, \nonumber \\
\bm{\sigma}= \mathbb{C}:\bm{\epsilon}-\bm{e}\cdot \bm{E},
\label{Eq:cons}
\end{align}
\end{linenomath}
where $\bm{D}$ refers to the electric displacement that causes the electrical polarization, while $\bm{e},~\bm{\kappa},~\mathbb{C}$  are the piezoelectric, the dielectric, and the elastic tensors, respectively. 

Energy minimization is adopted in the DEM to obtain the solution of the field variables. The problem statement reads as 
\begin{equation}\label{Eq:pi}
\begin{split}
    \text{Minimize:}\;\;\;\; \mathcal{E} &= \mathcal{E}_i + W_{ext},\\
    \text{subject to:}\;\;\;\; \bm{u} &= \bm{\overline u} \text{ on } \partial \Omega_{D_u},\\
    \text{and}\;\;\;\; \theta &= \overline{\theta} \text{ on } \partial \Omega_{D_{\theta}},\\
    \text{and}\;\;\;\; \bm{\sigma}\cdot \bm{n} &= \bm{t}_N \text{ on } \partial\Omega_{N}, \\
\end{split}
\end{equation}
where $\mathcal{E}$ represents the total energy of the system, $\mathcal{E}_i$ is the bulk internal energy, whose density is $\Psi$, and $W_{ext}$ stands for the work of external forces.
The internal energy density, $\Psi$, is expressed as
\begin{linenomath}
\begin{equation}
\Psi = \frac{1}{2}\bm{\epsilon}:\mathbb{C}:\bm{\epsilon} - \bm{E}\cdot\bm{e}\cdot \bm{E} -  \frac{1}{2}\bm{E}\cdot\bm{\kappa}\cdot \bm{E}.
\end{equation}
\end{linenomath}
We shall now present the example of a cantilever beam to show the application of DEm to solve PDEs involving electromechanical coupling.

\subsubsection{Cantilever beam}
\label{subsec5:piezo1}
\begin{figure}[t]
\centering
\tikzset{every picture/.style={line width=1.0pt}} 
\begin{tikzpicture}[x=0.75pt,y=0.75pt,yscale=-.8,xscale=.8]
\draw    (20,35) -- (20.,125) ;
\draw    (20,40) -- (8.3,50.67) ; \draw    (20,50) -- (8.3,60.67) ; 
\draw    (20,60) -- (8.3,70.67) ; \draw    (20,70) -- (8.3,80.67) ;
\draw    (20,80) -- (8.3,90.67) ; \draw    (20,90) -- (8.3,100.67) ;
\draw    (20,100) -- (8.3,110.67) ; \draw    (20,110) -- (8.3,120.67) ;
\draw    (20,120) -- (8.3,130.67) ; 
\draw   (20,50) -- (330,50) -- (330,110) -- (20,110) -- cycle ;
\draw [line width=0.5]      (38,137.57) -- (108.3,137.57)(45.03,79.67) -- (45.03,144) (101.3,132.57) -- (108.3,137.57) -- (101.3,142.57) (40.03,86.67) -- (45.03,79.67) -- (50.03,86.67)  ;
\draw    (164,140) -- (176,140) ; \draw    (159,135) -- (181,135) ;
\draw    (154,130) -- (186,130) ;
\draw    (170,110) -- (170,130) ;
\draw    (170,19.34) -- (170,50) ; \draw    (170,19.34) -- (221,19.34) ;
\draw   (221,19.34) .. controls (221,9.03) and (229.36,0.67) .. (239.66,0.67) .. controls (249.97,0.67) and (258.33,9.03) .. (258.33,19.34) .. controls (258.33,29.64) and (249.97,38) .. (239.66,38) .. controls (229.36,38) and (221,29.64) .. (221,19.34) -- cycle ;
\draw    (330,10) -- (330,50) ;
\draw [shift={(330,50)}, rotate = 270]    (10.93,-3.29) .. controls (6.95,-1.4) and (3.31,-0.3) .. (0,0) .. controls (3.31,0.3) and (6.95,1.4) .. (10.93,3.29)   ;

\draw (120,138) node [scale=0.9]  {$x$}; \draw (38,93) node [scale=0.9]  {$y$};
\draw (238,20) node [scale=1.1] [align=left] {V};  \draw (340,15) node [scale=1.1] [align=left] {F};
\draw (238,-8) node [scale=0.9] [align=left] {Electrode};
\draw (330,0) node [scale=0.9] [align=left] {Applied force};
\end{tikzpicture}
 \caption{
 Schematic diagram showing the electrical and the mechanical boundary conditions  for the cantilever nanobeam under study.}\label{Fig:BCs}
\end{figure}
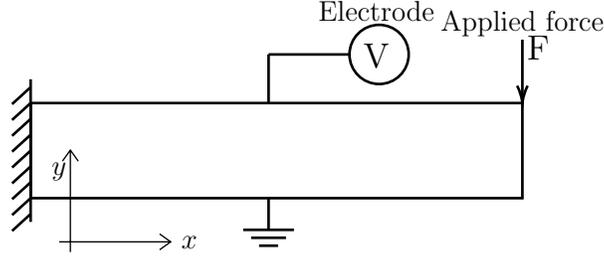

In this example, a cantilever beam configuration (the most common problem in evaluating the effect of piezoelectricity) is solved under plane strain conditions. The material is assumed to be isotropic and linearly elastic. The geometrical setup and the boundary conditions are shown in \autoref{Fig:BCs}. A fully connected network with three hidden layers of 150 neurons each is used to solve the electro-mechanically coupled problem. For the first two layers, we have considered hyperbolic 
tangent, \texttt{tanh} activation function; whereas for the last layer, 
linear activation function has been considered. To optimize the hyper-parameters, the loss function is calculated similar to the examples discussed in the previous sections. For training, we have used the ADAM optimizer followed by L-BFGS.

\begin{figure}
\centering
\begin{subfigure}{0.58\linewidth}  
  \includegraphics[width=7cm]{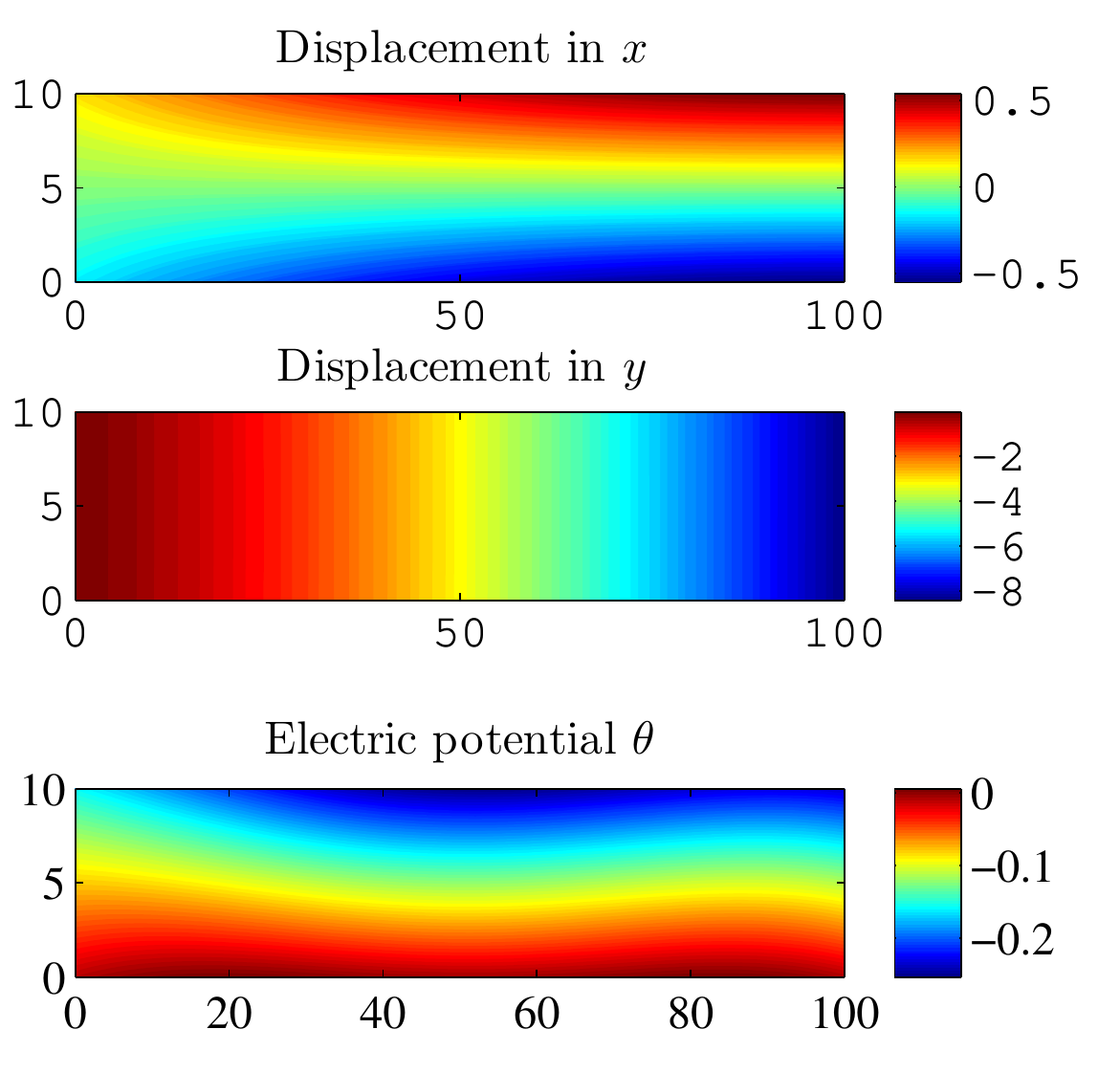}   
   \caption{}\label{Fig:DEM_F}
 \end{subfigure}%
 \begin{subfigure}{0.55\linewidth}  
  \includegraphics[width=7cm]{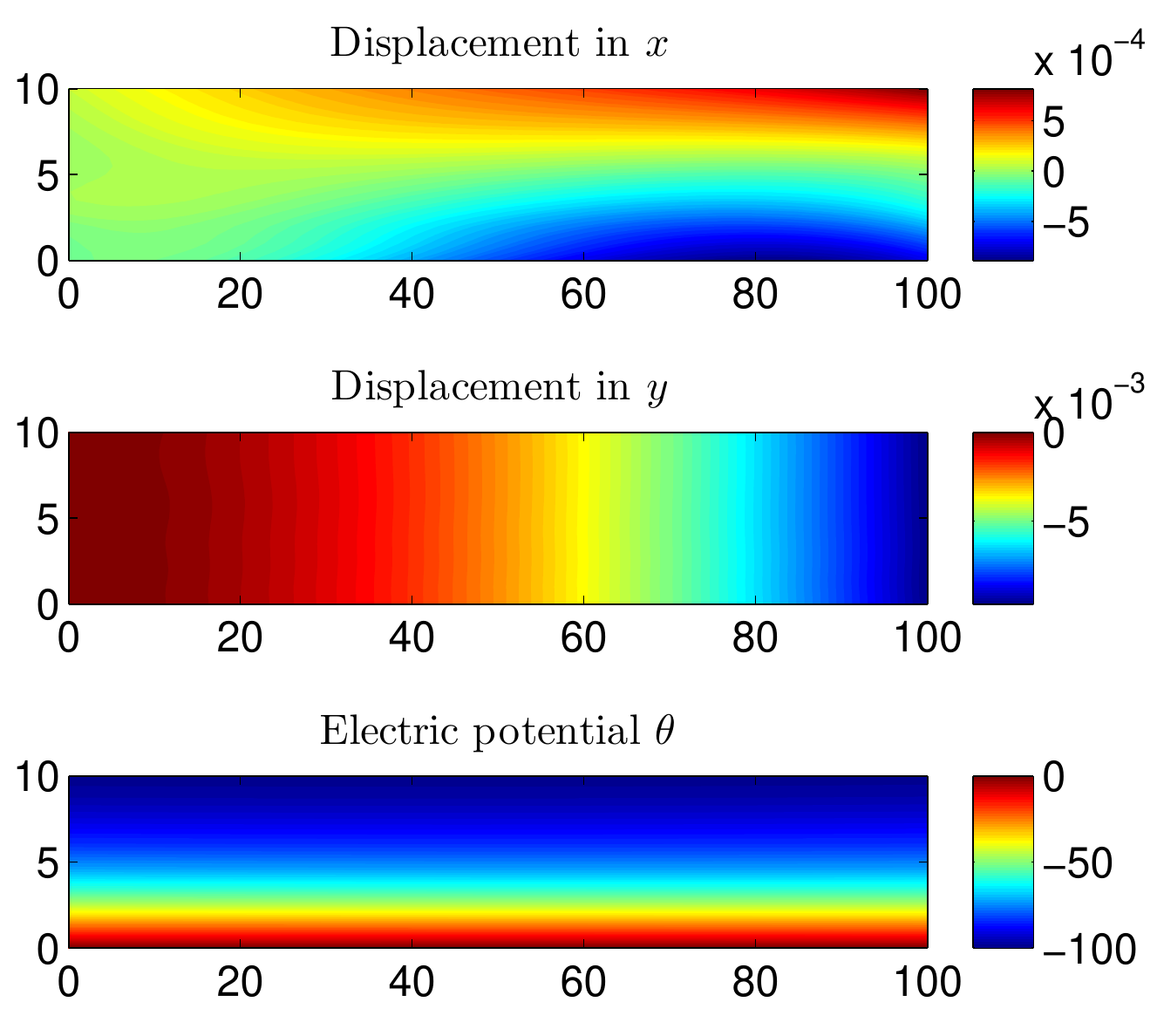}   
   \caption{}\label{Fig:DEM_E}
 \end{subfigure}    
\caption{The predicted values of three outputs of the system using DEM: (a) applying mechanical loading, and (b) applying electrical loading.}
\end{figure}

In this problem, we investigate two loading conditions: 
\begin{enumerate}
    \item pure mechanical loading
    \item pure electrical loading
\end{enumerate}
In the case of pure mechanical loading, a point load is applied as shown in \autoref{Fig:BCs}. Due to the electro-mechanical coupling, an electrical potential, $\theta$ is induced by the mechanical load. Assuming closed circuit configuration, $\theta$ is fixed to zero on the bottom surface while the electrode placed on the top surface undergoes a difference in electric potential as a result of the deformation. \autoref{Fig:DEM_E} shows the plots of the deformed shape and the electric potential obtained using the DEM approach. 

In the next step, we consider pure electrical loading. In this case, the electric potential on the top surface is fixed to $V=-100$ MV. The defomed configuration is presented in \autoref{Fig:DEM_E}. Both the cases considered in this example, demonstrate the ability of the DEM  to solve electro-mechanically coupled system without the aid of the classical numerical methods.

All the examples discussed previously involved second-order PDEs. To show the robustness of DEM, we take up the study of a fourth-order PDE in the next section.

\subsection{Kirchhoff Plate bending}
\label{subsec:kirchoff}
The Kirchhoff plate bending problem is a classical problem governed by a fourth-order PDE. The weak formulation of the problem involves the second-order derivatives of the transversal deflection. In traditional mesh-based methods, the essential requirement of $C^{1}$ continuity poses significant challenges. This, however, can be easily solved by the proposed deep collocation and the deep energy methods with a set of DNNs approximating the transversal deflection, which are proven to be effective in the bending analysis of Kirchhoff plate of various geometries and even with cut-outs.

The governing equation of Kirchhoff bending problem, expressed in terms of transversal deflection, is  
\begin{equation}
\bigtriangledown^{2}\left ( \bigtriangledown^{2}w \right )=\bigtriangledown^{4}w=\frac{p}{D},
\label{eq:governing_kir}
\end{equation}
where $\bigtriangledown^{4}\left (\cdot\right) =\frac{\partial^4 }{\partial x^4}+2\frac{\partial^4 }{\partial x^2\partial y^2}+\frac{\partial^4 }{\partial y^4}$ is commonly referred to as biharmonic operator. Similar to the previous examples, the problem statement in strong form is equivalent the minimization of the total potential energy of the system, $\mathcal{E}$, where
\begin{equation}
\mathcal{E}=\iint_{\Omega }\left (\frac{1}{2}\textit{\textbf{k}}^{T}\textit{\textbf{M}}-qw  \right )d\Omega-\int_{S_{3}}\bar{V}_{n}wdS+\int_{S_{2}+S_{3}}\bar{M}_{n}\frac{\partial w}{\partial n}dS	
\label{plate_energy}
\end{equation}
under uniformly distributed load. The problem is stated as:
\begin{equation}
w=\underset{v\ddot{\in H\left ( D \right )}}{argmin}\, \mathcal{E}\left ( v \right )
\end{equation}	
Additionally, the boundary conditions of the Kirchhoff plate taken into consideration can be classified into three types; simply-supported, clamped and free boundary conditions and is expressed as
\begin{equation}
\partial\Omega =\Gamma_{1}+\Gamma_{2}+\Gamma_{3}.
\label{boundary}
\end{equation}

To solve Kirchhoff bending problems with boundary constraints, the physical domain is first discretized with randomly distributed collocation points denoted by $\textit{\textbf{x}}\,_\Omega=(x_1,\ldots,x_{N_\Omega})^T$. Another set of collocation points are added to discretize the boundaries denoted by $\textit{\textbf{x}}\,_\Gamma(x_1,\ldots,x_{N_\Gamma})^T$.
The transversal deflection, $w$ can thus be approximated with the aforementioned deep feed-forward neural network $\textit{w}^h \left(\textit{\textbf{x}};\boldsymbol{p}\right)$. The loss function is constructed to find the approximate solution by minimizing either the mean squared error of the governing equation in residual form or the total energy, along with boundary conditions approximated by $\textit{w}^h \left(\textit{\textbf{x}};\boldsymbol{p}\right)$. 

 Substituting $\textit{w}^h \left(\textit{\textbf{x}}\,_\Omega;\boldsymbol{p}\right)$ into either Equation \ref{eq:governing_kir} or into \ref{plate_energy}, a physical informed deep neural network is obtained: $G\left(\textit{\textbf{x}}\,_\Omega;\boldsymbol{p}\right)$ for the strong form of the problem, and $\mathcal{E}_{p}\left(\textit{\textbf{x}}\,_\Omega;\boldsymbol{p}\right)$ the energy approach, respectively.

The boundary conditions can also be expressed by the physical informed neural network approximation $\textit{w}^h \left(\textit{\textbf{x}}\,_\Gamma;\boldsymbol{p}\right)$:

\noindent On clamped boundaries $\Gamma_{1}$, we have
\begin{equation}
  \textit{w}^h \left(\textit{\textbf{x}}\,_{\Gamma_1};\boldsymbol{p}\right)=\tilde{w}, \  \frac{\partial \textit{w}^h \left(\textit{\textbf{x}}\,_{\Gamma_1};\boldsymbol{p}\right)}{\partial n} = \tilde{ \boldsymbol{p}}_{n}.
\end{equation} 

\noindent On simply-supported boundaries $\Gamma_{2}$,
\begin{equation}
  \textit{w}^h \left(\textit{\textbf{x}}\,_{\Gamma_2};\boldsymbol{p}\right)=\tilde{w}, \ \tilde{ M}_{n}\left(\textit{\textbf{x}}\,_{\Gamma_2};\boldsymbol{p}\right)=\tilde{ M}_{n}, 
\label{bd2}
\end{equation} 

\noindent On free boundaries $\Gamma_{3}$,
\begin{equation}
  M_{n}\left(\textit{\textbf{x}}\,_{\Gamma_3};\boldsymbol{p}\right) =\tilde{ M}_{n} , \  \frac{\partial M_{ns}\left(\textit{\textbf{x}}\,_{\Gamma_3};\boldsymbol{p}\right) }{\partial s}+Q_{n}\left(\textit{\textbf{x}}\,_{\Gamma_3};\boldsymbol{p}\right)=\tilde{q}, 
\end{equation}  
It should be noted that $\textit{\textbf{n}},\textit{\textbf{s}}$ here refer to the normal and tangent directions along each boundary.

It is clear that all induced physical informed neural network in Kirchhoff bending analysis  $G\left(\textit{\textbf{x}};\boldsymbol{p}\right)$, $\mathcal{E}\left(\textit{\textbf{x}}\,_\Omega;\boldsymbol{p}\right)$, $M_{n}\left(\textit{\textbf{x}};\boldsymbol{p}\right)$, $M_{ns}\left(\textit{\textbf{x}};\boldsymbol{p}\right)$, $Q_{n}\left(\textit{\textbf{x}};\boldsymbol{p}\right)$  share the same parameters as $\textit{w}^h \left(\textit{\textbf{x}};\boldsymbol{p}\right)$. Considering the generated collocation points in domain and on boundaries, they can all be learned by minimizing the mean square error loss function:
\begin{equation}
L\left(\boldsymbol{p}\right)=MSE=MSE_{G}+MSE_{\Gamma_{1}}+MSE_{\Gamma_{2}}+MSE_{\Gamma_{3}},
\end{equation}
with
\begin{equation}
\begin{aligned}
&MSE_{G}=\frac{1}{N_d}\sum_{i=1}^{N_d}\begin{Vmatrix}
G\left(\textit{\textbf{x}}\,_\Omega;\boldsymbol{p}\right)
\end{Vmatrix}^2=\frac{1}{N_\Omega}\sum_{i=1}^{N_\Omega}\begin{Vmatrix}
\bigtriangledown^{4}w^h\left(\textit{\textbf{x}}\,_\Omega;\boldsymbol{p}\right)-\frac{p}{D}
\end{Vmatrix}^2,\\
&MSE_{\Gamma_{1}}=\frac{1}{N_{\Gamma_1}}\sum_{i=1}^{N_{\Gamma_1}}\begin{Vmatrix}
 \textit{w}^h \left(\textit{\textbf{x}}\,_{\Gamma_1};\boldsymbol{p}\right)-\tilde{w}
\end{Vmatrix}^2+\frac{1}{N_{\Gamma_1}}\sum_{i=1}^{N_{\Gamma_1}}\begin{Vmatrix}
\frac{\partial \textit{w}^h \left(\textit{\textbf{x}}\,_{\Gamma_1};\boldsymbol{p}\right)}{\partial n} - \tilde{ \boldsymbol{p}}_{n}
\end{Vmatrix}^2,\\
&MSE_{\Gamma_{2}}=\frac{1}{N_{\Gamma_2}}\sum_{i=1}^{N_{\Gamma_2}}\begin{Vmatrix}
 \textit{w}^h \left(\textit{\textbf{x}}\,_{\Gamma_2};\boldsymbol{p}\right)-\tilde{w}
\end{Vmatrix}^2+\frac{1}{N_{\Gamma_2}}\sum_{i=1}^{N_{\Gamma_2}}\begin{Vmatrix}
\tilde{ M}_{n}\left(\textit{\textbf{x}}\,_{\Gamma_2};\boldsymbol{p}\right)-\tilde{ M}_{n}
\end{Vmatrix}^2,\\
&MSE_{\Gamma_{3}}=\frac{1}{N_{\Gamma_3}}\sum_{i=1}^{N_{\Gamma_3}}\begin{Vmatrix}
 \tilde{ M}_{n}\left(\textit{\textbf{x}}\,_{\Gamma_3};\boldsymbol{p}\right)-\tilde{ M}_{n}
\end{Vmatrix}^2+\frac{1}{N_{\Gamma_3}}\sum_{i=1}^{N_{\Gamma_3}}\begin{Vmatrix}
\frac{\partial M_{ns}\left(\textit{\textbf{x}}\,_{\Gamma_3};\boldsymbol{p}\right) }{\partial s}+Q_{n}\left(\textit{\textbf{x}}\,_{\Gamma_3};\boldsymbol{p}\right)-\tilde{q}
\end{Vmatrix}^2,\\
\end{aligned}
\end{equation}
where $x\,_\Omega \in {\mathbb{R}^N} $, $\boldsymbol{p} \in {\mathbb{R}^K}$ are the neural network parameters. $L\left(\boldsymbol{p}\right)=0$, $\textit{w}^h \left(\textit{\textbf{x}};\boldsymbol{p}\right)$ is then a solution to transversal deflection.
Our goal is to find a set of parameters $\boldsymbol{p}$ such that the  approximated deflection $\textit{w}^h \left(\textit{\textbf{x}};\boldsymbol{p}\right)$ minimizes the loss $L\left(\boldsymbol{p}\right)$. If $L\left(\boldsymbol{p}\right)$ is a very small value, the approximation $\textit{w}^h \left(\textit{\textbf{x}};\boldsymbol{p}\right)$ is very closely satisfying governing equations and boundary conditions, namely
\begin{equation}
\boldsymbol{p}= \underset{\hat{\boldsymbol{p}} \in \mathbb{R}^K}{\operatorname{argmin}} \, L\left(\hat{\boldsymbol{p}}\right)
\end{equation}

However, in order to solve the problem of a thin plate with the deep energy method, it is necessary to construct the loss function in a different way:

\begin{equation}
\begin{matrix}
\mathcal{L}\left (\boldsymbol{p}\right ) =MAE_{\mathcal{E}}+MSE_{\Gamma_{1}}+MSE_{\Gamma_{2}},
\end{matrix}
\end{equation}
Here, only due to the fact that energy method is adopted, only Dirichlet boundary conditions among three boundary types need to be taken into account. This makes it easier than the deep collocation method. The latter needs a physical discovery of Neumann boundary conditions with DNN. In this case, as shown above, the twist moment and shear force along the boundaries need to be approximated with the physical informed deep neural networks.

The plate bending problems with DNN based method can be accordingly reduced to an optimization problem. In the deep learning Tensorflow framework, a variety of optimizers are available to help to gain an optimal solution, and the Adam and LBFGS optimizers are mainly adopted in numerical examples. 

Next, a series of numerical examples are tested to verify the feasibility of deep collocation method and deep energy method in the analysis of Kirchhoff plate bending problems.

\subsubsection{Simply-supported square plate on Winkler foundation}
\label{subsec6:kirchoff_prob1}
The deep collocation method is first applied to study the bending problems of various plates, which are chosen with analytical or exact solutions as comparisons.

To begin with, we consider a simply-supported square plate resting on Winkler foundation, which assumes that the foundation's reaction $p\left(x,y\right)$ can be described by $p\left(x,y\right)=\textit{k}w$, $\textit{k}$ being a constant called $foundation\;modulus$. For a plate on a continuous Winkler foundation, the governing Equation \ref{eq:governing_kir} can be written as
\begin{equation}
\bigtriangledown^{2}\left ( \bigtriangledown^{2}w \right )=\bigtriangledown^{4}w=\frac{\left(p-q\right)}{D}=\frac{\left(p-\textit{k}w\right)}{D}
\label{governingelastfound}
\end{equation}
Here, $D$ denotes the flexural stiffness of the plate depending on the plate thickness and material properties.   

The analytical solution for this numerical example is \cite{timoshenko1959theory}:
\begin{equation}
w=\frac{16p}{ab}\sum_{m=1,3,5,\cdots  }^{\infty}\sum_{n=1,3,5,\cdots  }^{\infty}\frac{\textrm{sin}\frac{m\pi x }{a} \textrm{sin}\frac{n\pi y }{b}}{mn\left [ \pi^4D\left ( \frac{m^2}{a^2} +\frac{n^2}{b^2}\right )^2+\textit{k} \right ]}
\end{equation}   

\begin{figure}
 \centering
 \begin{subfigure}[b] {0.42\linewidth}
   \centering
   \includegraphics[width=\linewidth]{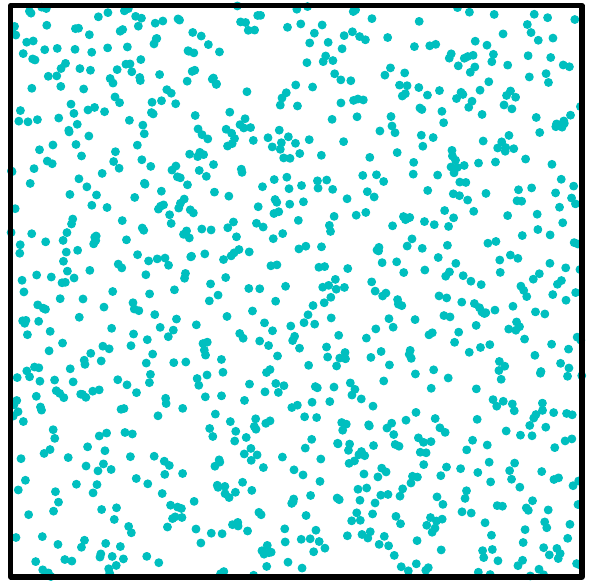}\hfill
   \caption{Collocation points in the square domain.}
 \end{subfigure}\hspace{5pt}
\begin{subfigure}[b] {0.56\linewidth}
   \centering
   \includegraphics[width=\linewidth]{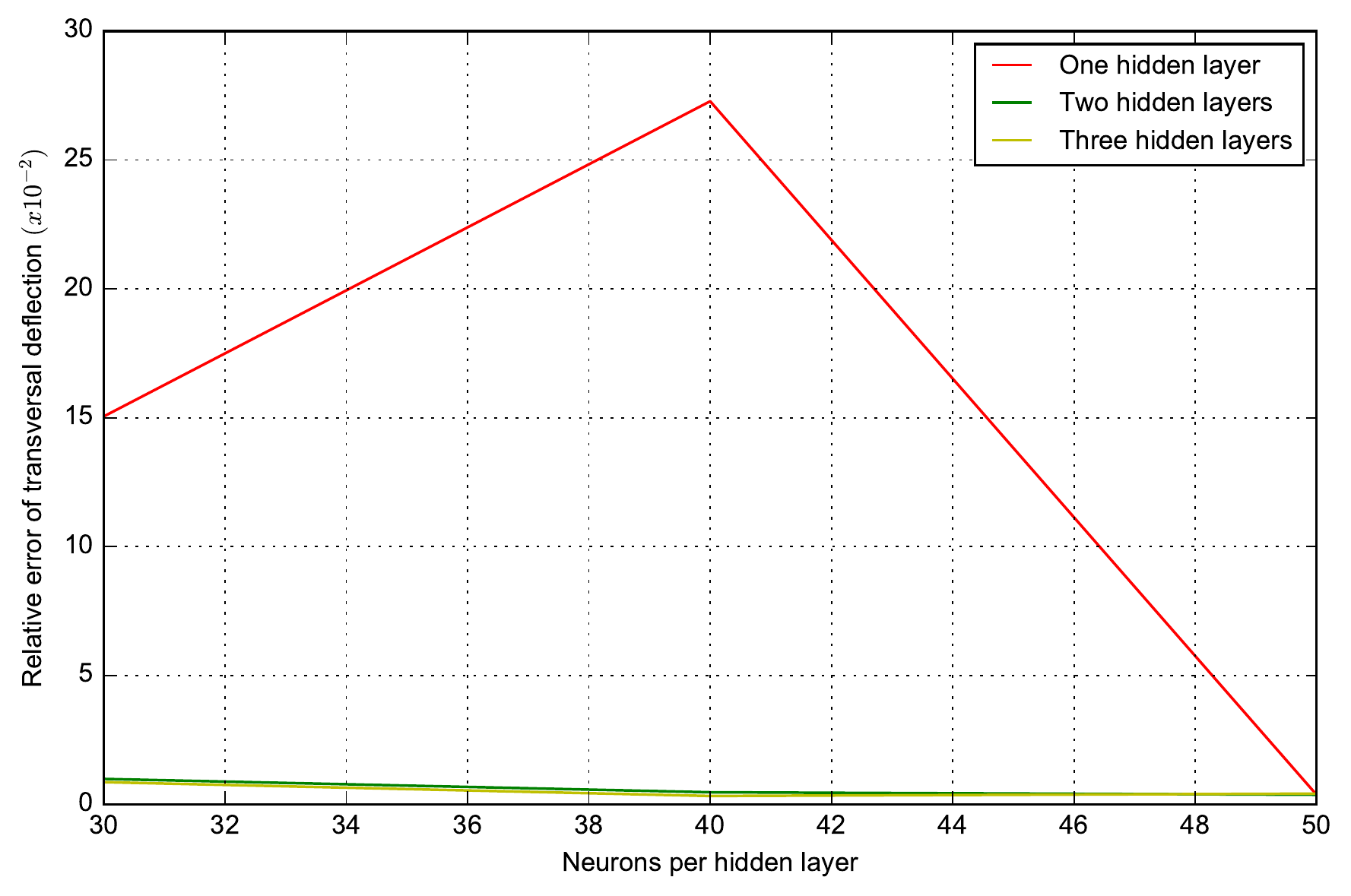}\hfill
   \caption{The relative error of deflection with varying hidden layers and neurons.}
 \end{subfigure}
 \caption{}
 \label{Scatterpoint}
 \end{figure}

\begin{figure}
\centering
\caption*{Table 1: Maximum deflection predicted by deep collocation method.}
\includegraphics[width=10cm]{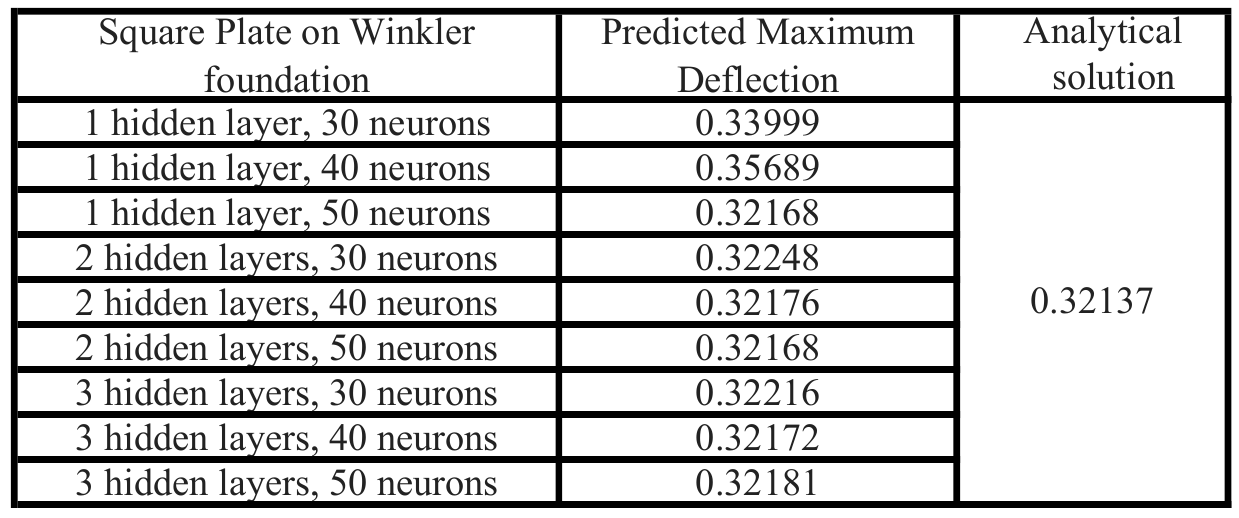}
\label{TableWK}
\end{figure}

\begin{figure}
 \centering
 \begin{subfigure}[b] {0.49\textwidth}
   \centering
   \includegraphics[width=\textwidth]{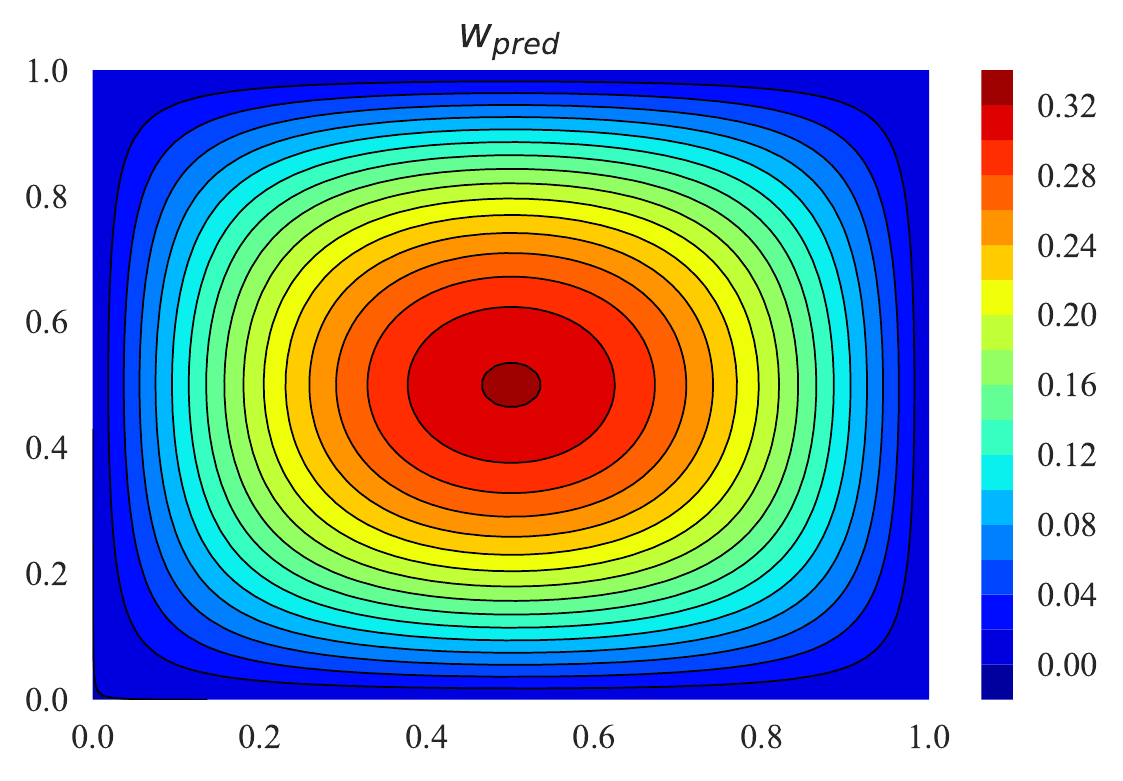}\hfill
   \caption{}\label{fig:contourpredssef}
 \end{subfigure}\hspace{5pt}
 \begin{subfigure}[b] {0.49\textwidth}
   \centering
   \includegraphics[width=\textwidth]{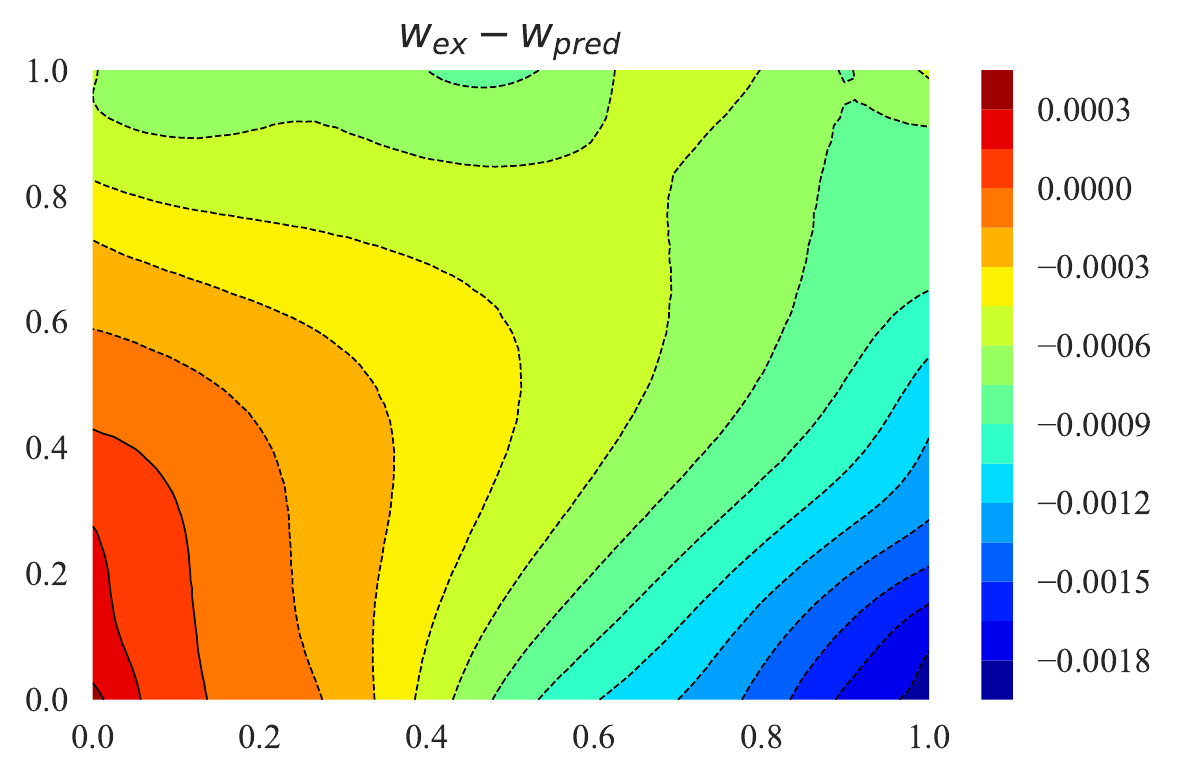}\hfill
   \caption{}\label{fig:errocontourpredssef}
 \end{subfigure}\hspace{5pt}
  \begin{subfigure}[b] {0.49\textwidth}
   \centering
   \includegraphics[width=\textwidth]{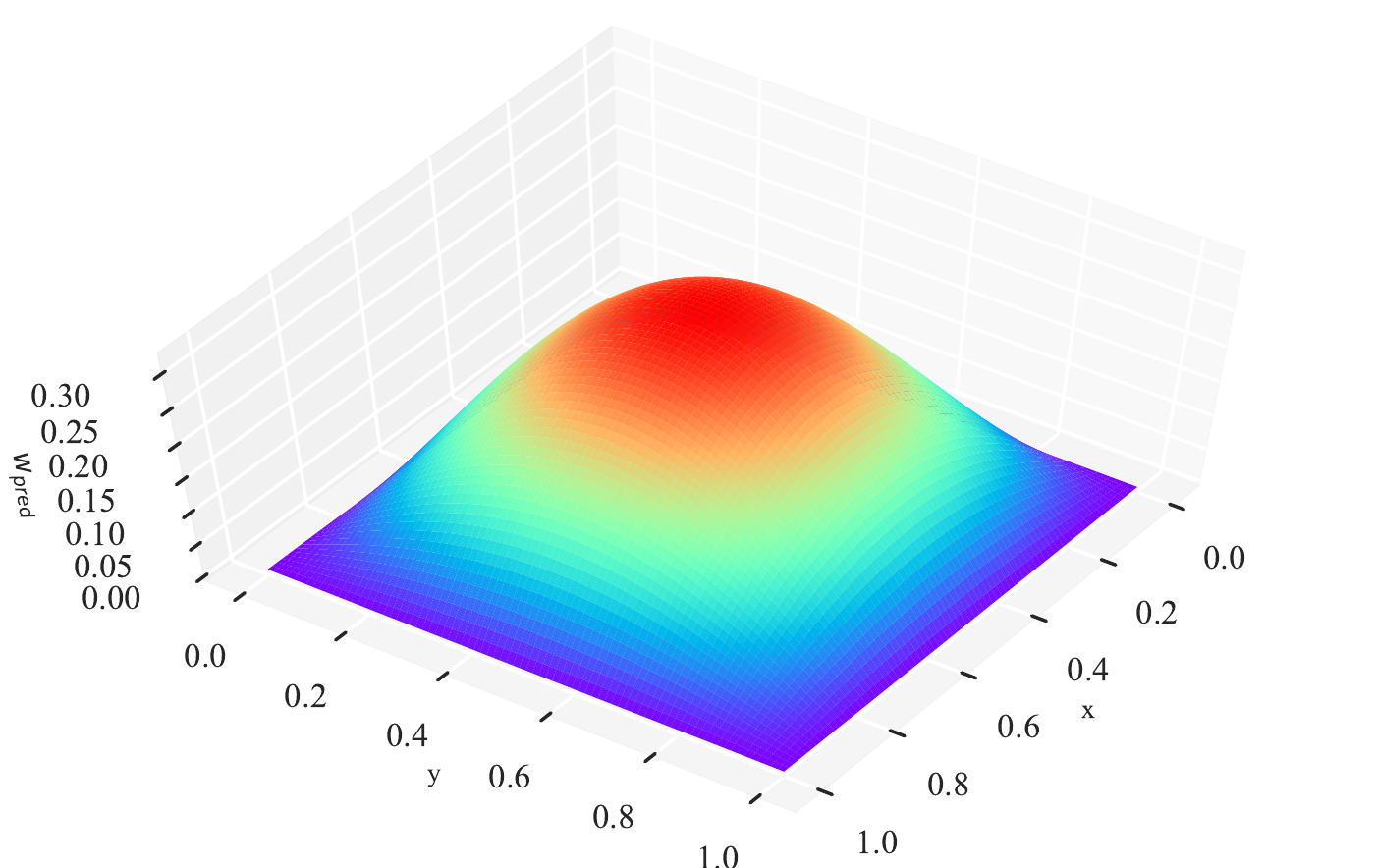}\hfill
   \caption{}\label{fig:Dflpredssef}
 \end{subfigure}\hspace{1pt}
 \begin{subfigure}[b] {0.49\textwidth}
   \centering
   \includegraphics[width=\textwidth]{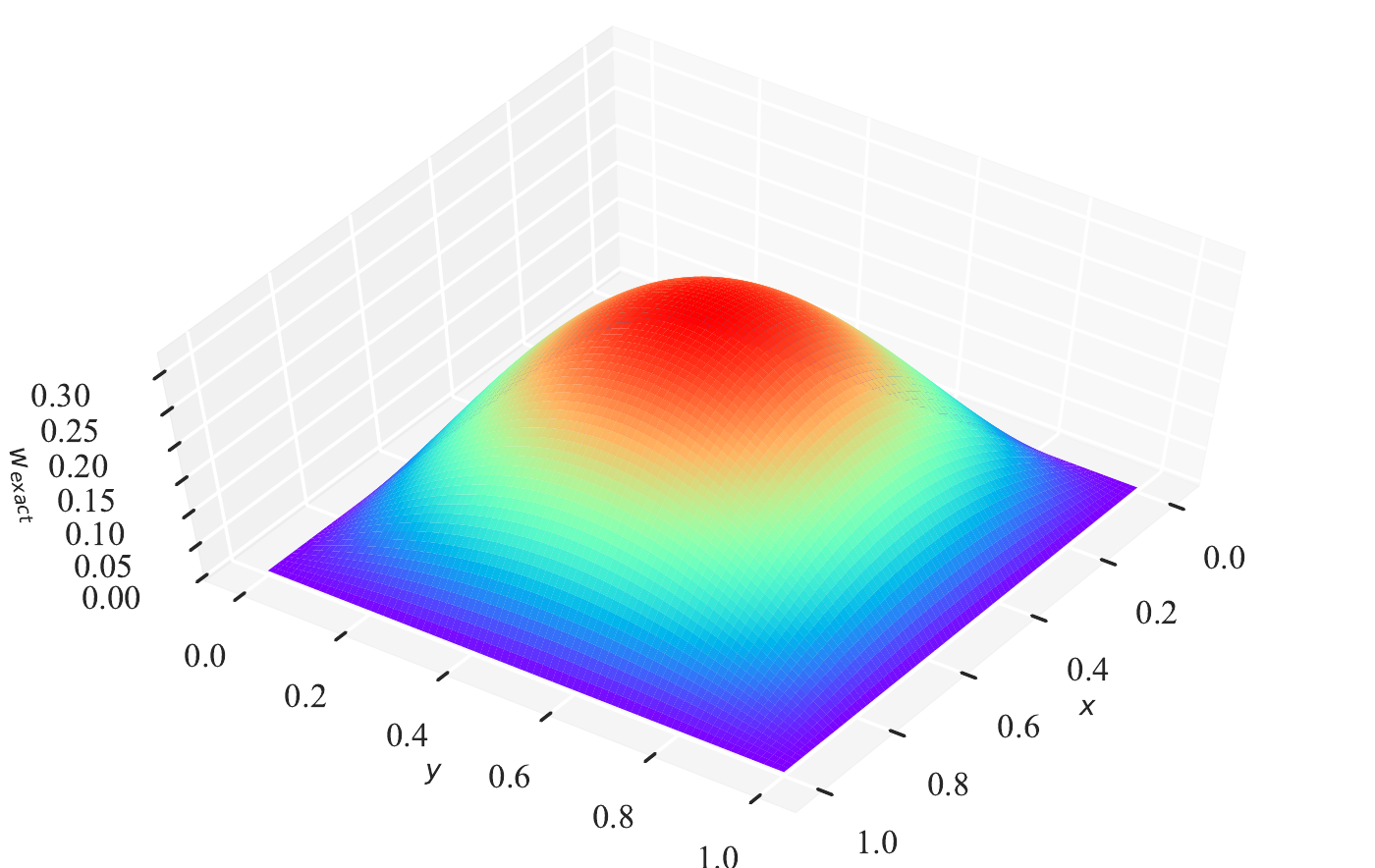}\hfill
   \caption{}\label{fig:Dflexactssef}
 \end{subfigure}\hspace{1pt}
 \caption{$\left(a\right)$ Predicted deflection contour $\left(b\right)$ Deflection error contour $\left(c \right)$ Predicted deflection $\left(d \right)$ Analytical deflection of the simply-supported plate on Winkler foundation with 3 hidden layers and 50 neurons.}
 \label{esneuron50point1}
\end{figure}

For this numerical example, the arrangement of collocation points is depicted in Figure~\ref{Scatterpoint}(a). Neural networks with different neurons and depth are applied in the calculation to study the feasibility of deep collocation in solving Kirchhoff plate bending problems. Table 1 lists the maximum deflection at the central point with varying hidden layers and neurons. It is clear that the results predicted by more hidden layers are more desirable and as hidden layer and neuron number grows, the maximum deflection becomes more accurate approaching the analytical serial solution for even two hidden layers. The relative error shown in Figure~\ref{Scatterpoint}(b) better depicts the advantages of deep neural network than shallow wide neural network. More hidden layers, with more neurons yield flatting of the relative error. Various contour plots are shown in Figure~\ref{esneuron50point1} and compared with the analytical solution. It is obvious that the deep collocation method predict the deflection of the whole plate in good accordance with the analytical solution.

\subsubsection{Clamped circular plate}
\label{subsec6:kirchoff_prob2}
\begin{figure}
 \centering
 \begin{subfigure}[b] {0.41\textwidth}
   \centering
   \includegraphics[width=\textwidth]{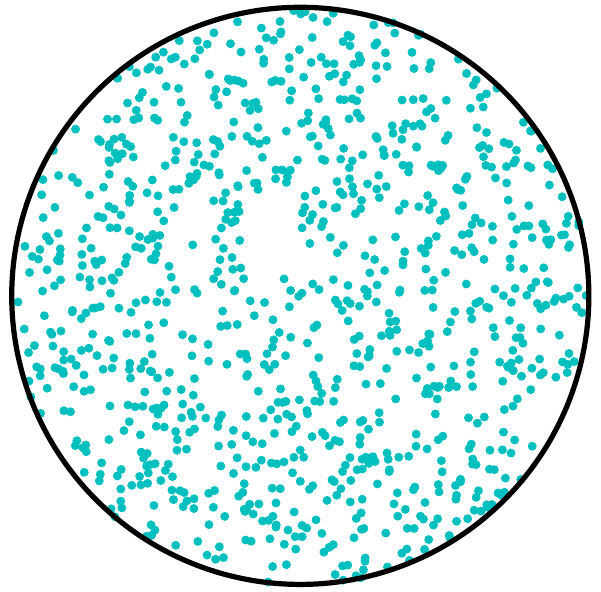}\hfill
   \caption{Collocation points in the circular domain}
 \end{subfigure}\hspace{5pt}
\begin{subfigure}[b] {0.57\textwidth}
   \centering
   \includegraphics[width=\textwidth]{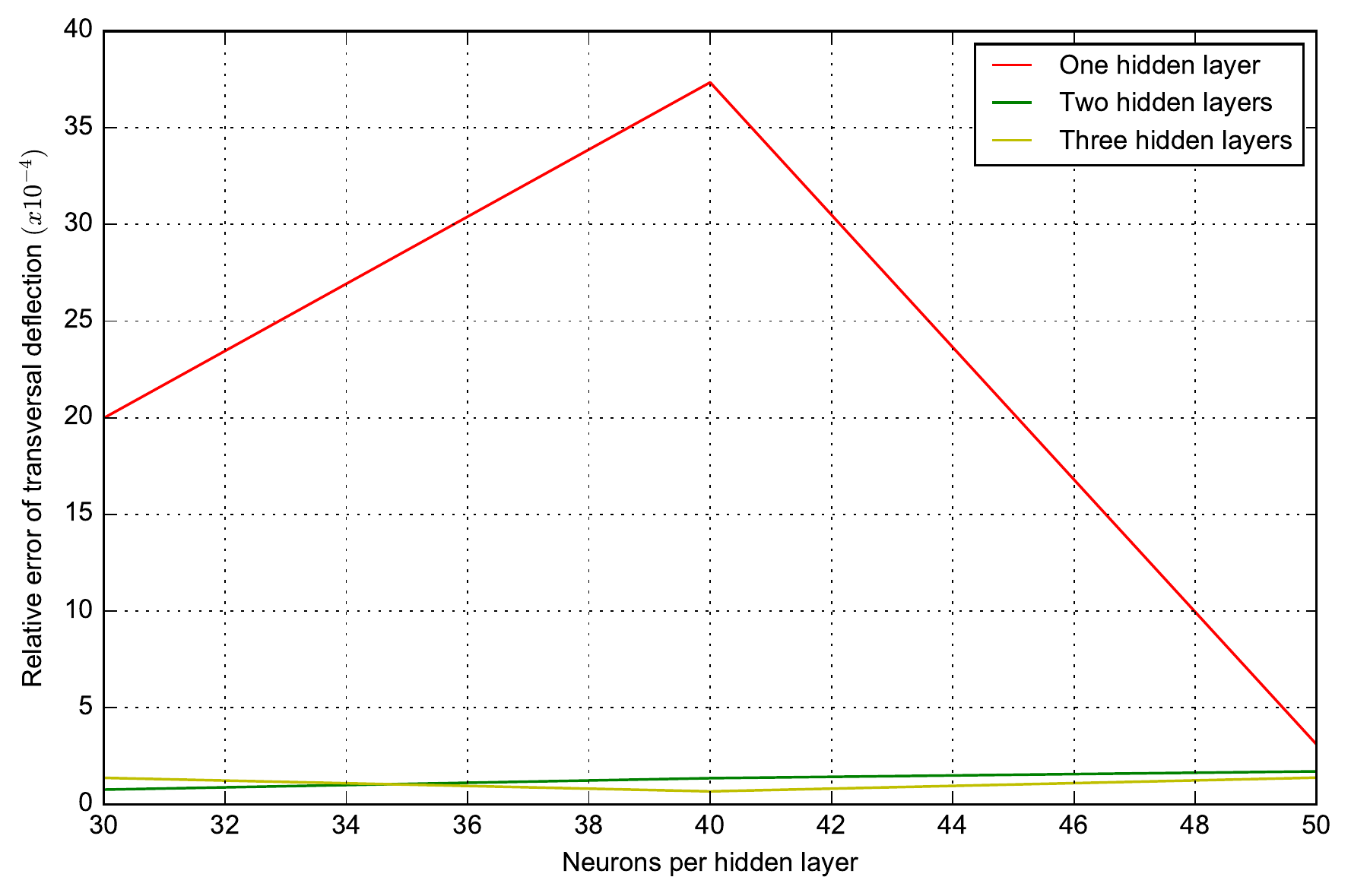}\hfill
   \caption{The relative error of deflection with varying hidden layers and neurons}
 \end{subfigure}
 \caption{}
 \label{Scatterpointcc}
 \end{figure}

Also, in this section, we study the feasibility of deep collocation method in the analysis of Kirchhoff plate bending problems.
We first apply the deep collocation method in studying a clamped circular plate with radius $R$ under a uniform load $p$. 1000 collocation points shown in Figure~\ref{Scatterpointcc}(a) are employed in the domain of the circular plate. The exact solution of this problem can be referred in \cite{timoshenko1959theory}:
\begin{equation}
w=\frac{p\left ( R^{2}-\left ( x^{2}+ y^{2}\right ) \right )^{2}}{64D},
\end{equation}   
$D$ denoting the flexural stiffness. The maximum deflection at the central of the circular plate with varying hidden layers and neurons is summarized in Table 3 and compared with the exact solution. The predicted maximum deflection become more accurate with increasing number of  neurons and hidden layers. The relative error for deflection of clamped circular plate with increasing hidden layers and neurons is depcited in Figure~\ref{Scatterpointcc}(b). As hidden layer number increases, the relative error curves flatten out and converges  to the exact solution. All neural networks perform well with a relative error magnitude of $1\times10^{-4}$.

\begin{figure}
\centering
\caption*{Table 2: Maximum deflection predicted by deep collocation method.}
\includegraphics[width=10cm]{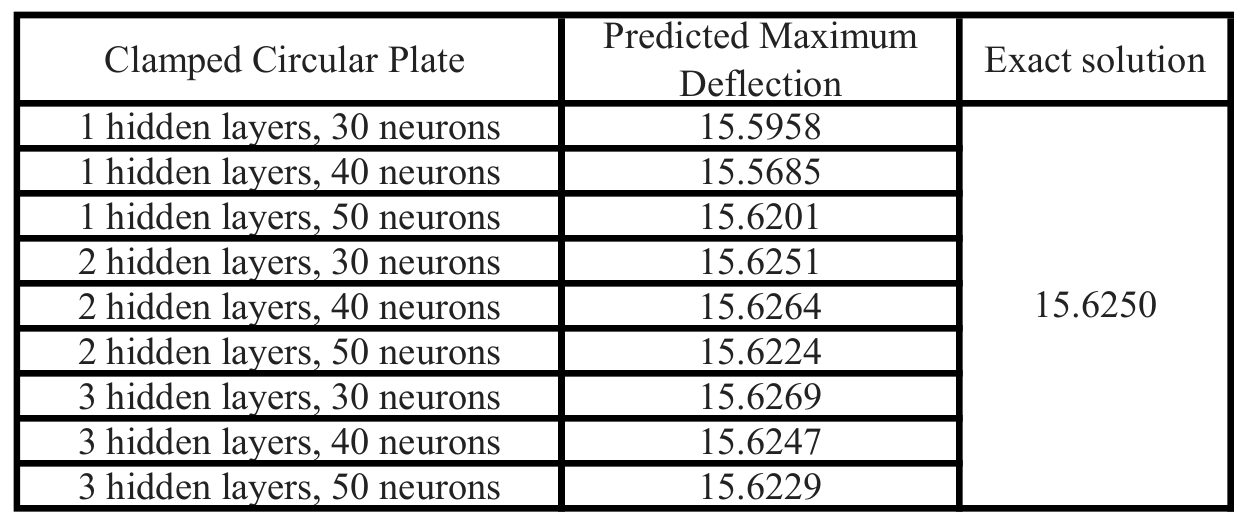} 
\end{figure}

\begin{figure}
 \centering
 \begin{subfigure}[b] {0.4\textwidth}
   \centering
   \includegraphics[width=\textwidth]{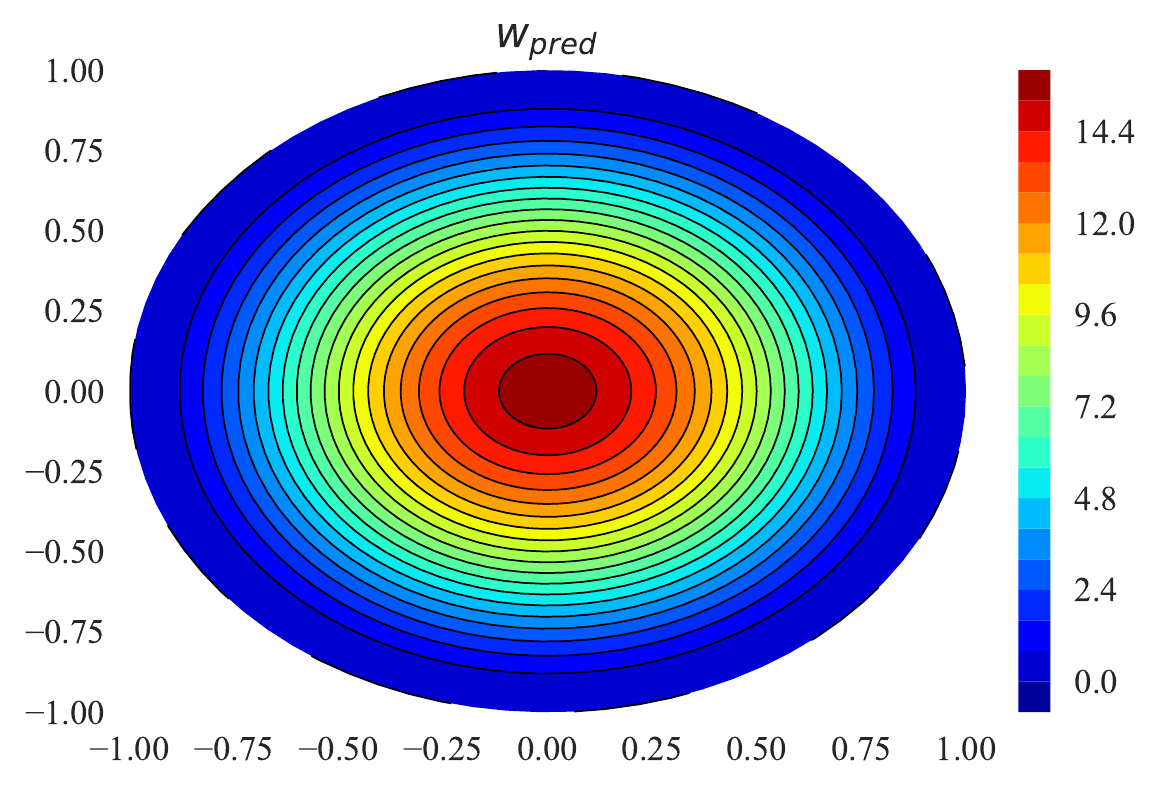}\hfill
   \caption{}\label{fig:contourpredssef_kir}
 \end{subfigure}\hspace{5pt}
 \begin{subfigure}[b] {0.4\textwidth}
   \centering
   \includegraphics[width=\textwidth]{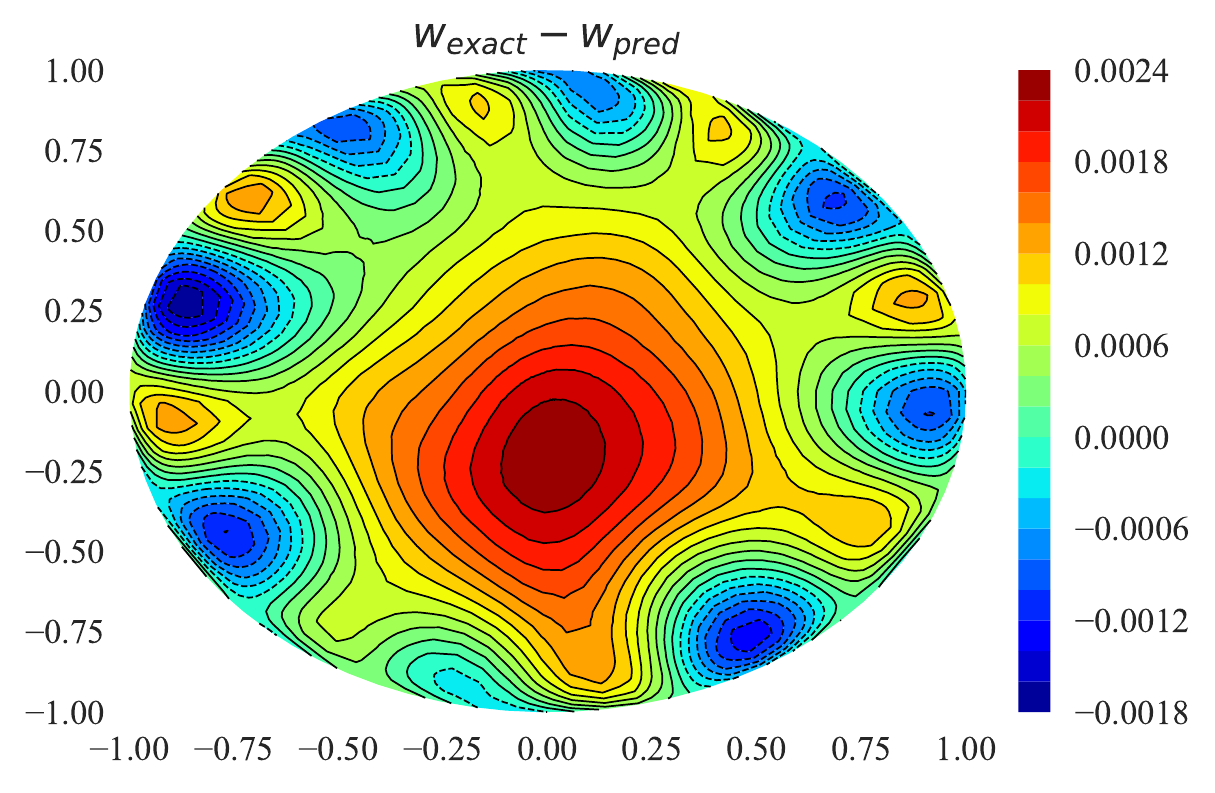}\hfill
   \caption{}\label{fig:errocontourpredssef_kir}
 \end{subfigure}\hspace{5pt}
  \begin{subfigure}[b] {0.48\textwidth}
   \centering
   \includegraphics[width=\textwidth]{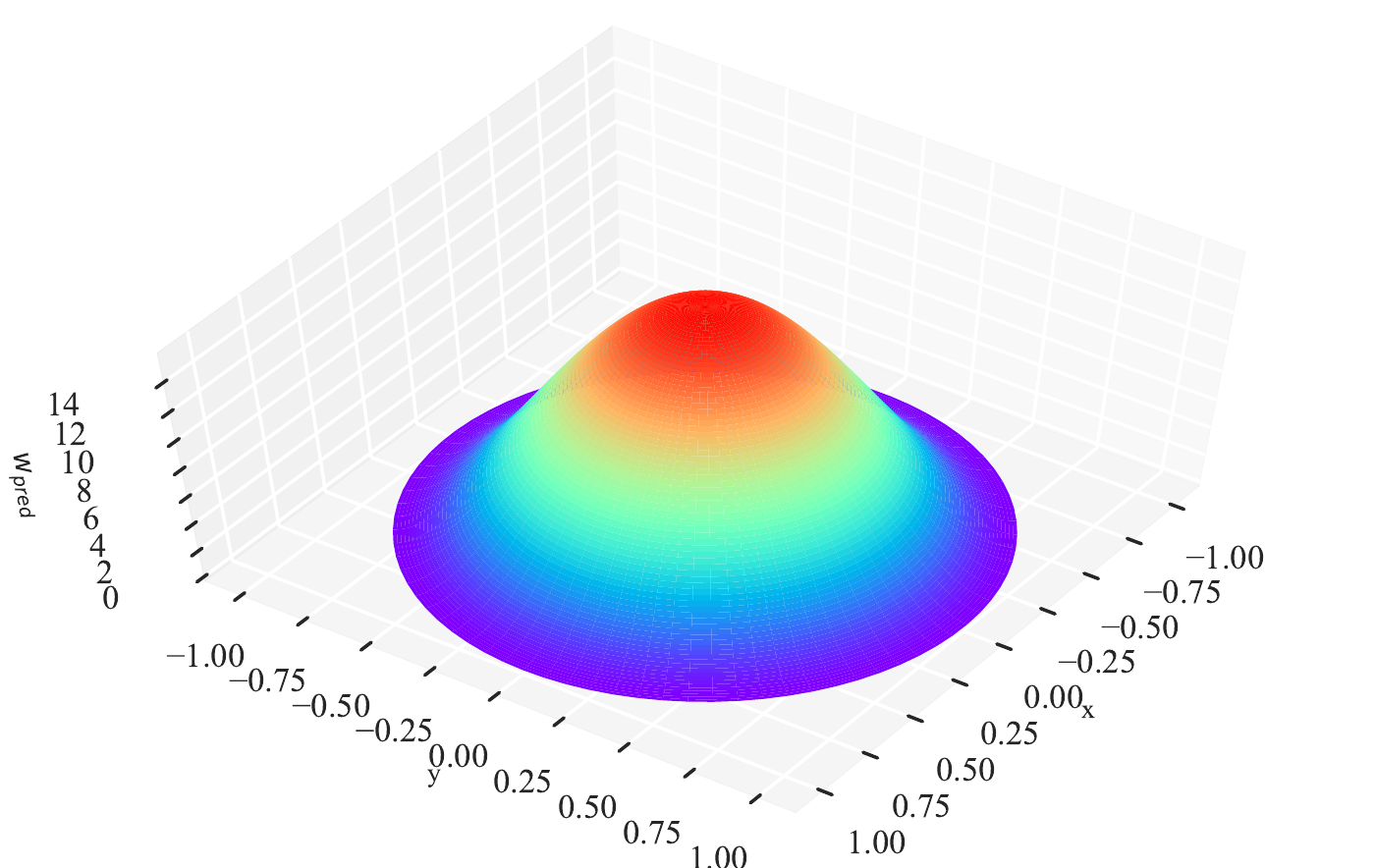}\hfill
   \caption{}\label{fig:Dflpredssef_kir}
 \end{subfigure}\hspace{1pt}
 \begin{subfigure}[b] {0.48\textwidth}
   \centering
   \includegraphics[width=\textwidth]{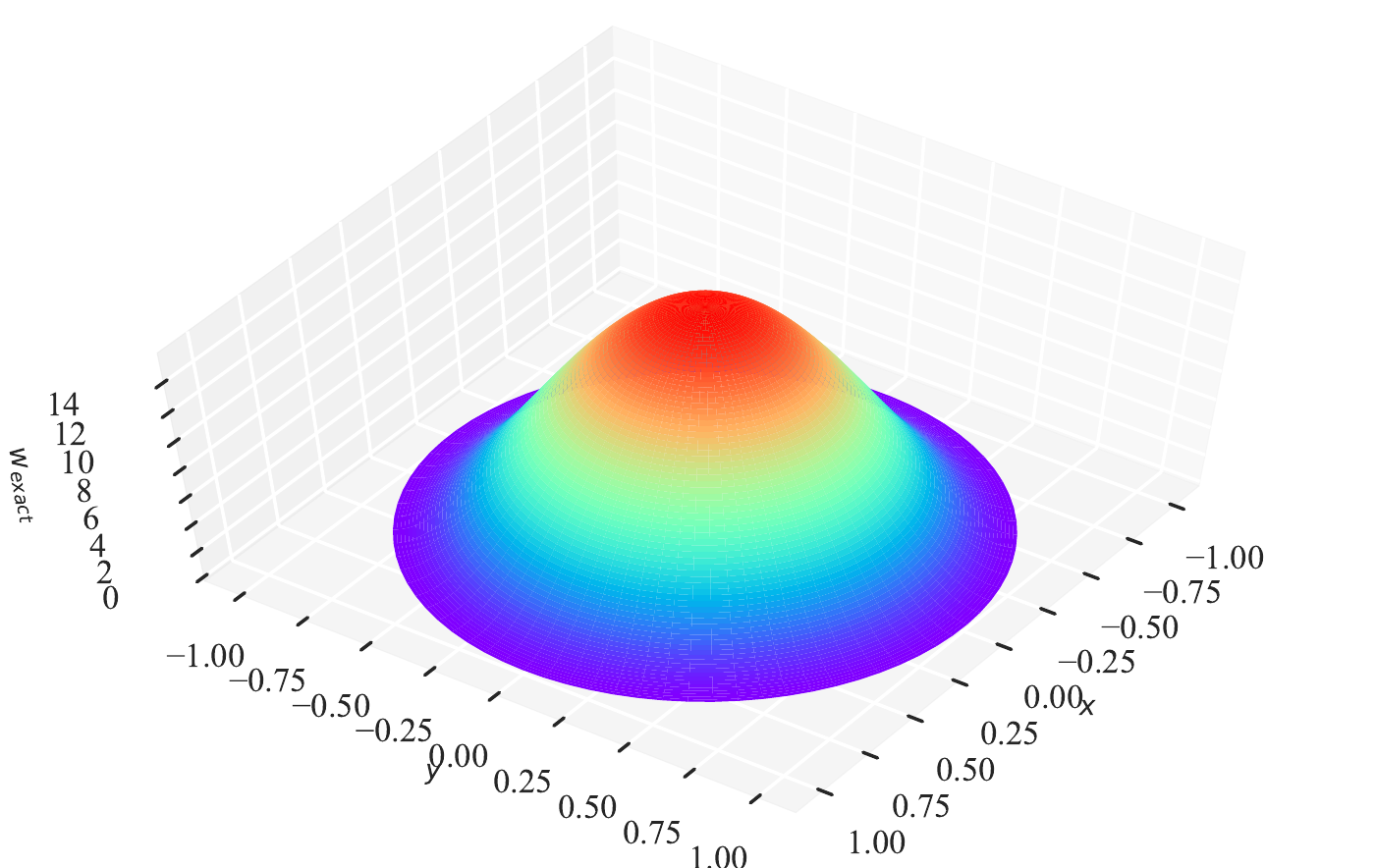}\hfill
   \caption{}\label{fig:Dflexactssef_kir}
 \end{subfigure}\hspace{1pt}
 \caption{$\left(a\right)$ Predicted deflection contour $\left(b\right)$ Deflection error contour $\left(c \right)$ Predicted deflection $\left(d \right)$ Exact deflection of the clamped circular plate with 3 hidden layers and 50 neurons}
 \label{ccesneuron50point1}
\end{figure}

The deformation contour, deflection error contour, predicted and exact deformation figures are displayed in Figure~\ref{ccesneuron50point1}. It is clear that the predicted deflection of this circular plate agrees well with the exact solution. 

\subsubsection{Simply-supported square plate under a sinusoidally distributed load}
\label{subsec6:kirchoff_prob3}
For the next two numerical examples, deep energy method is applied in the calculation. For deep energy method, we use a different DNN scheme based on Pytorch, and to further improve the primary results, different strategies have been proposed, including an autoencoder and tailored activation function, which has been proved to be feasible and efficient during the numerical experiment. The LBFGS optimizer is applied during the training of the deep neural network. The numerical experiments are executed on a 64-bit macOS Mojave server with Intel(R) Core(TM) i7-8850H CPU, 32GB memory.

First, a simply-supported square plate under a sinusoidal distribution transverse loading is studied with deep energy method. The distributed load is given by
\begin{equation}
\begin{array}{l}
p=\frac{p_{0}}{D}\textrm{sin}\left (\frac{\pi x}{a}  \right )\textrm{sin}\left (\frac{\pi y}{b}  \right ).
\end{array}
\end{equation} 
where $a$,$b$ are the length of the plate.The analytical solution for this problem is given by 
\begin{equation}
\begin{array}{l}
w=\frac{p_{0}}{\pi^{4} D\left (\frac{1}{a^{2}}+\frac{1}{b^{2}}  \right )^{2}}\textrm{sin}\left (\frac{\pi x}{a}  \right )\textrm{sin}\left (\frac{\pi y}{b}  \right ).
\end{array}
\end{equation}  

\begin{figure}
 \centering
 \begin{subfigure}[b] {0.4\textwidth}
   \centering
   \includegraphics[width=\textwidth]{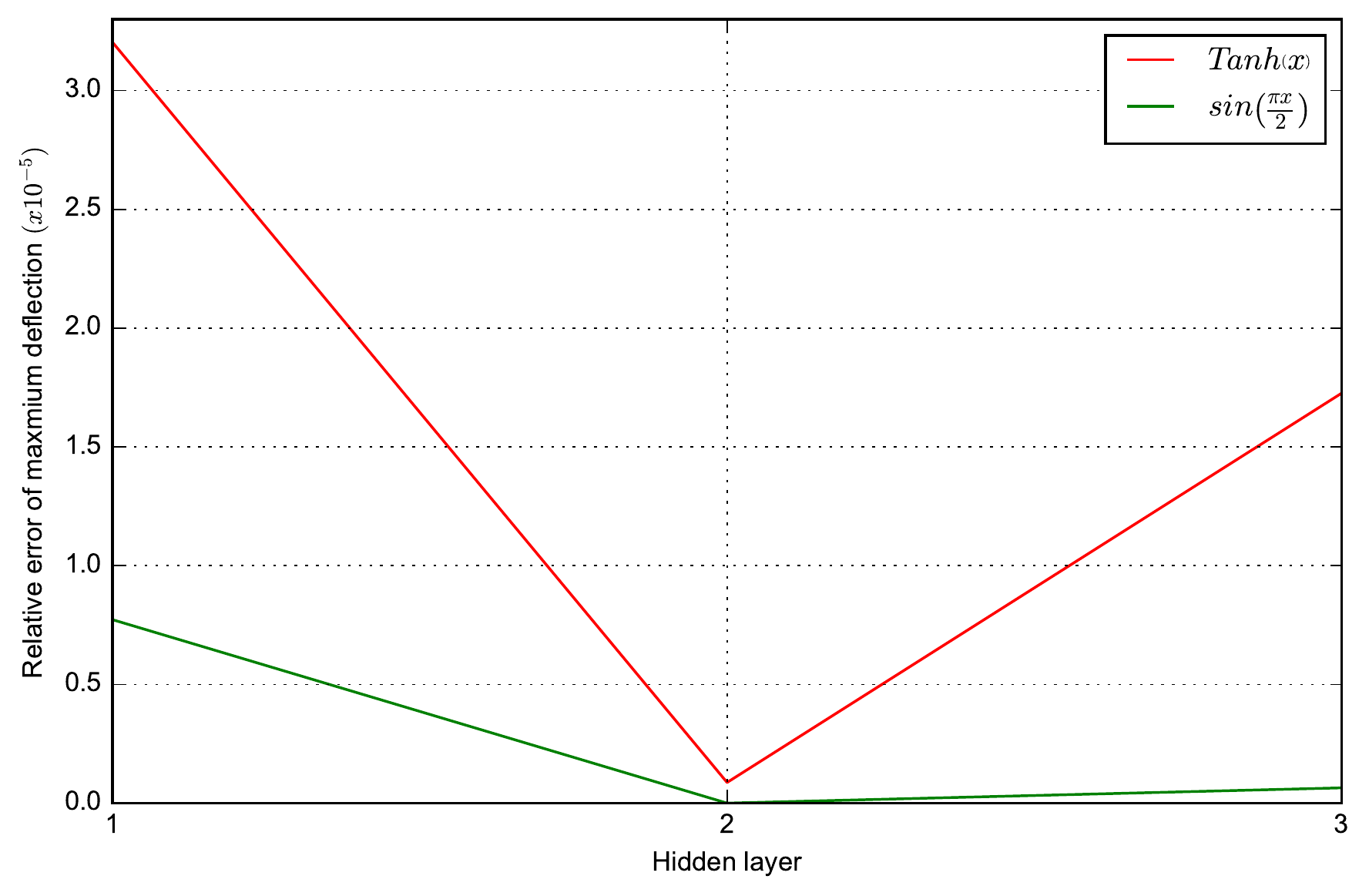}\hfill
   \caption{}
 \end{subfigure}\hspace{5pt}
 \begin{subfigure}[b] {0.4\textwidth}
   \centering
   \includegraphics[width=\textwidth]{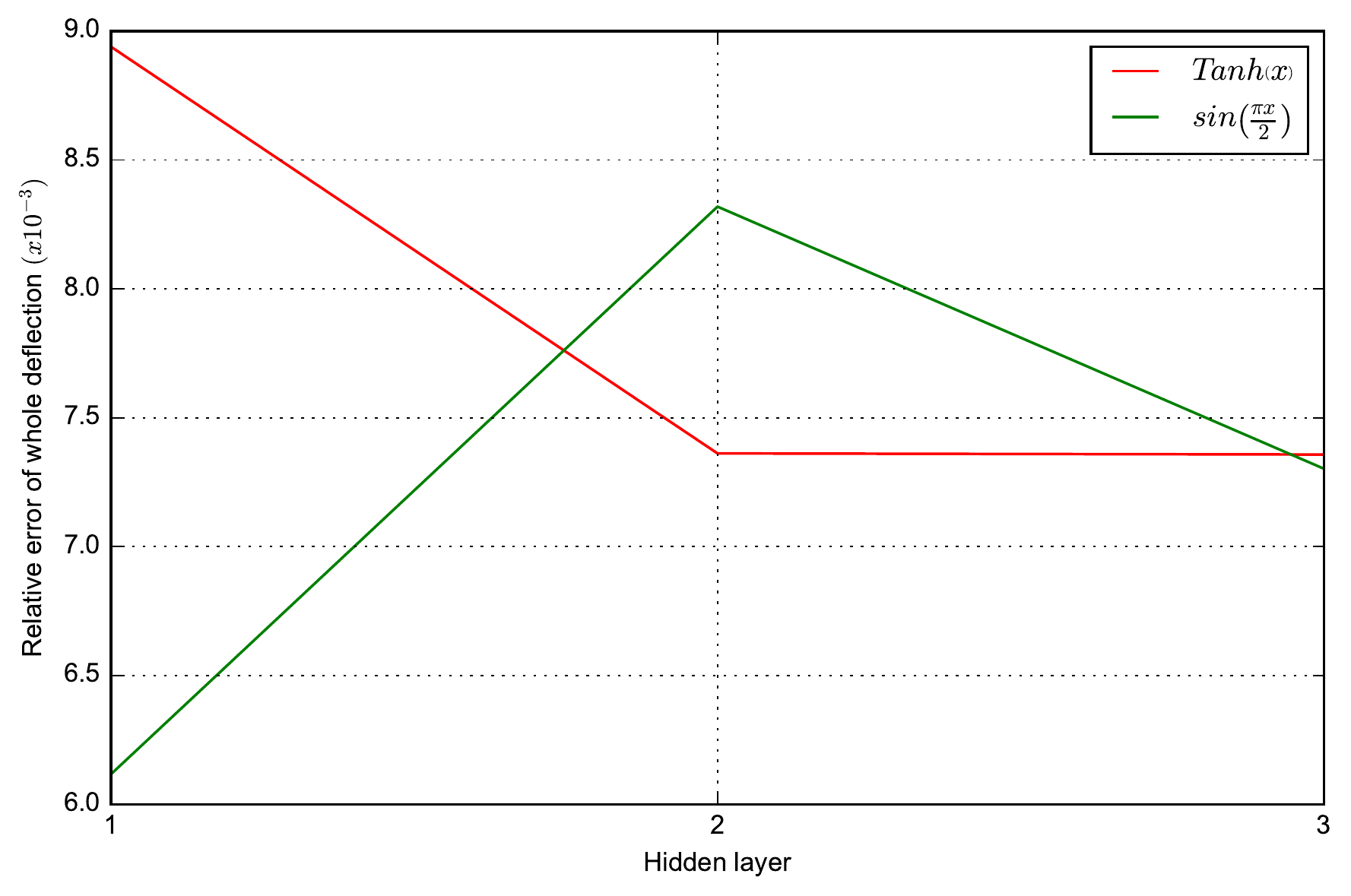}\hfill
   \caption{}
 \end{subfigure}\hspace{5pt}
 \caption{Relative error of $\left(a\right)$ maximum deflection and $\left(b\right)$ whole deflection predicted by Tanh and proposed activation function of a DNN for the simply-supported plate}
 \label{RelativerrDNNActivationFun}
\end{figure}

\begin{figure}
 \centering
 \begin{subfigure}[b] {0.4\textwidth}
   \centering
   \includegraphics[width=\textwidth]{Images/DEMfigMaxDflRelativeerror}\hfill
   \caption{}
 \end{subfigure}\hspace{5pt}
 \begin{subfigure}[b] {0.4\textwidth}
   \centering
   \includegraphics[width=\textwidth]{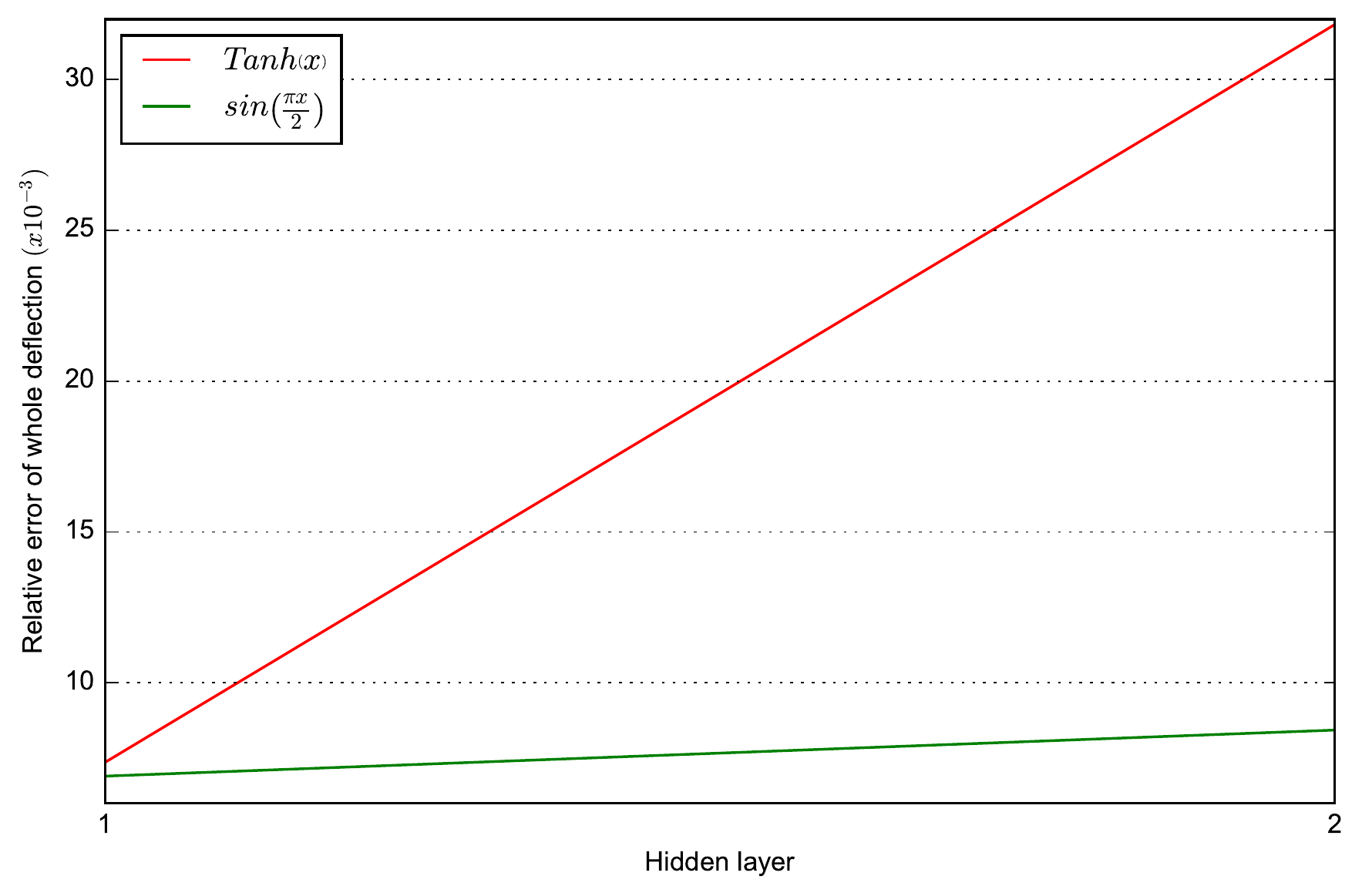}\hfill
   \caption{}
 \end{subfigure}\hspace{5pt}
 \caption{Relative error of $\left(a\right)$ maximum deflection and $\left(b\right)$ whole deflection predicted by Tanh and proposed activation function of a DNN with an autoencoder for the simply-supported plate}
 \label{RelativerrDNNCODActivationFun}
\end{figure}

For the detailed configuration of this numerical example, a $31x31$ random distributed collocation point set shown in Figure~\ref{Scatterpoint} is first generated in the plate domain. Based on this, the Monte Carlo integration scheme is adopted in the calculation of total potential energy. For this problem with sinusoidal load, the activation function was tailored accordingly as $f\left(x\right)=sin\left(\frac{\pi x}{2}\right)$, which improves the accuracy of the results. A neural network with 50 neurons each hidden layer is deployed to study the relative error of maximum deflection at central points and the whole deflection by the proposed activation function and classical Tanh activation function. And for the DNN, three hidden layers is deployed. As for DNN with an autoencoder, two encoding layer configurations $[2,1]*H$ and $[3]*H$ are considered, with $H$ the number of neurons on each encoding layer. It is shown clearly in Figure~\ref{RelativerrDNNActivationFun}, Figure~\ref{RelativerrDNNCODActivationFun} that the numericals gained by proposed activation function outweigh classical Tanh activation function for both two neural network schemes.  

\begin{figure}
\begin{subfigure}[b]{0.45\textwidth} 
  \centering\includegraphics[width=\textwidth]{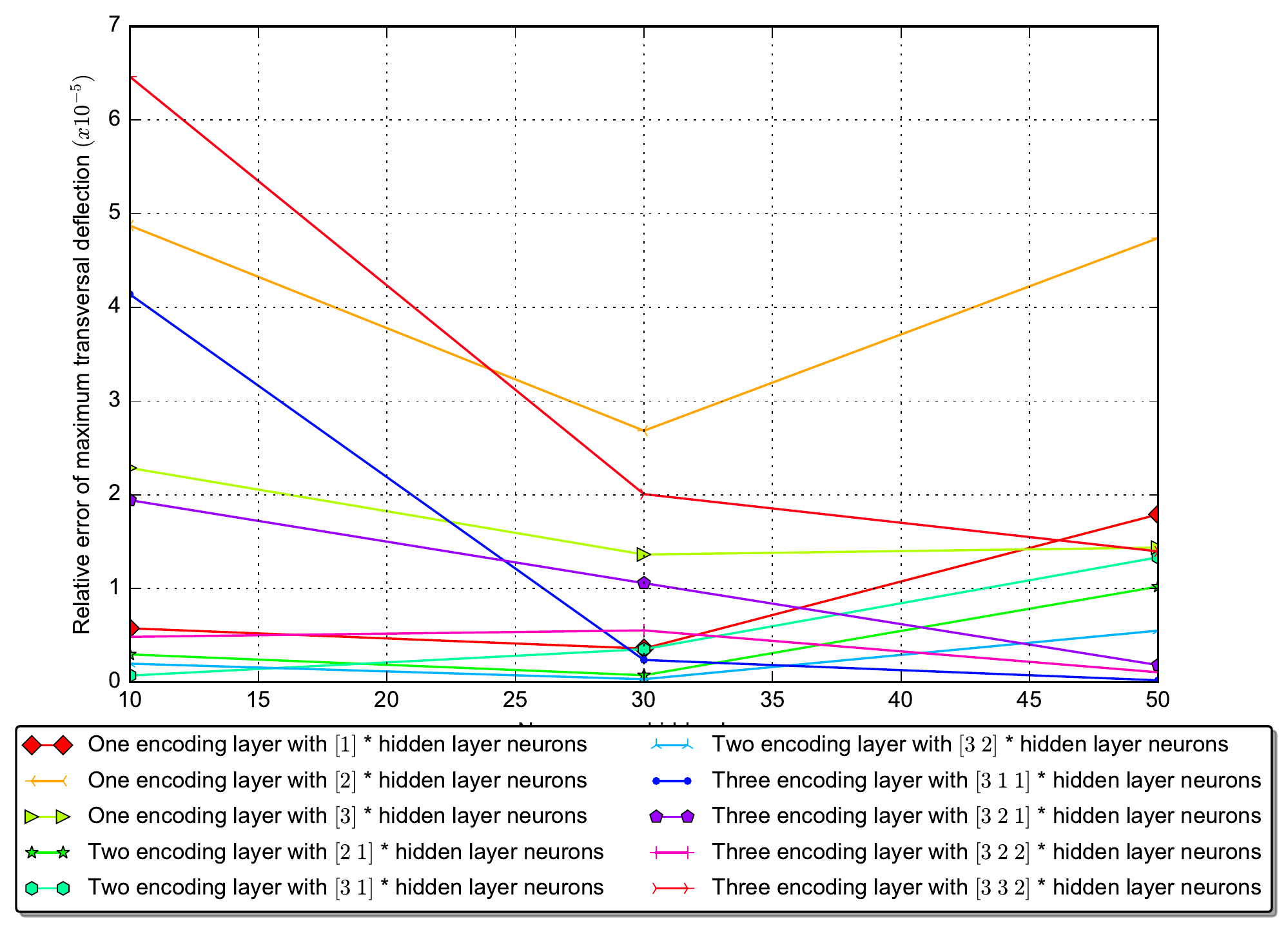}\hfill
   \caption{}
 \end{subfigure}\hspace{5pt}
 \begin{subfigure}[b]{0.45\textwidth} 
  \centering\includegraphics[width=\textwidth]{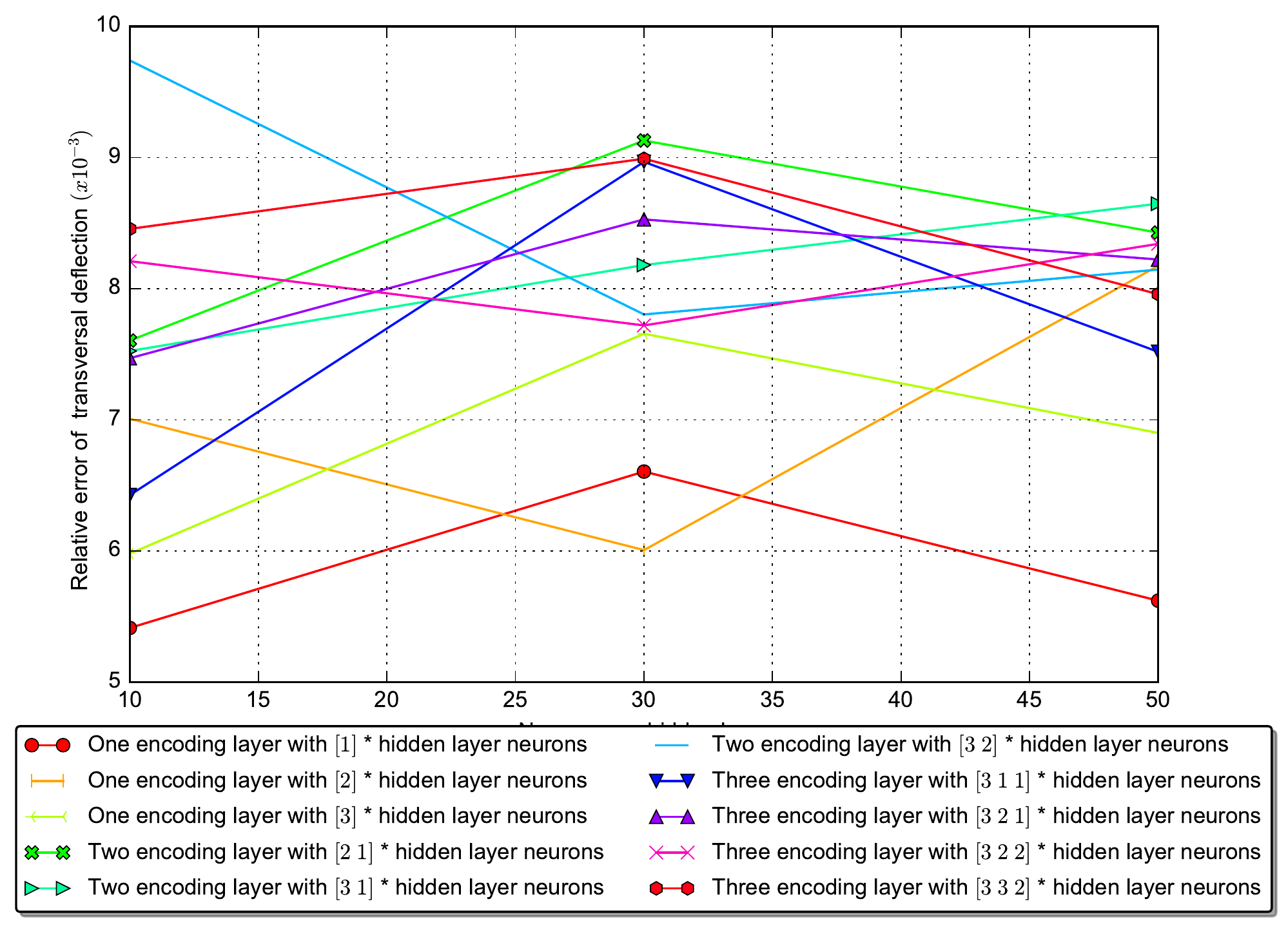}\hfill   
   \caption{}
 \end{subfigure}\hspace{5pt}
 \caption{Relative error of $\left(a\right)$ maximum deflection and $\left(b\right)$ whole deflection predicted by different encoding layer configurations of a DNN with an autoencoder for the simply-supported plate.}
\label{RelativerrDNNCODConf}
\end{figure}

\begin{figure}
\centering
\includegraphics[width=7cm]{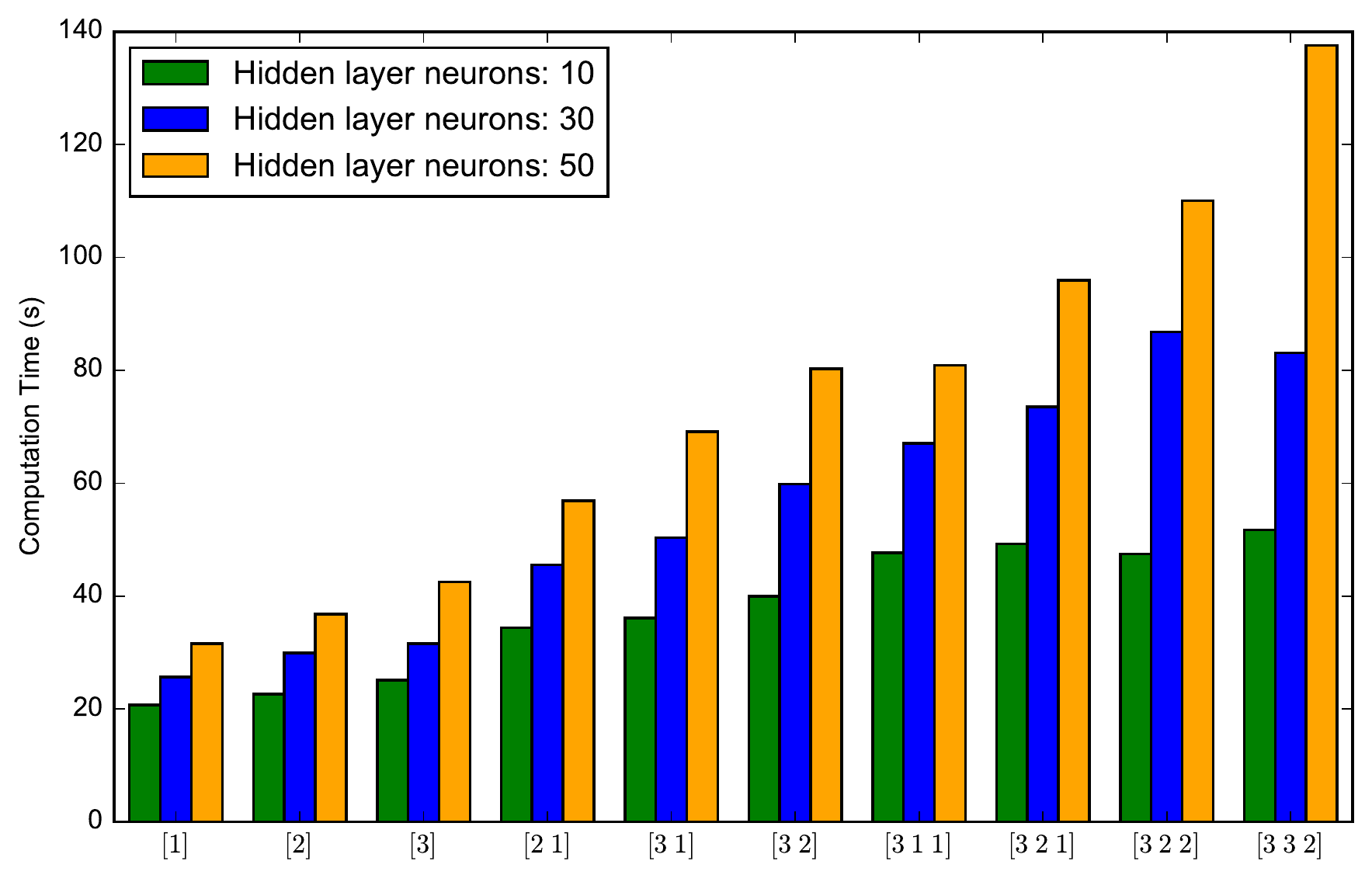}
\caption{The computational time of ten encoding layer configurations for the autoencoder}
\label{ComputationDNNCODConf}
\end{figure}

Moreover, we carry on studying an accurate and efficient configuration of the proposed DNN with an autoencoder, in hope for a better discovery of its physical patterns. For this numerical example, ten encoding layer types are studied and shown in Figure~\ref{RelativerrDNNCODConf}. We begin by studying the influence of different encoding layer with different neurons per layer on the accuracy and efficiency of this problem. We can see that with more encoding layer and neurons per layer, it will obtain a more stable and accurate results. Moreover, the corresponding computational cost of those ten schemes are listed in Figure~\ref{ComputationDNNCODConf}. It is clear that a two encoding layer autoencoder can meet both accuracy and efficiency ends. And the neurons on each scheme increase, it is clearly that the numerical solution converges to the analytical solution. 

Finally, the deep energy scheme is then compared with an open source IGA-FEM code from  on this problem \cite{nguyen2015isogeometric}, which is also based on the kirchhoff plate theory. The accuracy and computational time of both scheme are compared. For DEM, a  single encoding layer with 30 neurons is applied with DNN including increasing collocation points. And the computational time for DEM means the training time. It is clear that IGA-FEM is clearly faster and more accurate. But, the performance of DEM is still in the acceptance range. Once the DNN with an autoencoder is trained, it can be very fast used to predict deflection and stress in the whole plate. 

For deep energy method configuration, an auto-encoder is added to the deep neural networks to better capture the physical patterns of this problem. We also studied the optimal configuration of this auto encoder with different layers and neuron numbers, which will be beneficial to the further application of this deep energy method.

\begin{figure}
 \centering
 \begin{subfigure}[b]{0.4\textwidth}
   \centering
   \includegraphics[width=\textwidth]{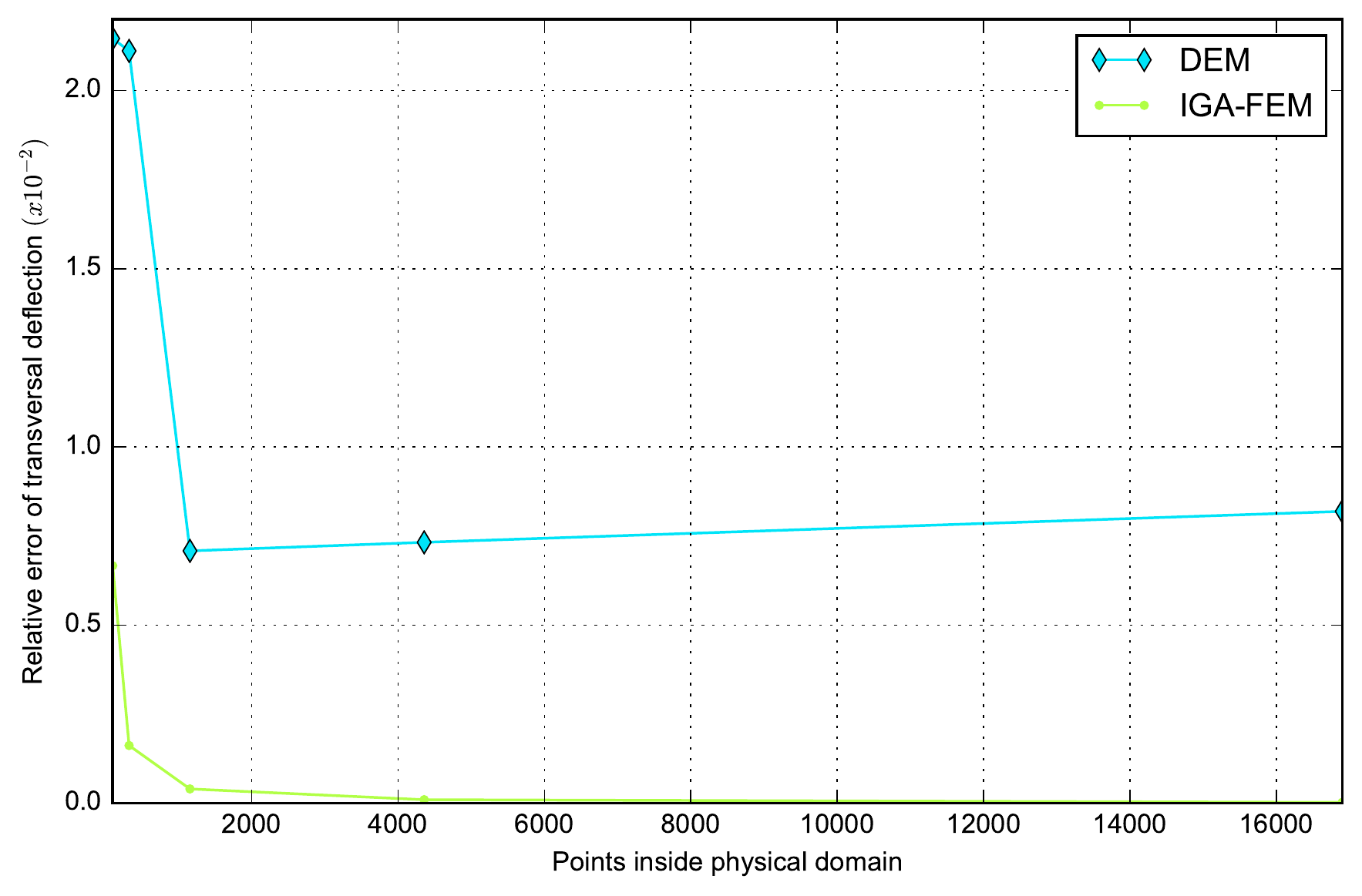}\hfill
   \caption{}
 \end{subfigure}\hspace{5pt}
 \begin{subfigure}[b] {0.4\textwidth}
   \centering
   \includegraphics[width=\textwidth]{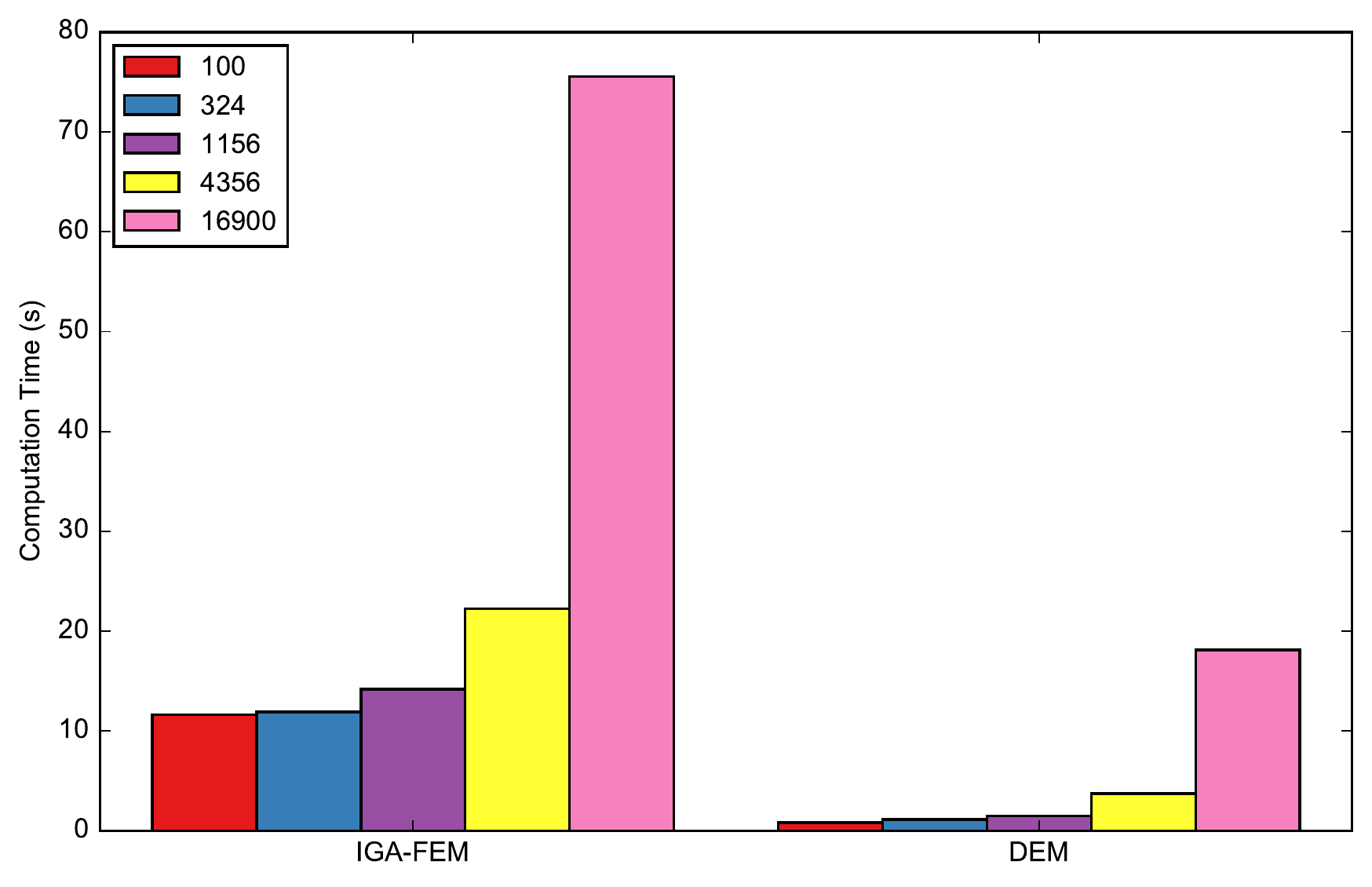}\hfill
   \caption{}
 \end{subfigure}\hspace{5pt}
 \caption{Relative error of $\left(a\right)$ deflection and $\left(b\right)$ predicted by the DNN with an autoencoder and IGA for the simply-supported plate.}
 \label{IGADNNCOD}
\end{figure}

\begin{figure}
 \centering
 \begin{subfigure}[b]{0.4\textwidth}
   \centering
   \includegraphics[width=\textwidth]{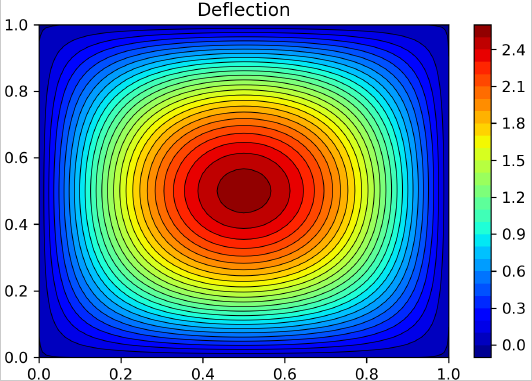}\hfill
   \caption{}\label{contourpredsspl}
 \end{subfigure}\hspace{5pt}
 \begin{subfigure}[b] {0.4\textwidth}
   \centering
   \includegraphics[width=\textwidth]{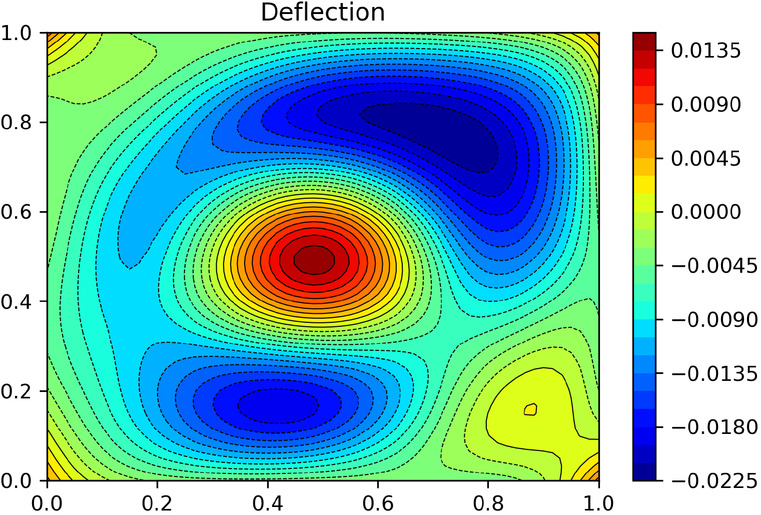}\hfill
   \caption{}\label{errocontourpredsspl}
 \end{subfigure}\hspace{5pt}
  \begin{subfigure}[b] {0.4\textwidth}
   \centering
   \includegraphics[width=\textwidth]{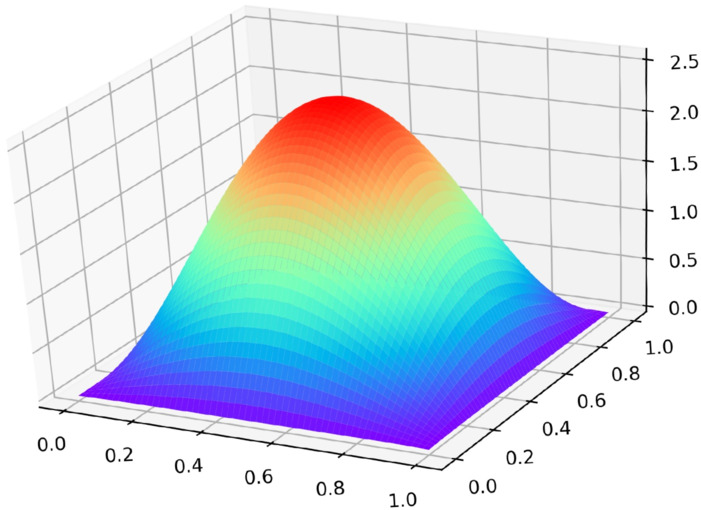}\hfill
   \caption{}\label{Fig:Dflpredsspl}
 \end{subfigure}\hspace{5pt}
 \begin{subfigure}[b] {0.4\textwidth}
   \centering
   \includegraphics[width=\textwidth]{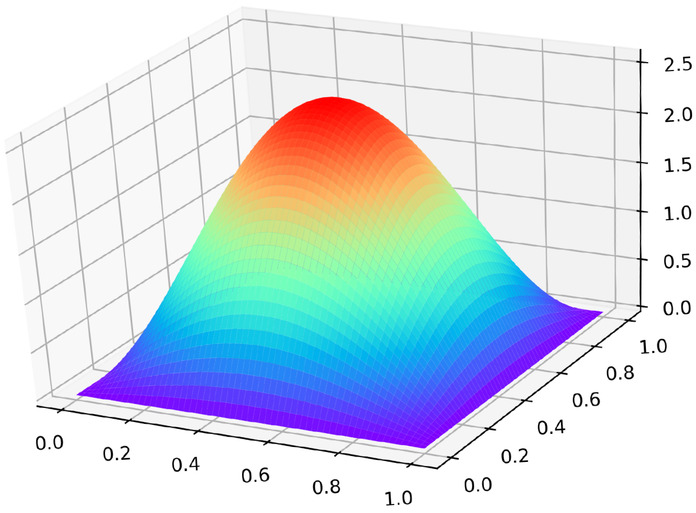}\hfill
   \caption{}\label{Dflexactsspl}
 \end{subfigure}\hspace{5pt}
 \caption{$\left(a\right)$ Predicted deflection contour $\left(b\right)$ Deflection error contour $\left(c \right)$ Predicted deflection $\left(d \right)$ Analytical deflection of the simply-supported square plate.}
 \label{contoursspl}
\end{figure}

To better reflect the deflection vector in the whole physical domain, the contour plot, contour error plot of deflection predicted by two encoding layers with [150,100] neurons are shown in Figure~\ref{contoursspl}, which agree well with  deflection obtained from analytical solution.

\subsubsection{Simply-supported annular plate}
\label{subsec6:kirchoff_prob4}
In this section, an annular plate, which is simply-supported on the outer circle and free on the inner circle, is  studied with deep energy method, for this problem, deep collocation method performs poorly for a feed-forward DNN, when considering free boundary conditions. However, for energy based method, only the essential boundary conditions needs to be taken into consideration, which simplifies the problem. The numerical results predicted by the deep energy method lend credence to the feasibility of deep energy method in analysis of plates with cut-out.

The analytical solution of this problem is \cite{timoshenko1959theory}:
\begin{equation}
w=\frac{qa^4}{64D}\left \{ -\left [ 1-\left ( \frac{r}{a} ^{4}\right ) \right ] +\frac{2\alpha_{1}}{1+\nu }\left [ 1-\left (\frac{r}{a}  \right )^{2} \right ]-\frac{4\alpha _{2} \beta^2}{1-\nu }\textup{log}\left ( \frac{r}{a} \right )\right \},
\end{equation}
where $\alpha_1=\left ( 3+\nu  \right )\left ( 1-\beta^2\right )-4\left ( 1+\nu \right )\beta^2\kappa$, $\alpha_2=\left ( 3+\nu  \right )+4\left ( 1+\nu \right )\kappa$, $\beta=\frac{b}{a}$, $ \kappa=\frac{\beta^2}{1-\beta^2}\textup{log}\beta$, with $a$, $b$ the outer and inner radius of the annular plate respectively.

\begin{figure}
\centering
\includegraphics[width=7cm]{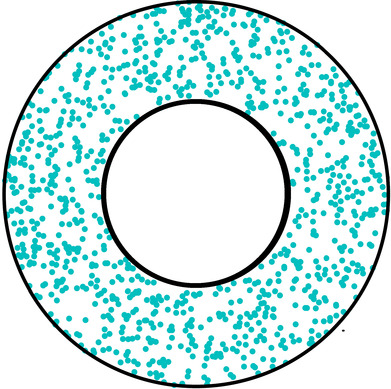} 
\caption{Collocation points in the annular domain.}
\label{asplscatterpointas}
\end{figure}

Likewise, we studied the deflection at a certain point to study the convergency of deflection with the varying encoding layers, which is chosen for annular plate as $(\frac{a+b}{2},0)$. In this numerical example, 1000 collocation points are generated in the physical domain. A DNN with an autoencoder is also constructed which will help better predict the physical pattern of this problem. During our numerical example, it is discovered that the proper point sampling inside physical domain affects the stability and accurate of this deep energy method. And compared with some other sampling method, the collocation points generated in Figure~\ref{asplscatterpointas} can be suitable for the numerical analysis of this problem.

\begin{figure}
\centering
\begin{subfigure}[b]{0.45\textwidth} 
  \centering\includegraphics[width=\textwidth]{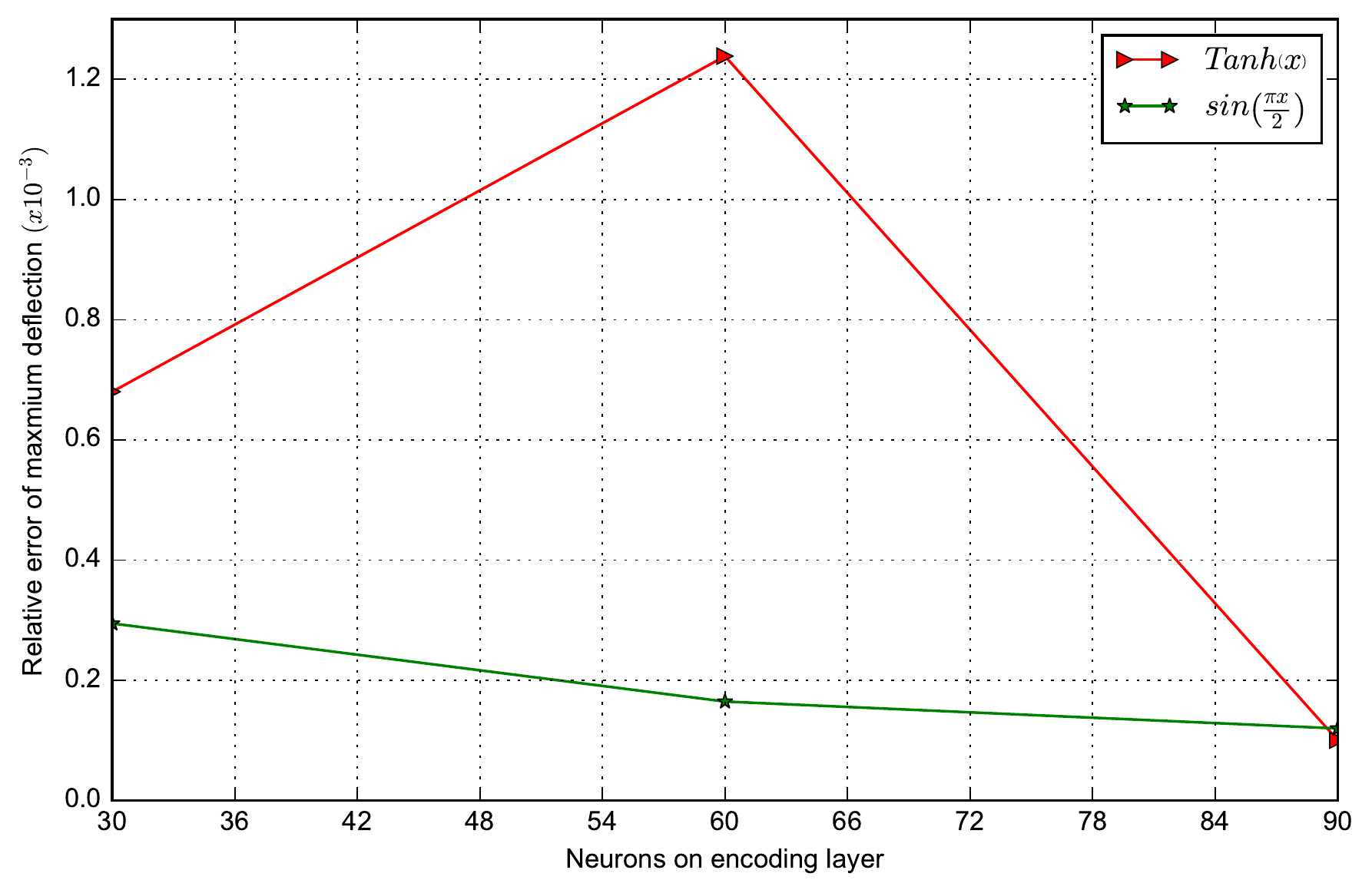}\hfill
   \caption{}
 \end{subfigure}\hspace{5pt}
 \begin{subfigure}[b]{0.45\textwidth} 
  \centering\includegraphics[width=\textwidth]{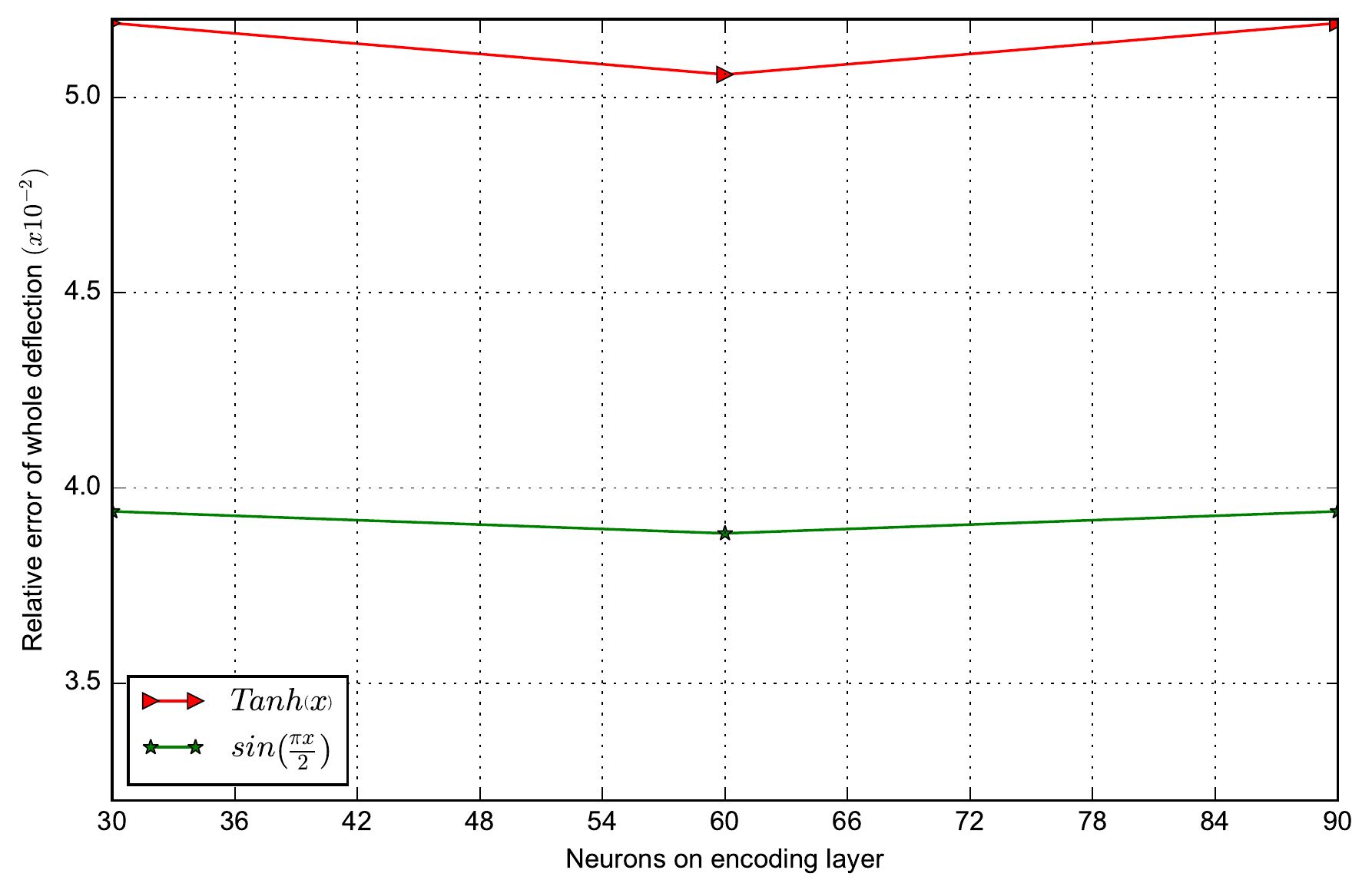}\hfill   
   \caption{}
 \end{subfigure}\hspace{5pt}
 \caption{Relative error of $\left(a\right)$ maximum deflection and $\left(b\right)$ whole deflection predicted by $\it{tanh}$ and proposed activation function of a DNN for the annular plate.}
\label{RelativerrDNNActivationFunAS}
\end{figure}

As for the numerical analysis, we first test the deep neural network with an autocoder with different activation function, namely the mostly used $tanh\left(x\right)$ and the proposed $sin\left(\frac{\pi x}{2}\right)$. Shown in Figure~\ref{RelativerrDNNActivationFunAS} is very clear that the proposed activation based DNN better predicts the deflection for the whole plate.

\begin{figure}
\centering
\includegraphics[width =0.7\textwidth]{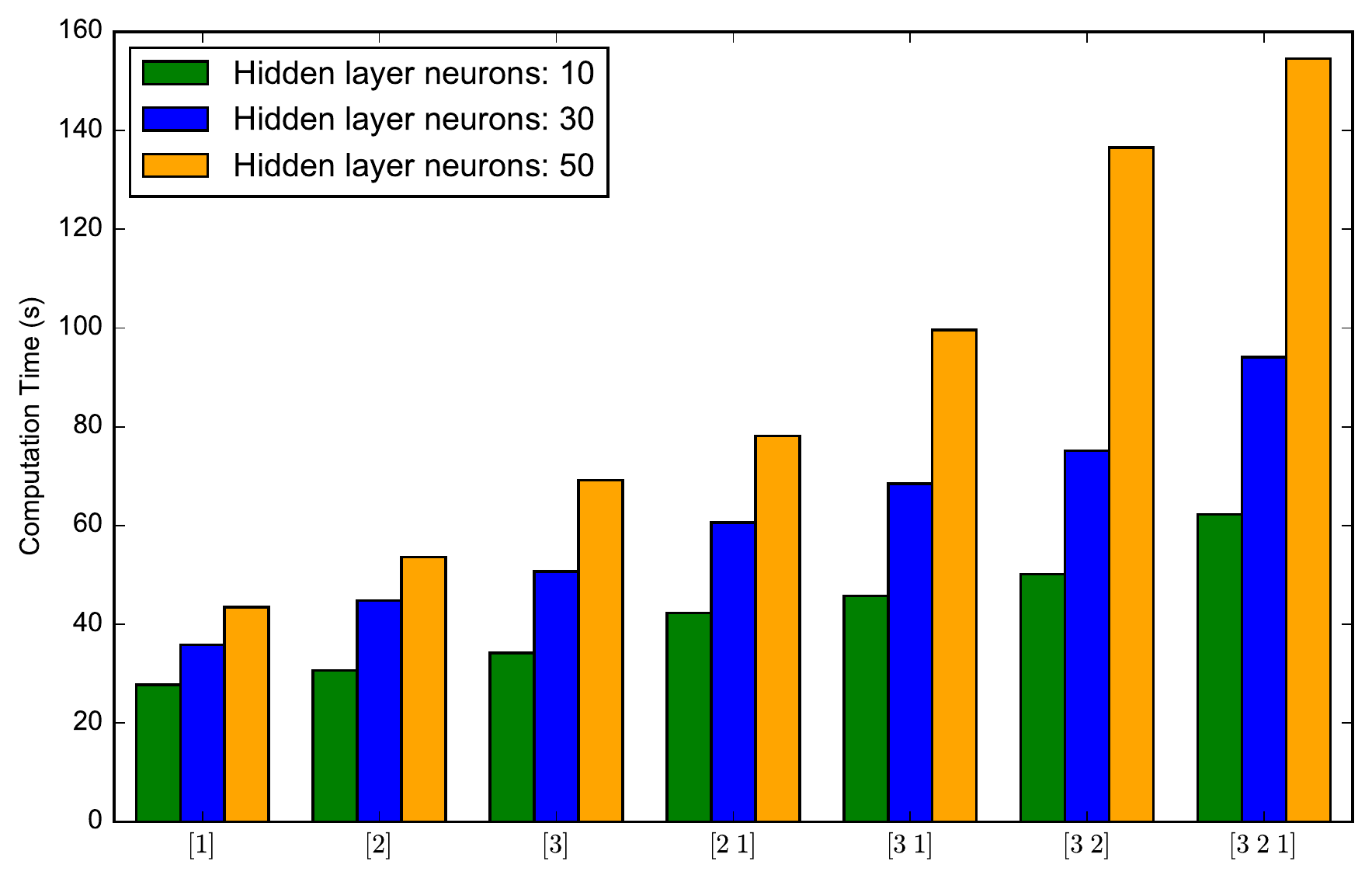}
\caption{The computational time of seven encoding layer configurations for the autoencoder.}
\label{ComputationDNNCODConfas}
\end{figure}

\begin{figure}
 \centering
 \begin{subfigure}[b] {0.4\textwidth}
   \centering
   \includegraphics[width=\textwidth]{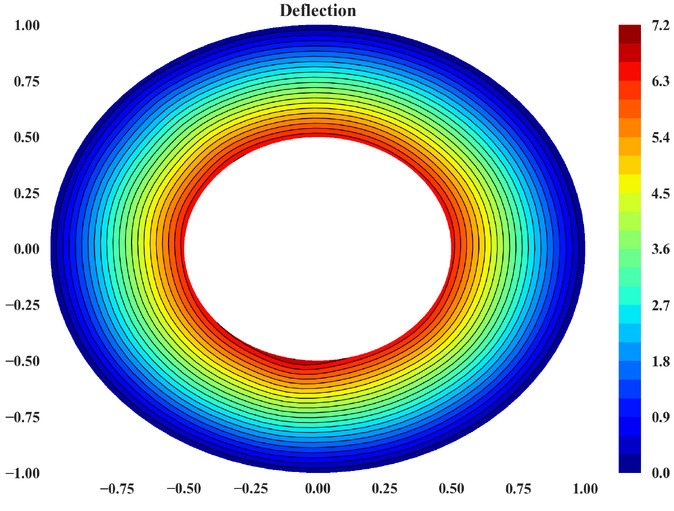}\hfill
   \caption{}\label{contourpredaspl}
 \end{subfigure}\hspace{5pt}
 \begin{subfigure}[b] {0.4\textwidth}
   \centering
   \includegraphics[width=\textwidth]{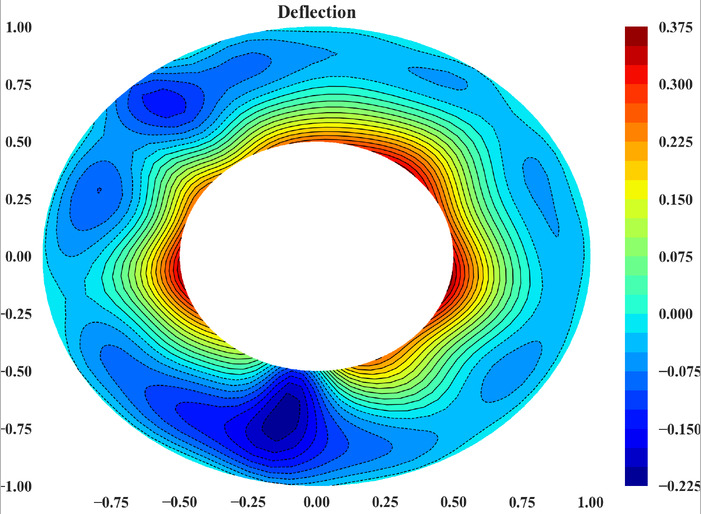}\hfill
   \caption{}\label{errorcontourpredaspl}
 \end{subfigure}\hspace{5pt}
  \begin{subfigure}[b] {0.4\textwidth}
   \centering
   \includegraphics[width=\textwidth]{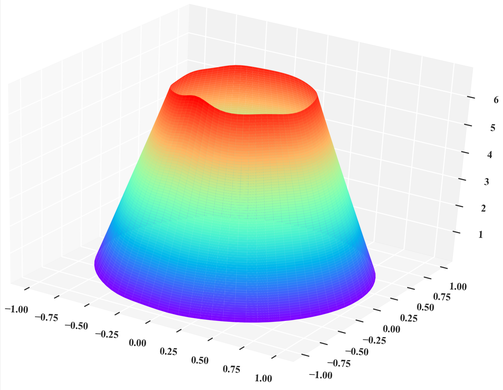}\hfill
   \caption{}\label{Dflpredaspl}
 \end{subfigure}\hspace{5pt}
 \begin{subfigure}[b] {0.4\textwidth}
   \centering
   \includegraphics[width=\textwidth]{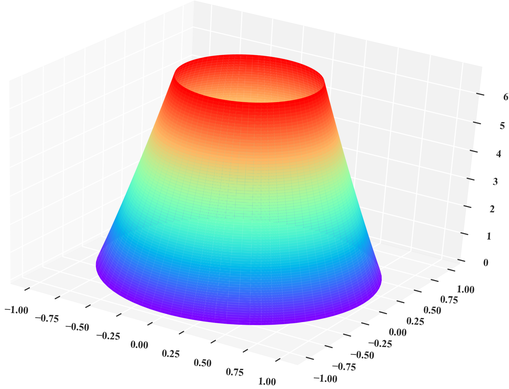}\hfill
   \caption{}\label{Dflexactaspl}
 \end{subfigure}\hspace{5pt}
 \caption{$\left(a\right)$ Predicted deflection contour $\left(b\right)$ Deflection error contour $\left(c \right)$ Predicted deflection $\left(d \right)$ Analytical deflection of the simply-supported annular plate with 3 hidden layers and 50 neurons.}
 \label{contouraspl}
\end{figure}

Further, we studied some chosen encoding layer configurations in affecting the deflection of the annular plate and its corresponding computational cost. Shown in Figure~\ref{contouraspl}, as the number of neurons and encoding layers increase, the deflection converges to the analytical solution. Combined with computational cost demonstrated in Figure~\ref{ComputationDNNCODConfas}, it can be concluded that two encoding layers can be sufficient for the bending analysis of Kirchhoff plate.

\begin{figure}
 \centering
 \begin{subfigure}[b]{0.45\textwidth}
   \centering \includegraphics[width=\textwidth]{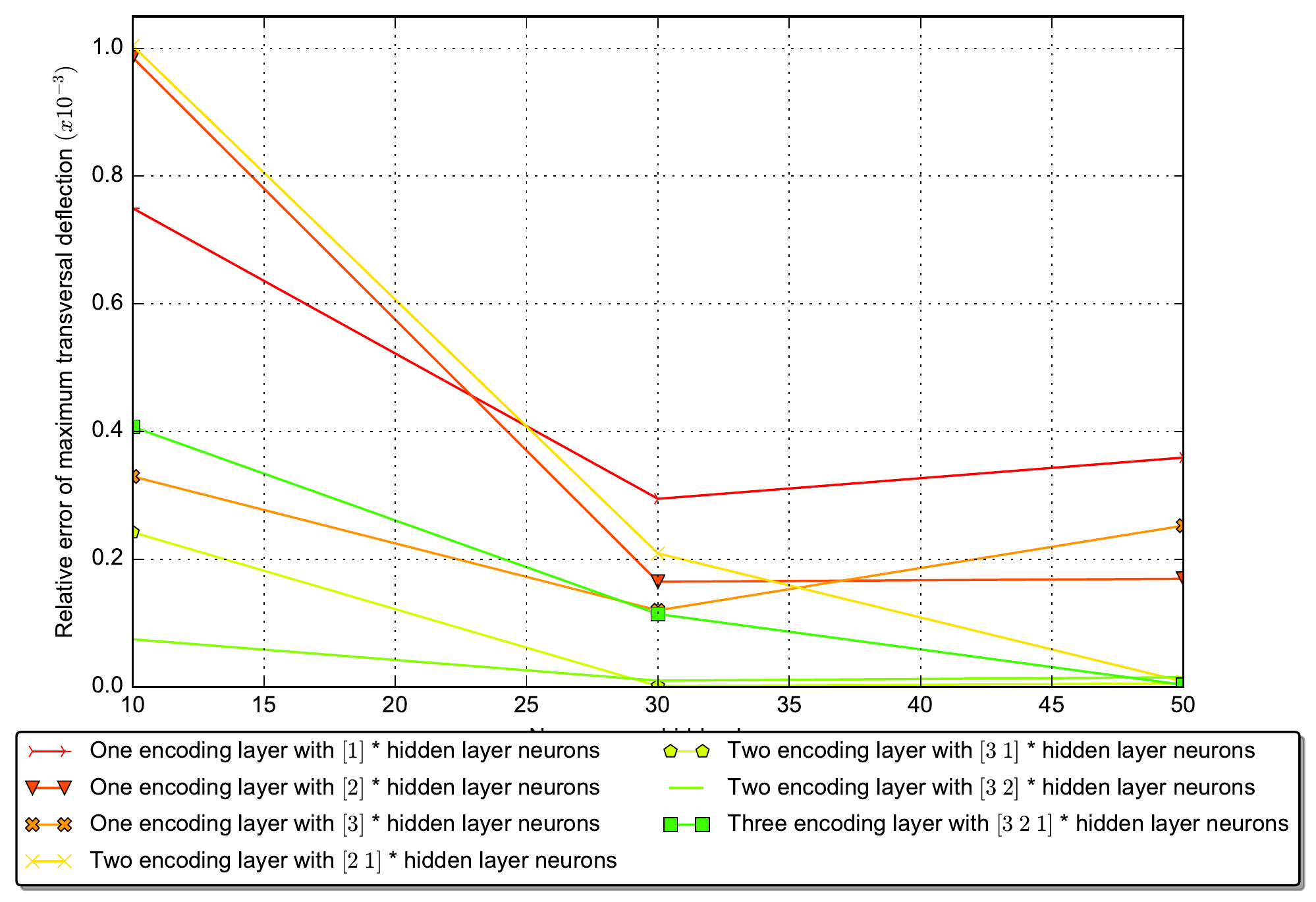}\hfill
   \caption{}
 \end{subfigure}\hspace{5pt}
 \begin{subfigure}[b] {0.45\textwidth}
   \centering \includegraphics[width=\textwidth]{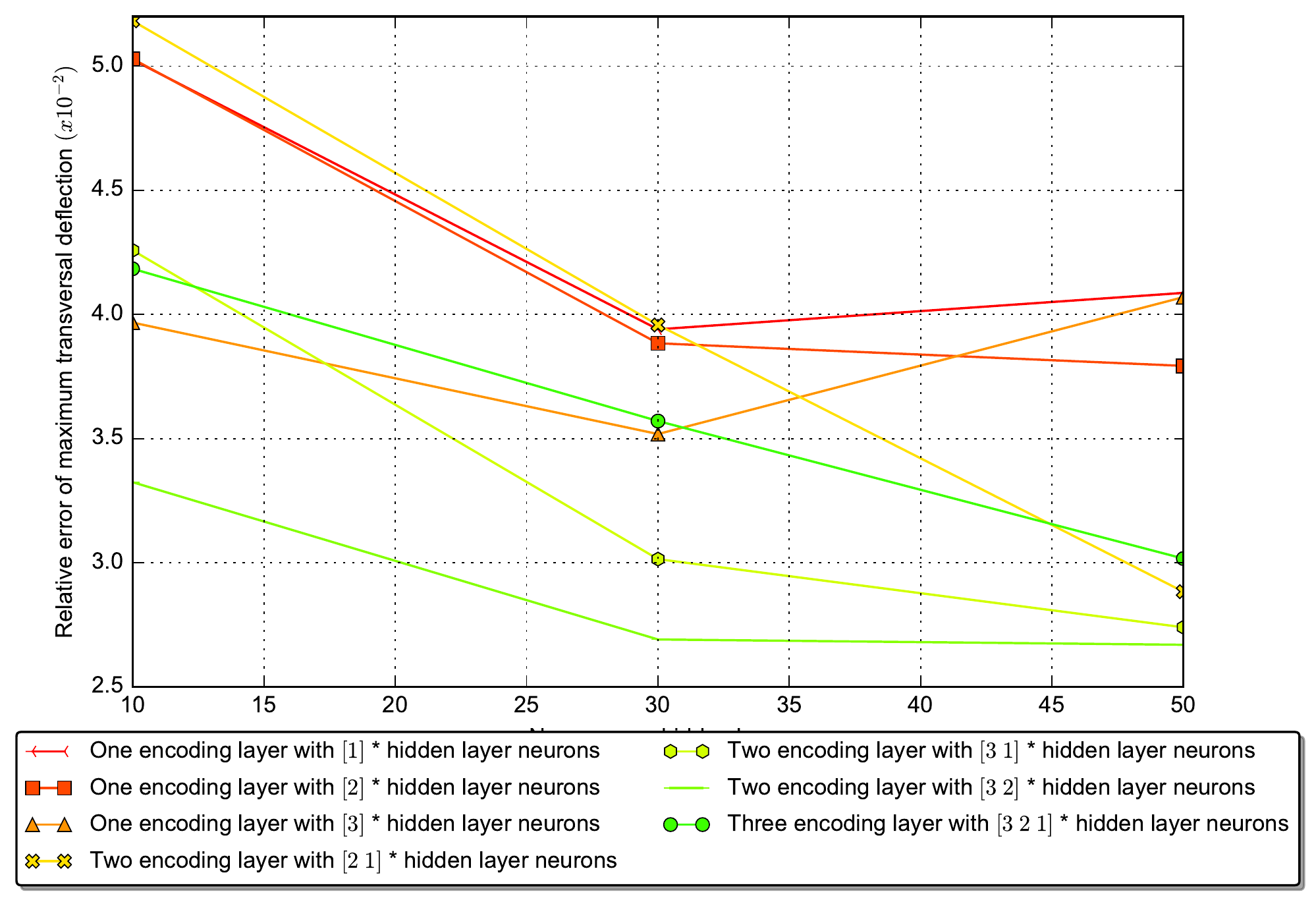}\hfill
   \caption{}
 \end{subfigure}\hspace{5pt}
 \caption{Relative error of $\left(a\right)$ maximum deflection and $\left(b\right)$ whole deflection predicted by different encoding layer configurations of a DNN with an autoencoder for the simply-supported plate.}
 \label{RelativerrDNNCODConfas}
\end{figure}

 Additionally, the deflection contour and the error contour are also shown in Figure~\ref{contouraspl}. We can conclude from Figure~\ref{contouraspl} that the transversal deflection of Kirchhoff plate predicted by our deep energy method agrees well with the analytical solution, even for plate with hole. 

\section{Concluding remarks}
\label{sec:conclusion}
The use of DNNs to solve boundary value problems has been explored. Several very relevant examples from computational mechanics have been solved using DNNs to build the approximation space. This constitutes a proof of concept for the possibility of approximating the solution of BVPs using concepts and tools coming from deep learning.

The first pillar of our approach is to use an energy characterizing the behaviour of the physical body under study. This energy is the basis for the construction of the loss function. The second fundamental idea is that the approximation space is defined by the architecture of the neural network. This two ideas lead to a finite sum nonconvex problem. In order to solve this problem, we remain faithful to the machine learning philosophy of using gradient based optimization approaches. This allows us to use standard libraries from the machine learning community such as TensorFlow. 

One of the advantages of the approach proposed here is that a wealth of concepts and tools developed by a very active community can be used. This also leads to ease of implementation. Near-mathematical notation can be used to define the loss function, which has the meaning of energy. Conceptually, this opens a door to the possibility of trying different mathematical models by defining the corresponding energies and implementing them in a very straightforward and transparent manner.

However, there are still some important caveats. The optimization problems that have to be solved are non-convex. The approximation spaces, although being very expressive, can be very difficult to analyze. Even linear problems lead to non-linear, non-convex discrete problems. This poses several challenges that have to be explored by the computational mechanics community.

\bibliographystyle{elsarticle-num}
\bibliography{biblio}
\end{document}